\pdfoutput=1

\documentclass[11pt]{article}

\usepackage[preprint]{acl}

\usepackage{times}
\usepackage{latexsym}
\usepackage{url}

\usepackage{listings}

\usepackage{algorithm}
\usepackage{algpseudocode}

\definecolor{backcolour}{rgb}{0.95,0.95,0.92}
\definecolor{codegreen}{rgb}{0,0.6,0}

\lstdefinestyle{myStyle}{
    backgroundcolor=\color{backcolour},   
    commentstyle=\color{codegreen},
    basicstyle=\ttfamily\footnotesize,
    breakatwhitespace=false,         
    breaklines=true,                 
    keepspaces=true,                 
    numbersep=5pt,                  
    showspaces=false,                
    showstringspaces=false,
    showtabs=false,                  
    tabsize=2,
}

\usepackage[most,skins,theorems]{tcolorbox}
\usepackage{mathtools}
\tcbset{
 aibox/.style={
 width=\linewidth,
 top=8pt,
 bottom=4pt,
 colback=blue!6!white,
 colframe=black,
 colbacktitle=black,
 enhanced,
 center,
 attach boxed title to top left={yshift=-0.1in,xshift=0.15in},
 boxed title style={boxrule=0pt,colframe=white,},
 }
}
\newtcolorbox{dmbox}[2][]{aibox,title=#2,#1}
\newtcolorbox{promptbox}{
  breakable,
  enhanced jigsaw,
  verbatim,
  colback=gray!5,
  colframe=gray!60,
  arc=2mm,
  boxrule=0.4pt,
  left=4mm,
  right=4mm,
  top=2mm,
  bottom=2mm,
  fontupper=\ttfamily\small
}

\usepackage{cleveref}

\usepackage{xspace}

\usepackage{amssymb}

\usepackage[T1]{fontenc}

\usepackage[utf8]{inputenc}

\usepackage{microtype}

\usepackage{inconsolata}

\usepackage{graphicx}

\newcommand{\datasource}[1]{\textbf{\texttt{#1}}}
\newcommand{\BLS}{\datasource{BLS}}
\newcommand{\USPF}{\datasource{EMM}}
\newcommand{\PRE}{\datasource{PRE}}
\newcommand{\SEC}{\datasource{SEC}}
\newcommand{\PG}{\datasource{PG}}
\newcommand{\PGS}{\datasource{PGS}}
\newcommand{\WLD}{\datasource{MFR}}
\newcommand{\ARXIV}{\datasource{cs.CL}}
\newcommand{\CSCL}{\texttt{cs.CL}}
\newcommand{\AVG}{\datasource{AVG}}
\newcommand{\FIND}{\datasource{FIND}}
\newcommand{\datasetname}{\FIND}

\newcommand{\model}[1]{\texttt{#1}}

\usepackage{colortbl} 
\usepackage{multirow}
\usepackage{listings} 
\usepackage{dsfont}

\colorlet{punct}{red!60!black}
\definecolor{background}{HTML}{EEEEEE}
\definecolor{delim}{RGB}{20,105,176}
\definecolor{ANALYSIS}{HTML}{006400}

\colorlet{numb}{magenta!60!black}
\lstdefinelanguage{json}{
 basicstyle=\normalfont\ttfamily,
 numbers=left,
 numberstyle=\scriptsize,
 basicstyle=\scriptsize,
 stepnumber=1,
 numbersep=8pt,
 showstringspaces=false,
 breaklines=true,
 frame=lines,
 backgroundcolor=\color{background},
 literate=
 *{0}{{{\color{numb}0}}}{1}
 {1}{{{\color{numb}1}}}{1}
 {2}{{{\color{numb}2}}}{1}
 {3}{{{\color{numb}3}}}{1}
 {4}{{{\color{numb}4}}}{1}
 {5}{{{\color{numb}5}}}{1}
 {6}{{{\color{numb}6}}}{1}
 {7}{{{\color{numb}7}}}{1}
 {8}{{{\color{numb}8}}}{1}
 {9}{{{\color{numb}9}}}{1}
 {:}{{{\color{punct}{:}}}}{1}
 {,}{{{\color{punct}{,}}}}{1}
 {\{}{{{\color{delim}{\{}}}}{1}
 {\}}{{{\color{delim}{\}}}}}{1}
 {[}{{{\color{delim}{[}}}}{1}
 {]}{{{\color{delim}{]}}}}{1},
}
\usepackage[most,skins,theorems]{tcolorbox}
\usepackage{mathtools}
\tcbset{
 aibox/.style={
 width=\linewidth,
 top=8pt,
 bottom=4pt,
 colback=blue!6!white,
 colframe=black,
 colbacktitle=black,
 enhanced,
 center,
 attach boxed title to top left={yshift=-0.1in,xshift=0.15in},
 boxed title style={boxrule=0pt,colframe=white,},
 }
}

\usepackage{dsfont} 

\usepackage{booktabs}

\usepackage{subcaption}

\usepackage{enumitem}
\usepackage{gb4e}
\noautomath
\usepackage{bbm} 
\usepackage{amsmath} 

\usepackage{xfrac}
\usepackage{cleveref}

\usepackage{pgfplots}
\usepgfplotslibrary{groupplots}
\usetikzlibrary{patterns}
\usepackage{ulem}

\definecolor{PURPLE}{HTML}{993893}
\definecolor{ORANGE}{HTML}{ea8d02}
\definecolor{OliveGreen}{HTML}{3C8031}

\usepackage{pgfplotstable}
\pgfplotsset{every tick label/.append style={font=\footnotesize},compat=1.17}
\pgfplotsset{compat = newest}

\usepgfplotslibrary{colorbrewer}

\definecolor{codegreen}{rgb}{0,0.6,0}
\lstdefinestyle{promptstyle}{  
    commentstyle=\color{codegreen},
    keywordstyle=\color{blue},
    basicstyle=\ttfamily\small,
    breakatwhitespace=false,        
    breaklines=true,                 
    captionpos=b,                    
    keepspaces=true,                 
    showspaces=false,                
    showstringspaces=false,
    showtabs=false,                  
    tabsize=2
}
\newcommand{\ccell}[2]{%
  \begingroup
  \setlength{\fboxsep}{1pt} 
  \colorbox{#1}{{\small #2}}
  \endgroup
}

\newcommand{\SPAN}[1]{%
\ccell{PURPLE!25}{%
  \scriptsize\ttfamily #1%
}%
}
\usepackage{fancyvrb}

\newcommand{\SPANEX}[1]{%
\colorbox{purple!25}{%
\scriptsize\ttfamily\detokenize{#1}%
}%
}
\newcommand{\SPANPLAIN}[1]{%
\scriptsize\ttfamily\detokenize{#1}%
}

\newcommand{\I}{\mathcal{I}} 

\newcommand{\E}{\mathcal{E}} 

\newcommand{\e}{\varepsilon} 
\newcommand{\ee}{$\varepsilon$} 

\newcommand{\D}{\mathcal{D}} 
\newcommand{\DD}{$\D$\xspace} 

\newcommand{\T}{\delta} 
\title{On Finding Inconsistencies in Documents}

\author{
 \textbf{Charles J. Lovering}$^{\diamondsuit}$\textrm{,}
 \textbf{Seth Ebner}$^{\diamondsuit}$\textrm{,}
 \\
 \textbf{Brandon Smock},
 \textbf{Michael Krumdick},
 \textbf{Saad Rabbani},
 \\
 \textbf{Ahmed Muhammad},
\textbf{Varshini Reddy},
 \textbf{Chris Tanner}\\
$^{\diamondsuit}$Core Contributor\\
\texttt{\{first\}.\{last\}@kensho.com}
 }

\begin{document}
\maketitle
\begin{abstract}
Professionals in academia, law, and finance audit their documents because inconsistencies can result in monetary, reputational, and scientific costs. Language models (LMs) have the potential to dramatically speed up this auditing process. To understand their abilities, we introduce a benchmark, \datasetname{} (\textbf{F}inding \textbf{IN}consistencies in \textbf{D}ocuments), where each example is a document with an inconsistency inserted manually by a domain expert. Despite the documents being long, technical, and complex, the best-performing model (\model{gpt-5}) recovered 64\% of the inserted inconsistencies. Surprisingly, \model{gpt-5} also found undiscovered inconsistencies present in the original documents. For example, on 50 arXiv papers, we judged 136 out of 196 of the model's suggestions to be legitimate inconsistencies missed by the original authors. However, despite these findings, even the best models miss almost half of the inconsistencies in \datasetname{}, demonstrating that inconsistency detection is still a challenging task.
\end{abstract}

\section{Introduction}
Documents such as financial statements and scientific articles are audited for accuracy. Automated inconsistency detection could reduce the burden of auditing and improve document quality \citep{yesnoerror2024,zhang2025reviewingscientificpaperscritical,wang2025finauditing,xi2025flaws}. Inconsistency detectors could also be useful in automated workflows, such as for inspecting chain-of-thought traces \cite{he-etal-2025-large} or reviewing generated reports.

When model capabilities reach sufficient performance, automated consistency checking could become ubiquitous, just as grammar and spelling checkers are today. Even models with low precision can be helpful because inconsistencies often are hard to find but easy to verify. Moreover, because the task requires processing a lot of information over long documents, achieving high recall is difficult, meaning the performance ceiling is high.

\begin{figure}[t!]
   \centering
   \includegraphics[width=\linewidth]{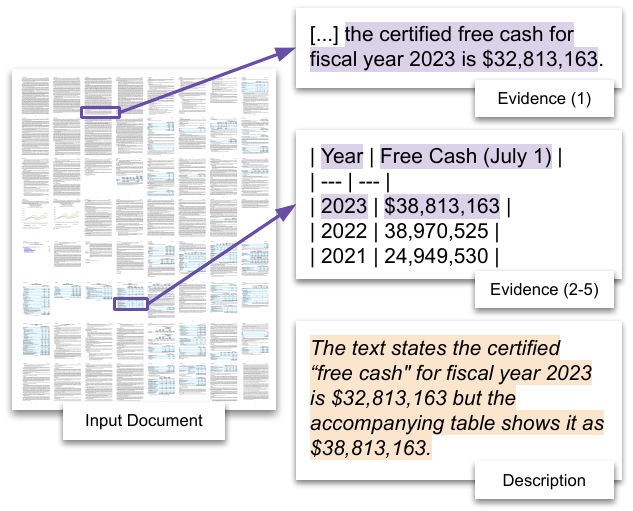}
   \caption{The task is to find inconsistencies within an input document. For each inconsistency, the model: (a) identifies the \textcolor{PURPLE}{evidence spans} that constitute the inconsistency and (b) generates a corresponding \textcolor{ORANGE}{description}.}
   \label{fig:teaser}
\end{figure}

\begin{dmbox}{Our Research Question}
How effectively do current language models find inconsistencies within documents?
\end{dmbox}

To approach this question, we introduce a benchmark, \datasetname{} (\textbf{F}inding \textbf{IN}consistencies in \textbf{D}ocuments), containing 375 test and 125 development problems. Each problem is a document with an inconsistency manually inserted by a financial expert. For each inconsistency it identifies, a model is expected to respond with \textit{evidence spans} from the document that demonstrate the inconsistency and a natural language \textit{description} of what makes those spans inconsistent. The documents are mostly financial reports, economic analyses, and general nonfiction. Documents in \datasetname{} are generally long: they have a median length of 35k tokens, and 25\% of documents contain over 78k tokens.

The length of the documents contributes to the challenge of the task. Auditing long documents is extremely resource intensive. The long context creates more opportunities for inconsistencies among related information. Experts interviewed in the course of this study noted that the review process to verify the consistency of tables in documents similar to those in \WLD{} (\Cref{sec:dataset}) averaged four hours. The full review process and error checking when preparing a document like a 10-K takes even longer, with large teams spending weeks on a single document.\footnote{Form 10-K is a report that a company files annually with the SEC in the United States.} Our results show performance decreases as document length increases (\Cref{sec:results}).

\paragraph{Problem Scope} \datasetname{} tests the ability of models to detect inconsistencies \textit{within} documents, excluding both \textit{resolving} inconsistencies and detecting inconsistencies that require \textit{external validation} (such as cross-referencing values against a database).

\paragraph{Contributions}  We {(1)} contribute \datasetname{}, an expert-annotated benchmark for inconsistency detection. We {(2)} show that the best-performing models recall over 60\% of inserted inconsistencies, and {(3)} find that, surprisingly, models find meaningful amounts of otherwise undiscovered inconsistencies with over 50\% precision. \textit{Models found five inconsistencies in late drafts of this paper (see \Cref{app:sec:paper:inconsistencies}).}

\section{Related Work}\label{sec:relatedwork}
\paragraph{Domains}
The ability to detect internal inconsistencies is useful in many domains, which prior work has explored. For example, \citet{deusser2023contradiction} and \citet{wang2025finauditing} examine inconsistencies in financial reports and filings. In the legal domain, \citet{andow2019policylint} consider privacy policies, while \citet{mantravadi2025legalwiz} and \citet{choudhury2025better} consider contracts and other legal documents. \citet{masuda2016detecting} consider system requirements and specifications. \citet{sagimbayeva2025misleading} construct a benchmark containing political statements. The Code.Debug task in $\infty$Bench \cite{zhang-etal-2024-bench} tests models' abilities to detect inconsistencies in code. Concurrent work has also investigated finding inconsistent statements in academic publications, often through the lens of automating the review process \cite{son2025aicoscientistsfailspota,zhang2025reviewingscientificpaperscritical,dycke2025automaticreviewersfaildetect,xi2025flaws,bianchi2025err}.

Broader domain datasets have also been constructed. ContraDoc \cite{li-etal-2024-contradoc} considers news, stories, and Wikipedia articles. WikiContradiction \cite{hsu-2021-contradiction} is based on Wikipedia articles tagged by editors as self-contradictory.

\paragraph{Knowledge Conflicts}
Information can be in conflict in many ways. For example, question answering and retrieval augmented generation systems must contend with retrieved documents being in conflict with each other or with a model's parametric knowledge \cite{xu-etal-2024-knowledge-conflicts,liu-roth-2025-conflicts,cattan2025dragged}. When conflicts occur, models perform worse \cite{chen-etal-2022-rich,hong-etal-2024-gullible,hou-2024-wikicontradict,liu-etal-2025-open,zeng2025worse} possibly because models detect these conflicts poorly \cite{pham-etal-2024-whos,jiayang-etal-2024-econ,kurfali-2025-conflicting,gokul2025contradictiondetectionragsystems}. Relatedly, corpus-level inconsistencies in Wikipedia have been studied by \citet{semnani2025detectingcorpuslevelknowledgeinconsistencies}. \citet{tyen-etal-2024-llms} and \citet{he-etal-2025-large} show that models struggle to detect errors within chain-of-thought streams and long reasoning chains.

Most similar to our work, \citet{kurfali-2025-conflicting} insert multiple conflicting ``needles'' into a document and ask a model a question involving the needles, finding that models frequently detect the conflict. We are also concerned with conflicts within a single document. However, unlike most prior work based on targeted queries, our work is based on finding all inconsistencies in a document.

\paragraph{Task Formulations}
We formulate our task to reflect the input-output interface of a useful tool, with extractive spans and explanatory descriptions. We simply prompt an LLM to find inconsistencies (as do \citet{deusser2023uncovering,li-etal-2024-contradoc,choudhury2025better,xi2025flaws}), leaving model development to future work. Other work has framed similar tasks using natural language inference \cite{hsu-2021-contradiction,deusser2023contradiction,song-etal-2025-introducing}, natural logic \cite{krishna2022proofver,aly-vlachos-2024-tabver}, and sheaves \cite{zadrozny2018sheafmodelcontradictionsdisagreements,huntsman2024prospectsinconsistencydetectionusing}.

\section{Experimental Design}
Our dataset consists of professionally prepared documents with at least the one known inconsistency we inserted in them. We task models to find all inconsistencies within a document. If we instead required the model to return at most one inconsistency, the model could fill the allotted slot in its response with an inconsistency from the original document that was undetected by the document's authors. Therefore, we formulate the task to be to return a list of inconsistencies. Our evaluation focuses on whether the model successfully returned the inserted inconsistency within that list.

\subsection{Data Definition}
An \textit{inconsistency} $\I$ has two parts (\Cref{fig:teaser}). First, the evidence $\E = \{\e_1, \dots, \e_n\}$ is a set of spans from the document $\D$. Second, the description $\T$ is a natural language explanation of how the evidence spans are inconsistent.

There is a unique span in the document (by reading order) where an inconsistency can first be detected. We call this span the \textit{trigger} (and include it in the evidence). The remaining evidence consists of the nearest set of spans preceding the trigger that are necessary to identify the inconsistency. Guidelines with examples of how to  handle span selection edge cases are linked in \Cref{app:sec:open}.

\subsection{Task Definition}\label{sec:design:task}
A model $M$ is to return all inconsistencies within a document $\D$ from data source $\mathcal{S}$. Each predicted inconsistency $\hat{\I}$ must conform to the XML-like formatting specified in the prompt (see \Cref{app:sec:span}) to be considered valid.

\subsection{Metric Definitions}\label{sec:design:metrics}
We evaluate the evidence, description, and their combination (the full task) with different metrics. Because the model returns a list of candidate inconsistencies, we must compare each candidate to the expected inconsistency. For each document, we score each candidate against the expected inconsistency and keep the best match, and then average these per-document scores across the dataset:
\begin{align*}
\Lambda &\vcentcolon= \frac{1}{|\mathcal{S}|} \sum_{(\D, \I)\in \mathcal{S}} \;\;\underset{\hat{\I} \in M(\D)}{\max}\;\; \lambda(\hat{\I}\,;\, \I, \D),
\end{align*}
where $\lambda$ is the underlying metric (evidence, description, or full task), $M(\D)$ is the model's list of predicted inconsistencies for document $\D$ from source $\mathcal{S}$, and $\I$ is the expected inconsistency. $\hat{\I}$ is a predicted inconsistency (or answer). We call $\Lambda$ the \textit{response-level} score to distinguish it from the underlying per-candidate metric $\lambda$. In all result tables below, we report the response-level scores $\Lambda_\E$, $\Lambda_\T$, and $\Lambda_\I$ for evidence, description, and full task, respectively.

\paragraph{Evidence Metric}
Our evidence metric measures how well the predicted spans align with the expert annotated spans. This lexical method is fast and easily reproducible but sensitive to character level differences, offering a different set of tradeoffs compared to the neural metrics used below. It uses bipartite matching with weights based on longest common substrings to align the predicted and reference spans. The metric's output ranges from 0 to 1, where a score of 1 for a predicted inconsistency indicates an exact match against the expected inconsistency, and 0 indicates no alignment. \Cref{app:sec:evidence_metric} and \Cref{app:alg} contain the design and algorithm details as well as discussion of a lenient version of the metric. We report the response-level score $\Lambda_\E \in [0,1]$ in \Cref{tbl:results:metrics}.

\paragraph{Description Metric}\label{sec:design:description_metric}
To evaluate the predicted description against the reference description, we use BLEURT \cite{sellam-etal-2020-bleurt}, a trained neural text generation metric that evaluates how well the predicted text conveys the meaning of the reference text.\footnote{We use the recommended \texttt{BLEURT-20} checkpoint from the \url{https://github.com/lucadiliello/bleurt-pytorch} library as the underlying model.} Though unbounded, its range is typically between 0 and 1.\footnote{\url{https://github.com/google-research/bleurt/blob/cebe7e6/README.md\#interpreting-bleurt-scores}} 
These scores are best used to rank models on the same set of documents. We report the response-level score $\Lambda_\T$ in \Cref{tbl:results:metrics}.

\paragraph{Task Metric} 
We use LLM-as-a-judge (LMJ) to holistically compare a predicted inconsistency (answer) to the expected inconsistency. The LMJ score is conditioned on the predicted and expected inconsistencies (including all the evidence and descriptions) and the document. It returns a boolean (0 or 1).\footnote{Like \citet{arora2025healthbench}, we found \texttt{gpt-4.1} to be an effective judge, reaching an overall 0.96 Cohen's kappa with human judgments on 200 individual responses sampled with replacement from the development set with answers generated by all our tested models. Our LMJ grading prompt was informed by \citet{wei2024measuring}. These results suggest for our dataset that the LMJ is a useful proxy for a human grader.}  Because the metric is boolean, the response level version of this metric, $\Lambda_\I$, functions like recall for the inserted inconsistency.\footnote{Our setup allows multiple answers per response to be graded positively, but this is rare (\Cref{app:sec:results:duplicates}).} (In \Cref{sec:analysis:precision}, we estimate precision by manually grading inconsistencies from a top-performing model.) 

\section{Dataset}\label{sec:dataset}
To build \datasetname{}, we collected professional documents, described below, in the public domain (\BLS{}, \SEC{}, \USPF{}, \PG{}) or with direct usage permission (\PRE{}). \datasetname{} has 125 development and 375 test examples each containing an annotated inconsistency. The documents in this dataset are unlikely to be in current model training corpora; most were released in 2025 (some in late 2024) and \PRE{} is a private source. However, the auxiliary analysis data (\WLD{}, \ARXIV{}) are not date-protected (see \Cref{app:sec:dataset}). Our goal was to create an informative benchmark with expert annotations---not necessarily an evergreen dataset. Guidelines used for the annotation and data creation process are linked in \Cref{app:sec:open}.

\begin{figure}[]
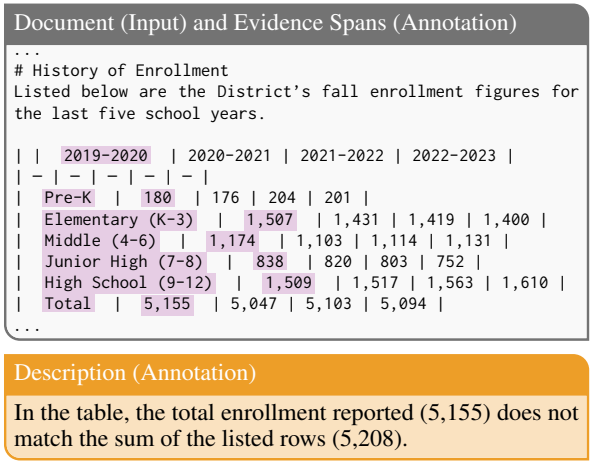

\small
\begin{tcolorbox}[
        enhanced jigsaw, 
        colback=gray!5!white, 
        colframe=gray!90!black, 
        boxrule=1pt, 
        arc=5pt, 
        sharp corners=downhill, 
        left=0pt, right=0pt, top=0pt, bottom=0pt,
        fontupper=\scriptsize\ttfamily,
        title=Document (Input) and Evidence Spans (Annotation)
    ]
    $\dots$\\
\# History of Enrollment\\
Listed below are the District’s fall enrollment figures for the last five school years.\\

|  | \SPAN{2019-2020} | 2020-2021 | 2021-2022 | 2022-2023 | \\
| --- | --- | --- | --- | --- |\\
| \SPAN{Pre-K} | \SPAN{180} | 176 | 204 | 201 |\\
| \SPAN{Elementary (K-3)} | \SPAN{1,507} | 1,431 | 1,419 | 1,400 |\\
| \SPAN{Middle (4-6)} | \SPAN{1,174} | 1,103 | 1,114 | 1,131 |\\
| \SPAN{Junior High (7-8)} | \SPAN{838} | 820 | 803 | 752 |\\
| \SPAN{High School (9-12)} | \SPAN{1,509} | 1,517 | 1,563 | 1,610 |\\
| \SPAN{Total} | \SPAN{5,155} | 5,047 | 5,103 | 5,094 | \\
$\dots$
\end{tcolorbox}
 \begin{tcolorbox}[
        enhanced jigsaw, 
        colback=ORANGE!25, 
        colframe=ORANGE!85, 
        boxrule=1pt, 
        arc=5pt, 
        sharp corners=downhill, 
        left=0pt, right=0pt, top=0pt, bottom=0pt,
        fontupper=\footnotesize,
        title=Description  (Annotation)
    ]
In the table, the total enrollment reported (5,155) does not match the sum of the listed rows (5,208).
\end{tcolorbox}

\caption{\textit{Numeric} inconsistency from \USPF{}.}
\label{fig:dataset:example:numeric}
\end{figure}

\subsection{Sources}\label{sec:design:sources}
We use documents from a variety of professional sources and have financial experts insert inconsistencies for the purposes of the task. Document samples are in \Cref{app:sec:dataset} (\Cref{app:fig:dataset:example:BLS,app:fig:dataset:example:PG,app:fig:dataset:example:PRE,app:fig:dataset:example:SEC,app:fig:dataset:example:USPF,app:fig:dataset:example:WLD}), and \Cref{fig:dataset:example:numeric,fig:dataset:example:non,fig:dataset:example:meta} show examples of inserted inconsistencies. The sources used in \datasetname{} are:

\paragraph{\BLS{}} (Bureau of Labor Statistics) Reports on employment and other economic trends in the U.S.\footnote{\url{https://www.bls.gov/}}

\paragraph{\PRE{}} S\&P presale reports that analyze pending bond deals, e.g., evaluating credit risk.\footnote{\url{https://www.spglobal.com/ratings/en/regulatory/presale-reports}}

\paragraph{\SEC{}} 10-Q reports covering finances and business operations, filed by public U.S. companies.\footnote{\url{https://www.sec.gov/edgar/search/}}

\paragraph{\USPF{}} Bond disclosures that U.S. municipalities must file when they issue debt.\footnote{\url{https://emma.msrb.org/}}

\paragraph{\PG{}} Non-fiction books from Project Gutenberg.\footnote{\url{https://www.gutenberg.org/}}\\

Additionally, we examine \textcolor{ANALYSIS}{\textit{analysis-only}} documents with live inconsistencies from:

\paragraph{\textcolor{ANALYSIS}{\ARXIV{}}} NLP and computational linguistics papers from the \CSCL{} category on arXiv.\footnote{\url{https://arxiv.org/list/cs.CL/recent}} Because arXiv provides version histories of papers, we were able to select documents where papers' authors found and fixed (at least) one inconsistency.\footnote{We manually inspected the diffs between successive versions of papers to identify such corrections of inconsistencies. Note that this process is biased toward inconsistencies that humans (the papers' authors) were able to find and resolve.} We also include ``control'' documents sampled from the \CSCL{} category without conditioning on the existence of a known inconsistency.

\paragraph{\textcolor{ANALYSIS}{\WLD{}}} Miscellaneous financial reports containing inconsistencies found in the wild by analysts, covering annual and ad-hoc financial reports. Most of the inconsistencies involve long tables.

\subsection{Why Mostly Finance Sources?}
The desiderata for our dataset were (1) permissible licenses, (2) high quality documents, and (3) diversity of content (tables and analyses, document lengths). The chosen finance and economic sources covered these needs. {Also, prior work (\Cref{sec:relatedwork}) suggests that models have low recall in identifying inconsistencies, but recall is critical in the finance domain, as even minor inconsistencies can result in penalties, making \datasetname{} an appropriate testbed for model development.}

\subsection{Annotators and Quality Control}
To build a practical taxonomy of inconsistency types, we interviewed financial experts and then collaborated with a financial expert annotation team to create a annotation scheme. Then, the annotation team inserted inconsistencies into the documents paired with iterative review and validation. After annotation, we reviewed the dataset again. On average, creating, annotating, and reviewing each inconsistency took two hours. The full process is outlined in \Cref{app:fig:annotation}.

\subsection{Inconsistency Types}
The main types we consider are high-level distinctions of what type of information is in conflict. These types are listed below, with additional axes---modality (text, table, or both), scope (short or long range), and number of evidence spans---described in our guidelines. For each document, we provided a setting of the axes for the annotators to use when creating the inconsistency. Because creating inconsistencies is difficult, these settings served as suggestions, and the types were re-labeled after annotation based on the inconsistencies actually created. The distribution of inconsistency types is shown in \Cref{tab:data-type-distribution} and includes:

\begin{exe}
\ex \label{experiments:ex2:1} {\textit{Numeric} inconsistencies, e.g., incorrect sums or averages. See \Cref{fig:dataset:example:numeric}.}
\ex \label{experiments:ex2:2} {\textit{Non-numeric}, conceptual, or logical inconsistencies, e.g., incorrect analysis or application of domain knowledge. See \Cref{fig:dataset:example:non}.}
\ex \label{experiments:ex2:3}  {\textit{Structural} inconsistencies, e.g., the absence of expected information or an incorrect reference to a table. See \Cref{fig:dataset:example:meta}.}
\end{exe}

\begin{table}[ht!]
\small
\centering
\begin{tabular}{@{}l|@{\hspace{4pt}}r@{\hspace{4pt}}r@{\hspace{4pt}}r@{\hspace{4pt}}r@{\hspace{4pt}}r@{\hspace{4pt}}r|@{\hspace{4pt}}r}
\toprule
         &              & \BLS{} & \PRE{} & \SEC{} & \USPF{} & \PG{} & Total \\
\midrule
\multirow{3}{4em}{Content}  & Numeric      & 48  & 48  & 40  & 40 & 28   & 204   \\
         & Non-numeric  & 13  & 14  & 12  & 10 & 14   & 63    \\
         & Structural         & 14  & 13  & 23  & 25 & 33   & 108  \\
\bottomrule
\end{tabular}
\caption{Distribution of inconsistency types in \datasetname{}.}
\label{tab:data-type-distribution}
\end{table}

\begin{figure}[]
\small
\begin{tcolorbox}[
        enhanced jigsaw, 
        colback=gray!5!white, 
        colframe=gray!90!black, 
        boxrule=1pt, 
        arc=5pt, 
        sharp corners=downhill, 
        left=0pt, right=0pt, top=0pt, bottom=0pt,
        fontupper=\scriptsize\ttfamily,
        title=Document (Input) and Evidence Spans (Annotation)
    ] 
\SPAN{Deferred revenue expected to be recognized as revenue more} \SPAN{than one year} subsequent to the balance sheet date is \SPAN{included in short-term liabilities} on the condensed consolidated balance sheets.
\end{tcolorbox}
 \begin{tcolorbox}[
        enhanced jigsaw, 
        colback=ORANGE!25, 
        colframe=ORANGE!85, 
        boxrule=1pt, 
        arc=5pt, 
        sharp corners=downhill, 
        fontupper=\footnotesize,
        left=0pt, right=0pt, top=0pt, bottom=0pt,
        title=Description  (Annotation)
    ]
The sentence incorrectly classifies deferred revenue expected after one year as a short-term liability. This creates a contradiction because such revenue should be reported as a long-term liability, based on standard accounting timelines.
\end{tcolorbox}

\caption{\textit{Non-numeric} inconsistency from \SEC{}.}
\label{fig:dataset:example:non}
\end{figure}



\begin{figure}[]
\small
\begin{tcolorbox}[
        enhanced jigsaw, 
        colback=gray!5!white, 
        colframe=gray!90!black, 
        boxrule=1pt, 
        arc=5pt, 
        sharp corners=downhill, 
        left=0pt, right=0pt, top=0pt, bottom=0pt,
        fontupper=\scriptsize\ttfamily,
        title=Document (Input) and Evidence Spans (Annotation)
    ]
\SPAN{These individuals would have preferred full-time employment} \SPAN{but were working part time because their hours had been} \SPAN{reduced or they were unable to find full-time jobs.} \SPAN{(See table 3.)} $\dots$

\#\#\# \SPAN{Table 3}. \SPAN{Employed people by disability status,} \SPAN{occupation, and sex, 2024 annual averages}

| Occupation | People with a disability | People with no disability | $\dots$
\end{tcolorbox}
 \begin{tcolorbox}[
        enhanced jigsaw, 
        colback=ORANGE!25, 
        colframe=ORANGE!85, 
        boxrule=1pt, 
        arc=5pt, 
        sharp corners=downhill, 
        fontupper=\footnotesize,
        left=0pt, right=0pt, top=0pt, bottom=0pt,
        title=Description  (Annotation)
        \footnotesize
    ]
The paragraph gives information on the workers disability employed part time and full time and in the end of the paragraph it states "see Table 3" but Table 3 instead gives information on the "Employed people by disability status, occupation, and sex..."
\end{tcolorbox}

\caption{\textit{Structural} inconsistency from \BLS{}.}
\label{fig:dataset:example:meta}
\end{figure}

\subsection{Dataset Analysis}\label{sec:dataset:analysis}
\datasetname{} consists of the first five sources listed in \Cref{sec:design:sources}, each with 75 test documents. See \Cref{tbl:dataset:test} for dataset statistics. The average document length varies from 10k tokens (\BLS{}) to 123k tokens (\USPF{}). Evidence statistics are similar across sources (albeit with high variance): approximately 5 spans per document and 11 tokens per evidence span.\footnote{Token counts measured with \texttt{tiktoken} (cl100k\_base), \url{https://github.com/openai/tiktoken}.} The remaining sources (\WLD{} and \ARXIV{}) are outside the main dataset. Notably, \WLD{} has a higher mean of 19 evidence spans per document but shorter average span length, mainly due to its inconsistencies pertaining to sums over long tables.

\begin{figure*}[ht!]
\centering
\input{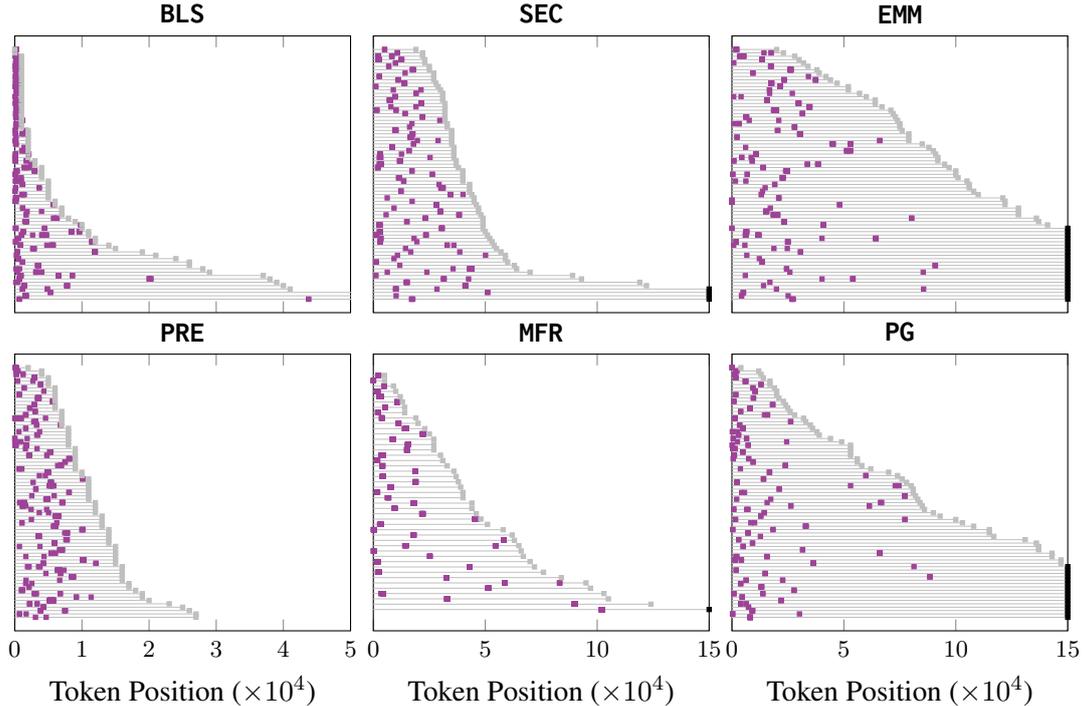}
\caption{\textbf{Absolute \textcolor{PURPLE}{Evidence Locations} and Test Document Lengths.} Each row corresponds to a document, with each row ending at a gray square indicating document length. \textcolor{PURPLE}{Purple squares} mark evidence locations. Black squares indicate that the document exceeds the displayed length. (For \BLS{} and \PRE{} the axis ends at $5\times10^4$.)}
\label{fig:evidence:sm}
\end{figure*}

\Cref{fig:evidence:sm} shows that most inconsistencies are located in the first half of the document, with the average position of the evidence being 30\% through the document (\Cref{tbl:dataset:test}). This reveals a limitation of our annotation approach: position within the documents was not explicitly varied during annotation. 

\begin{table}
\small
\centering
\begin{tabular}{@{}l@{\hspace{4pt}}r@{\hspace{4pt}}r@{\hspace{4pt}}r@{\hspace{4pt}}r@{\hspace{4pt}}r@{}}
\toprule
 & Docs (\DD) & kTok/\DD & \ee/\DD & Tok/\ee & \ee~Pos\% \\
\midrule
\BLS{} & 75 & 10$_{\pm14.8}$ & 5$_{\pm3.3}$ & 16$_{\pm60.9}$ & 28$_{\pm23.4}$ \\[1pt]
\PRE{} & 75 & 12$_{\pm5.5}$ & 7$_{\pm10.6}$ & 9$_{\pm7.9}$ & 37$_{\pm23.1}$ \\[1pt]
\SEC{} & 75 & 63$_{\pm102.8}$ & 5$_{\pm4.6}$ & 11$_{\pm8.6}$ & 37$_{\pm22.3}$ \\[1pt]
\USPF{} & 75 & 123$_{\pm82.7}$ & 6$_{\pm8.0}$ & 11$_{\pm9.8}$ & 24$_{\pm20.8}$ \\[1pt]
\PG{} & 75 & 109$_{\pm106.6}$ & 4$_{\pm2.6}$ & 8$_{\pm6.0}$ & 22$_{\pm24.1}$ \\
\midrule
\datasetname{} & 375 & 63$_{\pm89.6}$ & 5$_{\pm6.7}$ & 11$_{\pm28.3}$ & 30$_{\pm23.6}$ \\
\midrule
\textcolor{ANALYSIS}{\WLD{}} & 45 & 48$_{\pm33.6}$ & 19$_{\pm9.0}$ & 4$_{\pm1.3}$ & 37$_{\pm26.8}$ \\[1pt]
\textcolor{ANALYSIS}{\ARXIV{}} \footnotesize{Idnt} & 25 & 20$_{\pm10.1}$ & --- & --- & --- \\[1pt]
\textcolor{ANALYSIS}{\ARXIV{}} \footnotesize{Ctrl} & 25 & 25$_{\pm25.5}$ & --- & --- & --- \\
\bottomrule
\end{tabular}
\caption{\textbf{Statistics by document source}: number of documents (\DD), mean tokens per document (in thousands) (kTok/\DD), mean evidence spans per document (\ee/\DD), mean tokens per evidence span (Tok/\ee), and mean relative position of evidence within documents (\ee~Pos\%). Subscripts denote standard deviations. \datasetname{} row reports statistics pooled over the sources above; the bottom section reports  sources used for \textcolor{ANALYSIS}{analysis} in \Cref{sec:analysis:precision,sec:analysis:hard}.}
\label{tbl:dataset:test}
\end{table}

\section{Methods}\label{sec:methods}
\paragraph{Open Source} Our code will be released after review. Dataset and annotation information are in \Cref{app:sec:open}. Prompts are in \Cref{app:sec:prompts}. 

\paragraph{Models.} {\sloppy
We test {\small\model{gpt-oss-20b}}, {\small\model{gpt-oss-120b}},
{\small\model{gemma-3-12b-it}}, {\small\model{gemma-3-27b-it}} (open-source), and
{\small\model{claude-v4-sonnet}}, {\small\model{gpt-5-mini}}, {\small\model{gpt-5}}, {\small\model{o3-mini}},
{\small\model{o3}}, {\small\model{gemini-2.5-flash}}, {\small\model{gemini-2.5-pro}}
(closed-source).
}
\paragraph{Context Length Limits} Because of hardware limitations and compute costs, we set upper limits on the number of input tokens (105k) and the number of output tokens (20k). In the test set 83 out of 375 documents (at most, depending on the model and tokenizer) had their text truncated. However, only for the \model{gemma-3} models was any evidence included in the truncated text, and even then only for two documents. Moreover, because no model approaches a saturated score on this benchmark, we expect this truncation to be a reasonable tradeoff in terms of cost.

\begin{table}
\small
\centering
\begin{tabular}{@{}l@{\hspace{4pt}}r@{\hspace{4pt}}r@{\hspace{4pt}}r@{\hspace{4pt}}r@{\hspace{4pt}}r@{}}
\toprule
Model & Valid & kTok/\DD & $\hat{\I}$/\DD & \ee/$\hat{\I}$ & Tok/\ee \\
\midrule
gemma-3-12b-it & 99.8 & 0.6 & 3.4 & 2.4 & 16.8 \\[1pt]
gemma-3-27b-it & 99.8 & 0.9 & 4.7 & 2.5 & 21.7 \\[1pt]
gpt-oss-20b & 96.0 & 5.5 & 2.3 & 5.9 & 8.0 \\[1pt]
gpt-oss-120b & 99.5 & 2.7 & 1.9 & 3.2 & 16.9 \\
\midrule
sonnet-v4 & 97.6 & 4.9 & 1.8 & 4.9 & 8.3 \\[1pt]
gpt-5-mini & 100.0 & 5.9 & 2.6 & 2.6 & 31.9 \\[1pt]
gpt-5 & 99.0 & 11 & 3.2 & 2.5 & 25.7 \\[1pt]
o3-mini & 99.5 & 2.9 & 1.4 & 5.0 & 10.3 \\[1pt]
o3 & 99.5 & 5.4 & 1.9 & 2.3 & 23.2 \\[1pt]
gemini-2.5-flash & 94.3 & 12 & 4.3 & 4.1 & 27.5 \\[1pt]
gemini-2.5-pro & 99.5 & 13 & 5.1 & 3.7 & 17.8 \\
\bottomrule
\end{tabular}
\caption{\textbf{Model response statistics.} Format validity rate (\%), mean tokens per document (in thousands) (kTok$/\D$), mean answers per response ($\hat{\I}/\D$), mean evidence spans per answer ($\e/\hat{\I}$), and mean tokens per evidence span (Tok/\ee).}
\label{tbl:results_list_basics}
\end{table}

\begin{table*}[ht!]
\small
\centering
\begin{tabular}{@{}l@{\hspace{3pt}}r@{\hspace{3pt}}r@{\hspace{3pt}}r@{\hspace{3pt}}r@{\hspace{3pt}}r@{\hspace{3pt}}|@{\hspace{3pt}}r@{}}
\toprule
& \multicolumn{6}{c}{\textbf{Evidence Score ($\Lambda_\E$)}} \\
\midrule
Model & \BLS{} & \PRE{} & \SEC{} & \USPF{} & \PG{} & \AVG{} \\
\midrule
gemma-3-12b-it & 18 & 5 & 6 & 3 & 4 & 7 \\[1pt]
gemma-3-27b-it & 23 & 8 & 4 & 1 & 0 & 7 \\[1pt]
gpt-oss-20b & 30 & 11 & 3 & 2 & 2 & 10 \\[1pt]
gpt-oss-120b & 35 & 14 & 7 & 5 & 4 & 13 \\
\midrule
sonnet-v4 & \textbf{44} & \textbf{35} & 13 & 14 & 9 & \textbf{23} \\[1pt]
gpt-5-mini & 37 & 27 & 25 & 17 & 9 & \textbf{23} \\[1pt]
gpt-5 & \textbf{43} & \textbf{36} & \textbf{42} & \textbf{27} & \textbf{18} & \textbf{33} \\[1pt]
o3-mini & 23 & 5 & 1 & 0 & 1 & 6 \\[1pt]
o3 & \textbf{40} & 29 & 25 & 17 & 10 & \textbf{24} \\[1pt]
gemini-2.5-flash & \textbf{42} & 25 & 19 & 13 & 5 & \textbf{21} \\[1pt]
gemini-2.5-pro & \textbf{45} & \textbf{38} & \textbf{36} & \textbf{27} & 12 & \textbf{32} \\
\bottomrule
\end{tabular}
\begin{tabular}{@{}@{\hspace{3pt}}r@{\hspace{3pt}}r@{\hspace{3pt}}r@{\hspace{3pt}}r@{\hspace{3pt}}r@{\hspace{3pt}}|@{\hspace{3pt}}r@{}}
\toprule
\multicolumn{6}{c}{\textbf{Description Score ($\Lambda_\T$)}} \\
\midrule
\BLS{} & \PRE{} & \SEC{} & \USPF{} & \PG{} & \AVG{} \\
\midrule
43 & 37 & 35 & 34 & 33 & 36 \\[1pt]
45 & 38 & 36 & 34 & 34 & 37 \\[1pt]
45 & 32 & 27 & 28 & 28 & 32 \\[1pt]
47 & 36 & 34 & 34 & 32 & 37 \\
\midrule
\textbf{53} & 43 & 30 & 33 & 32 & \textbf{38} \\[1pt]
46 & 43 & 39 & 38 & 38 & \textbf{41} \\[1pt]
50 & \textbf{46} & \textbf{44} & \textbf{41} & \textbf{43} & \textbf{45} \\[1pt]
39 & 31 & 19 & 23 & 22 & 27 \\[1pt]
49 & 42 & 39 & 35 & 37 & \textbf{40} \\[1pt]
\textbf{53} & 45 & 38 & 39 & 36 & \textbf{42} \\[1pt]
\textbf{54} & \textbf{48} & \textbf{42} & \textbf{43} & \textbf{44} & \textbf{46} \\
\bottomrule
\end{tabular}
\begin{tabular}{@{}r@{\hspace{3pt}}r@{\hspace{3pt}}r@{\hspace{3pt}}r@{\hspace{3pt}}r|@{\hspace{4pt}}r@{}}
\toprule
\multicolumn{6}{c}{\textbf{Task Score ($\Lambda_\I$)}} \\
\midrule
\BLS{} & \PRE{} & \SEC{} & \USPF{} & \PG{} & \AVG{} \\
\midrule
57 & 25 & 7 & 3 & 9 & 20 \\[1pt]
61 & 27 & 7 & 8 & 12 & 23 \\[1pt]
21 & 3 & 1 & 0 & 3 & 6 \\[1pt]
28 & 3 & 0 & 0 & 1 & 6 \\
\midrule
\textbf{83} & 49 & 20 & 21 & 23 & 39 \\[1pt]
80 & 53 & 47 & \textbf{37} & 39 & 51 \\[1pt]
\textbf{87} & \textbf{67} & \textbf{73} & \textbf{43} & \textbf{52} & \textbf{64} \\[1pt]
56 & 17 & 8 & 3 & 12 & 19 \\[1pt]
\textbf{85} & 49 & 57 & 31 & 40 & 53 \\[1pt]
75 & 48 & 36 & 25 & 31 & 43 \\[1pt]
\textbf{88} & \textbf{71} & 61 & \textbf{40} & \textbf{45} & \textbf{61} \\
\bottomrule
\end{tabular}
\caption{Evidence, description, and task scores across the \datasetname{} test set (response-level metrics). For all three scores, higher is better. Scores are presented as percents (out of 100).}
\label{tbl:results:metrics}
\end{table*}

\section{Results}
\label{sec:results}

Below, we present results on \datasetname{}. {Across all results tables,} we bold the top-performing model and all models with no statistically significant difference from the top model ($\alpha = 0.05$). We use a paired t-test, and for the \AVG{} column we use a clustered test \citep{miller2024adding} over the sources. Standard errors are in \Cref{app:tbl:results_list_lexical_se}.
\paragraph{Model Response Statistics}
\Cref{tbl:results_list_basics} covers basic statistics of how different models responded. All models consistently generated output in the valid format (8 of 11 models gave valid output over 99\% of the time, and the worst case was still over 94\%). No model consistently saturated the available output token budget of 20k tokens. However, the top performing models, \model{gpt-5} and \model{gemini-2.5-pro}, used among the most tokens and generated among more answers per response compared to other closed-source models.

\paragraph{Overall Performance} 
The response-level metrics tell two basic stories (\Cref{tbl:results:metrics}). First, the open-source models underperform closed-source models: besides \model{o3-mini}, all closed-source models outperform all open-source models. Second, larger models outperform the smaller models within a series, with \model{gpt-5} and \model{gemini-2.5-pro} performing best overall, each with over 60\% recall. The description score averages fall in a narrow range (27--46, with 7 of 11 models scoring between 36--42), suggesting that the descriptions are of similar quality across models. The rankings formed by the three metrics over the model and source scores show strong agreement: pairwise Spearman rank correlations range from 0.86 to 0.92 (all $p < 0.001$).

\paragraph{Performance Per Data Source} The order of performance across sources for \model{gpt-5} and \model{gemini-2.5-pro} (the best models) largely matches the order of average document length, with worse performance on longer documents. Models did not saturate their output token budgets, indicating that lower recall on longer documents results from missed inconsistencies rather than output length constraints. 

For the tested models, \BLS{} was the easiest task (highest model scores), whereas \USPF{} and \PG{} were the most challenging (lowest model scores). Qualitatively, inconsistencies in \BLS{} are simpler because the documents tend to contain only a few tables and straightforward analysis. 

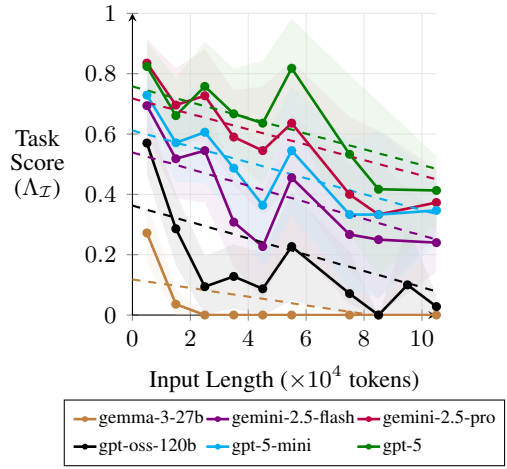
\begin{figure}[]
\centering
\begin{tikzpicture}[baseline]
\begin{axis}[
    width=4cm, height=4cm,
    xlabel={Input Length ($\times 10^4$ tokens)}, xlabel style={font=\small},
    ylabel={\shortstack{Task\\Score\\($\Lambda_\I$)}}, ylabel style={font=\small, rotate=-90},
    xmin=0, xmax=10.5, ymin=0, ymax=1,
    xtick={0,2,4,6,8,10}, ytick={0,0.2,0.4,0.6,0.8,1.0},
    grid=major, grid style={line width=.1pt, draw=gray!20},
    legend style={at={(0.5,-0.28)}, anchor=north, font=\scriptsize, nodes={inner xsep=0.45pt}},
    legend columns=3, legend cell align=left, legend image post style={scale=0.5},
    axis lines=left, scale only axis
]
\addplot[brown!20,fill=brown!20,fill opacity=0.3,draw=none,forget plot] coordinates {
    (0.50,0.174) (1.50,0.000) (2.50,0.000) (3.50,0.000) (4.50,0.000) (5.50,0.000) (7.50,0.000) (8.50,0.000) (10.50,0.000) (10.50,0.000) (8.50,0.000) (7.50,0.000) (5.50,0.000) (4.50,0.000) (3.50,0.000) (2.50,0.000) (1.50,0.086) (0.50,0.369)
} \closedcycle;
\addplot[violet!20,fill=violet!20,fill opacity=0.3,draw=none,forget plot] coordinates {
    (0.50,0.596) (1.50,0.386) (2.50,0.373) (3.50,0.161) (4.50,0.048) (5.50,0.242) (7.50,0.035) (8.50,0.000) (10.50,0.143) (10.50,0.337) (8.50,0.506) (7.50,0.498) (5.50,0.668) (4.50,0.407) (3.50,0.454) (2.50,0.718) (1.50,0.650) (0.50,0.793)
} \closedcycle;
\addplot[purple!20,fill=purple!20,fill opacity=0.3,draw=none,forget plot] coordinates {
    (0.50,0.756) (1.50,0.575) (2.50,0.573) (3.50,0.433) (4.50,0.332) (5.50,0.431) (7.50,0.143) (8.50,0.055) (10.50,0.263) (10.50,0.484) (8.50,0.612) (7.50,0.657) (5.50,0.842) (4.50,0.758) (3.50,0.746) (2.50,0.882) (1.50,0.818) (0.50,0.915)
} \closedcycle;
\addplot[black!20,fill=black!20,fill opacity=0.3,draw=none,forget plot] coordinates {
    (0.50,0.465) (1.50,0.166) (2.50,0.000) (3.50,0.022) (4.50,0.000) (5.50,0.048) (7.50,0.000) (8.50,0.000) (9.50,0.000) (10.50,0.000) (10.50,0.066) (9.50,0.296) (8.50,0.000) (7.50,0.211) (5.50,0.407) (4.50,0.205) (3.50,0.235) (2.50,0.196) (1.50,0.405) (0.50,0.675)
} \closedcycle;
\addplot[cyan!20,fill=cyan!20,fill opacity=0.3,draw=none,forget plot] coordinates {
    (0.50,0.634) (1.50,0.441) (2.50,0.437) (3.50,0.328) (4.50,0.158) (5.50,0.332) (7.50,0.086) (8.50,0.055) (10.50,0.238) (10.50,0.455) (8.50,0.612) (7.50,0.580) (5.50,0.758) (4.50,0.569) (3.50,0.646) (2.50,0.775) (1.50,0.702) (0.50,0.824)
} \closedcycle;
\addplot[green!50!black!20,fill=green!50!black!20,fill opacity=0.3,draw=none,forget plot] coordinates {
    (0.50,0.742) (1.50,0.536) (2.50,0.609) (3.50,0.517) (4.50,0.431) (5.50,0.653) (7.50,0.272) (8.50,0.125) (10.50,0.301) (10.50,0.526) (8.50,0.708) (7.50,0.795) (5.50,0.983) (4.50,0.842) (3.50,0.817) (2.50,0.906) (1.50,0.786) (0.50,0.905)
} \closedcycle;
\addplot[brown,line width=1pt,mark=*,mark size=1.2pt] coordinates {
    (0.50,0.272) (1.50,0.036) (2.50,0.000) (3.50,0.000) (4.50,0.000) (5.50,0.000) (7.50,0.000) (8.50,0.000) (10.50,0.000)
};
\addlegendentry{gemma-3-27b}
\addplot[violet,line width=1pt,mark=*,mark size=1.2pt] coordinates {
    (0.50,0.694) (1.50,0.518) (2.50,0.545) (3.50,0.308) (4.50,0.227) (5.50,0.455) (7.50,0.267) (8.50,0.250) (10.50,0.240)
};
\addlegendentry{gemini-2.5-flash}
\addplot[purple,line width=1pt,mark=*,mark size=1.2pt] coordinates {
    (0.50,0.835) (1.50,0.696) (2.50,0.727) (3.50,0.590) (4.50,0.545) (5.50,0.636) (7.50,0.400) (8.50,0.333) (10.50,0.373)
};
\addlegendentry{gemini-2.5-pro}
\addplot[black,line width=1pt,mark=*,mark size=1.2pt] coordinates {
    (0.50,0.570) (1.50,0.286) (2.50,0.094) (3.50,0.128) (4.50,0.087) (5.50,0.227) (7.50,0.071) (8.50,0.000) (9.50,0.100) (10.50,0.028)
};
\addlegendentry{gpt-oss-120b}
\addplot[cyan,line width=1pt,mark=*,mark size=1.2pt] coordinates {
    (0.50,0.729) (1.50,0.571) (2.50,0.606) (3.50,0.487) (4.50,0.364) (5.50,0.545) (7.50,0.333) (8.50,0.333) (10.50,0.347)
};
\addlegendentry{gpt-5-mini}
\addplot[green!50!black,line width=1pt,mark=*,mark size=1.2pt] coordinates {
    (0.50,0.824) (1.50,0.661) (2.50,0.758) (3.50,0.667) (4.50,0.636) (5.50,0.818) (7.50,0.533) (8.50,0.417) (10.50,0.413)
};
\addlegendentry{gpt-5}
\addplot[brown,dashed,line width=0.8pt,forget plot] coordinates {
    (0.00,0.118) (0.21,0.115) (0.43,0.112) (0.64,0.109) (0.86,0.106) (1.07,0.103) (1.29,0.100) (1.50,0.097) (1.71,0.094) (1.93,0.091) (2.14,0.088) (2.36,0.085) (2.57,0.081) (2.79,0.078) (3.00,0.075) (3.21,0.072) (3.43,0.069) (3.64,0.066) (3.86,0.063) (4.07,0.060) (4.29,0.057) (4.50,0.054) (4.71,0.051) (4.93,0.048) (5.14,0.045) (5.36,0.042) (5.57,0.038) (5.79,0.035) (6.00,0.032) (6.21,0.029) (6.43,0.026) (6.64,0.023) (6.86,0.020) (7.07,0.017) (7.29,0.014) (7.50,0.011) (7.71,0.008) (7.93,0.005) (8.14,0.002) (8.36,-0.001) (8.57,-0.005) (8.79,-0.008) (9.00,-0.011) (9.21,-0.014) (9.43,-0.017) (9.64,-0.020) (9.86,-0.023) (10.07,-0.026) (10.29,-0.029) (10.50,-0.032)
};
\addplot[violet,dashed,line width=0.8pt,forget plot] coordinates {
    (0.00,0.539) (0.21,0.533) (0.43,0.527) (0.64,0.521) (0.86,0.515) (1.07,0.509) (1.29,0.504) (1.50,0.498) (1.71,0.492) (1.93,0.486) (2.14,0.480) (2.36,0.474) (2.57,0.468) (2.79,0.462) (3.00,0.456) (3.21,0.450) (3.43,0.445) (3.64,0.439) (3.86,0.433) (4.07,0.427) (4.29,0.421) (4.50,0.415) (4.71,0.409) (4.93,0.403) (5.14,0.397) (5.36,0.391) (5.57,0.386) (5.79,0.380) (6.00,0.374) (6.21,0.368) (6.43,0.362) (6.64,0.356) (6.86,0.350) (7.07,0.344) (7.29,0.338) (7.50,0.332) (7.71,0.327) (7.93,0.321) (8.14,0.315) (8.36,0.309) (8.57,0.303) (8.79,0.297) (9.00,0.291) (9.21,0.285) (9.43,0.279) (9.64,0.273) (9.86,0.268) (10.07,0.262) (10.29,0.256) (10.50,0.250)
};
\addplot[purple,dashed,line width=0.8pt,forget plot] coordinates {
    (0.00,0.718) (0.21,0.712) (0.43,0.707) (0.64,0.701) (0.86,0.696) (1.07,0.690) (1.29,0.685) (1.50,0.680) (1.71,0.674) (1.93,0.669) (2.14,0.663) (2.36,0.658) (2.57,0.652) (2.79,0.647) (3.00,0.641) (3.21,0.636) (3.43,0.630) (3.64,0.625) (3.86,0.619) (4.07,0.614) (4.29,0.608) (4.50,0.603) (4.71,0.597) (4.93,0.592) (5.14,0.586) (5.36,0.581) (5.57,0.576) (5.79,0.570) (6.00,0.565) (6.21,0.559) (6.43,0.554) (6.64,0.548) (6.86,0.543) (7.07,0.537) (7.29,0.532) (7.50,0.526) (7.71,0.521) (7.93,0.515) (8.14,0.510) (8.36,0.504) (8.57,0.499) (8.79,0.493) (9.00,0.488) (9.21,0.482) (9.43,0.477) (9.64,0.472) (9.86,0.466) (10.07,0.461) (10.29,0.455) (10.50,0.450)
};
\addplot[black,dashed,line width=0.8pt,forget plot] coordinates {
    (0.00,0.363) (0.21,0.357) (0.43,0.351) (0.64,0.345) (0.86,0.339) (1.07,0.334) (1.29,0.328) (1.50,0.322) (1.71,0.316) (1.93,0.310) (2.14,0.305) (2.36,0.299) (2.57,0.293) (2.79,0.287) (3.00,0.281) (3.21,0.276) (3.43,0.270) (3.64,0.264) (3.86,0.258) (4.07,0.252) (4.29,0.247) (4.50,0.241) (4.71,0.235) (4.93,0.229) (5.14,0.223) (5.36,0.218) (5.57,0.212) (5.79,0.206) (6.00,0.200) (6.21,0.194) (6.43,0.189) (6.64,0.183) (6.86,0.177) (7.07,0.171) (7.29,0.165) (7.50,0.160) (7.71,0.154) (7.93,0.148) (8.14,0.142) (8.36,0.136) (8.57,0.131) (8.79,0.125) (9.00,0.119) (9.21,0.113) (9.43,0.107) (9.64,0.102) (9.86,0.096) (10.07,0.090) (10.29,0.084) (10.50,0.078)
};
\addplot[cyan,dashed,line width=0.8pt,forget plot] coordinates {
    (0.00,0.612) (0.21,0.606) (0.43,0.600) (0.64,0.595) (0.86,0.589) (1.07,0.583) (1.29,0.578) (1.50,0.572) (1.71,0.566) (1.93,0.561) (2.14,0.555) (2.36,0.549) (2.57,0.544) (2.79,0.538) (3.00,0.533) (3.21,0.527) (3.43,0.521) (3.64,0.516) (3.86,0.510) (4.07,0.504) (4.29,0.499) (4.50,0.493) (4.71,0.487) (4.93,0.482) (5.14,0.476) (5.36,0.470) (5.57,0.465) (5.79,0.459) (6.00,0.453) (6.21,0.448) (6.43,0.442) (6.64,0.436) (6.86,0.431) (7.07,0.425) (7.29,0.419) (7.50,0.414) (7.71,0.408) (7.93,0.402) (8.14,0.397) (8.36,0.391) (8.57,0.385) (8.79,0.380) (9.00,0.374) (9.21,0.368) (9.43,0.363) (9.64,0.357) (9.86,0.352) (10.07,0.346) (10.29,0.340) (10.50,0.335)
};
\addplot[green!50!black,dashed,line width=0.8pt,forget plot] coordinates {
    (0.00,0.758) (0.21,0.752) (0.43,0.747) (0.64,0.741) (0.86,0.735) (1.07,0.730) (1.29,0.724) (1.50,0.719) (1.71,0.713) (1.93,0.708) (2.14,0.702) (2.36,0.696) (2.57,0.691) (2.79,0.685) (3.00,0.680) (3.21,0.674) (3.43,0.668) (3.64,0.663) (3.86,0.657) (4.07,0.652) (4.29,0.646) (4.50,0.641) (4.71,0.635) (4.93,0.629) (5.14,0.624) (5.36,0.618) (5.57,0.613) (5.79,0.607) (6.00,0.601) (6.21,0.596) (6.43,0.590) (6.64,0.585) (6.86,0.579) (7.07,0.573) (7.29,0.568) (7.50,0.562) (7.71,0.557) (7.93,0.551) (8.14,0.546) (8.36,0.540) (8.57,0.534) (8.79,0.529) (9.00,0.523) (9.21,0.518) (9.43,0.512) (9.64,0.506) (9.86,0.501) (10.07,0.495) (10.29,0.490) (10.50,0.484)
};
\end{axis}
\end{tikzpicture}
\caption{\textbf{Task Score ($\Lambda_\I$) vs. Input Length}. Dashed lines correspond to linear regression coefficients (p~<~0.05); See \Cref{app:tab:input_length_recall} for the coefficients.}
\label{fig:input_size_corr}
\end{figure}

\paragraph{Performance vs. Input Length and Inconsistency Type} \Cref{fig:input_size_corr} shows that input length correlates with task score, based on a linear regression conditioned on the data sources and inconsistency types as control variables (all effects are significant). For visual clarity, the figure displays only some of the models. See \Cref{app:tab:input_length_recall} for all regression coefficients. The inconsistency types also explain some of the result variance; models perform worse on structural examples compared to the other types. Both top-performing models' task scores degrade 2.6 percent for every 10k token increase in input length. Across models, the uneven counts of documents per length bin and performance on \PG{} explain the jump at 50k tokens (\Cref{tab:dataset_bin_distribution}). Note that the observed association between performance and input length is correlative, not causal.

\section{Analysis}
\subsection{Models Find Otherwise Undiscovered Inconsistencies}
\label{sec:analysis:precision}
\label{sec:analysis:arxiv}
We study predicted inconsistencies that were not matched to the expected inconsistencies and find that many are otherwise undetected inconsistencies in the original documents.

\paragraph{Methods} These inconsistencies are sometimes nuanced, calling for expert evaluation. To make this tractable, we limit this analysis only to outputs from the best model (\model{gpt-5} according to \Cref{tbl:results:metrics}). Furthermore, we manually grade the outputs on only 25 documents sampled from each source (200 total), including \WLD{} and both subsets of \ARXIV{}. Including \ARXIV{} expands the number of domains evaluated and provides a closer, albeit qualitative, look at how effective a tool could be in practice.

For documents in \datasetname{} and \WLD{}, the annotators of \datasetname{} were the graders, and for \ARXIV{}, two authors of this work served as the graders. The three grading categories are whether the predicted inconsistency is (1) real, (2) not an inconsistency, or (3) the grader is unsure. Our financial experts had relatively low agreement scores over the three grading categories. Cohen's kappa was 0.31 with agreement on 58.8\% of items. Because of the low agreement scores and the complexity of the task, the experts subsequently deliberated together to arrive at an agreement for each model prediction. For \ARXIV{}, similarly, Cohen's kappa was 0.38 with agreement on 67.8\% of items, and the disagreements were resolved through a deliberation phase. 

\paragraph{Results} On \datasetname{} (\Cref{tbl:analysis:precision}), the precision of \model{gpt-5}'s predictions averages above 50\%. We also report the usefulness of the answers, which captures cases where the model prediction is helpful, if not pointing out a strict inconsistency. This averaged almost 70\%. For the analysis-only sources, \WLD{} and \ARXIV{}, performance was also high: On both subsets of \ARXIV{}, precision exceeded 70\% and usefulness was over 90\%, indicating a small number of false positives. These high precision scores, paired with the number of inconsistencies predicted per response being 1.7 to 4.4, suggest that \textbf{models are finding previously undetected inconsistencies at meaningful rates.} Examples are shown in \Cref{app:fig:dataset:example:meta:unanticipated:ROW_73,app:fig:dataset:example:meta:unanticipated:ROW_86,app:fig:dataset:example:meta:unanticipated:ROW_119,app:fig:dataset:example:meta:unanticipated:ROW_228-2}.
\begin{table}[t]
\small
\centering
\begin{tabular}{@{}l|@{\hspace{4pt}}r@{\hspace{4pt}}r@{\hspace{4pt}}r@{\hspace{4pt}}r@{\hspace{4pt}}r|@{\hspace{4pt}}r|@{\hspace{4pt}}r@{\hspace{4pt}}r@{\hspace{4pt}}r@{}}
\toprule
 &  &  &  &  &  &  &  & \multicolumn{2}{c}{\textcolor{ANALYSIS}{\ARXIV{}}} \\
 & \BLS{} & \PRE{} & \SEC{} & \USPF{} & \PG{} & \AVG{} & \textcolor{ANALYSIS}{\WLD{}} & \textcolor{ANALYSIS}{\footnotesize{Ctrl}} & \textcolor{ANALYSIS}{\footnotesize{Idnt}} \\
\midrule
$\hat{\I}/\D$ & 1.7 & 2.6 & 2.4 & 4.0 & 2.4 & 2.6 & 4.0 & 4.4 & 3.9 \\
\midrule
P \scriptsize{(\%)} & 65.1 & 46.2 & 42.4 & 65.7 & 43.3 & 52.5 & 50.5 & 72.1 & 70.1 \\[1pt]
U \scriptsize{(\%)} & 69.8 & 70.8 & 64.4 & 74.7 & 53.3 & 66.6 & 56.6 & 91.9 & 90.7 \\
\bottomrule
\end{tabular}

\caption{$\hat{\I}/\D$ denotes the number of predicted answers per document. Precision (P) denotes the percent of answers that were judged as inconsistent. Useful (U) denotes the percent of answers that were judged as either inconsistent or helpful for a user. Predicted answers come from \model{gpt-5}. For \ARXIV{}, scores for the control dataset are denoted ``Ctrl'' and for the identified dataset are denoted ``Idnt''.}
\label{tbl:analysis:precision}
\end{table}

\subsection{Models Find Known Naturally Occurring Inconsistencies}\label{sec:analysis:hard}

Here, we focus on known naturally occurring inconsistencies---ones that either external experts discovered or the original document authors decided were worth fixing---and find that models do recover some of them.

\WLD{} has mostly numeric inconsistencies in long tables found by financial experts in public financial documents. While the precision for the suggested errors was high ($>50$\%) (see \Cref{tbl:analysis:precision}), the highest recall scores range from 7 to 11 percent, with the open source models finding only 0 to 4 percent of the errors (\Cref{app:tbl:results_list_hard_se}). All three metrics are lower than for data in \datasetname{} (\Cref{tbl:results:metrics}).\footnote{The standard errors are high (\Cref{app:tbl:results_list_hard_se}) in part because the total number of documents in this source is 45.}

For the identified subset of \ARXIV{}, those with a known inconsistency, we manually grade the model responses for recall. The known inconsistency was recovered in 12 of 25 documents. Again, the papers' authors fixed these inconsistencies in subsequent drafts, suggesting they were to some degree meaningful. These results suggest a mixed outlook: current models find inconsistencies with relatively high precision, making them useful, but depending on the document type, still miss other inconsistencies, meaning that alone they are not fully reliable.

\section{Discussion}\label{sec:discussion}
\textbf{Models are useful tools for inconsistency detection.}
This task is a practical needle-in-the-haystack task; the models find the needles at a high enough rate to be useful. \textbf{Moreover, the models find unanticipated ``needles'' in published documents.}
The observations in \Cref{sec:analysis:precision,sec:analysis:hard} suggest that using models in this way during document preparation can be helpful. In fact, we did so on a late version of our draft and found five novel inconsistencies, which are documented in \Cref{app:sec:paper:inconsistencies}. (None changed the conclusions of our work.) Despite the promising performance, the recall on this long-context task is about 60\% for the best models, which means that \textbf{the model misses almost half the expected inconsistencies within a set of documents,} suggesting more research is needed to make these capabilities fully reliable.

\section{Limitations}
\paragraph{Our task formulation focuses on finding inconsistencies, not fixing them.} Because the inconsistencies in \datasetname{} are based on information internal to the document, there is no fact checking or data verification component to the task (which would involve access to ground truth information in databases or other documents). Resolving the identified inconsistencies often requires information external to the document, such as from trusted knowledge bases or institutional knowledge.

\paragraph{Our task is part extractive, and so perhaps difficult for LLMs.} Our formulation of the task is part extractive (evidence) and part abstractive (description). Two limitations of using LLMs for finding evidence are that the predictions do not contain offsets into the document and consequently that checks are required to ensure the evidence text does indeed appear in the document.

\paragraph{The span boundaries can be subjective.} Decisions regarding evidence span boundaries and segmentation can be subjective, leading to some ambiguity in what the correct span(s) should be, so we include a lenient version of the evidence metric that ignores the delineation between one span and the next (see \Cref{app:sec:evidence_metric} for more discussion). For example, consider: ``the scores are always positive and we show that in Table 2.'' If this were evidence, then it could be captured with one span or two, such as ``scores are always positive'' and ``show that in Table 2''. To help mitigate this limitation, we encourage our annotators to prefer shorter spans, and we accommodate both strict and lenient measures of success during evaluation.

\paragraph{\datasetname{} has some distributional limitations.} The documents are all in English due to the language knowledge of the annotators and researchers. Each example considers a single document, while finding inconsistencies across documents is also an interesting use case \cite{semnani2025detectingcorpuslevelknowledgeinconsistencies}. The locations of the inserted inconsistencies are biased toward the first halves of the documents (see \Cref{tbl:dataset:test} and \Cref{fig:evidence:sm}). Additionally, none of the inconsistencies concern visual information in plots, which presents an exciting avenue for future work.

\paragraph{The inconsistencies found are meaningful but limited in scope.} Our interviews with subject matter experts (largely in the finance domain) highlighted how even simple inconsistencies can have significant monetary penalties. However, within the domain of \CSCL{} papers, the range of inconsistencies found by \model{gpt-5} can be understood in a different (if compatible) way. First, some of the papers were not intended to be the final version and so the authors may have been more lax when preparing the documents. Moreover, while some inconsistencies cast their results in a better light, there were no examples where the inconsistency changed the conclusions of the work. So, while the articles would be made more sound by resolving the identified inconsistencies, the model did not find any flaws that rose to the point of misleading the reader. \textit{Likewise, when we tested this draft, we found small issues we were happy to have been shown and to fix, but they did not change the conclusions of our work.}

\section*{Acknowledgments}
Special thanks to Blake MacDonald, Craig Schmidt, and Adam Wiemerslage for their helpful discussions and advice. Thanks also to Sudhker Gundlapally, Shruti Hajirnis, and Akshata Joshi for help with understanding inconsistencies in financial documents. Thanks to arXiv for use of its open access interoperability. We appreciate all the dataset sources for their services. 

\bibliography{custom,anthology-1,anthology-2}
\appendix

\counterwithin{figure}{section}
\counterwithin{table}{section}
\counterwithin{algorithm}{section}
\renewcommand{\thefigure}{\thesection.\arabic{figure}}
\renewcommand{\thealgorithm}{\thesection.\arabic{algorithm}}
\renewcommand{\thetable}{\thesection.\arabic{table}}

\section*{Housekeeping}\label{app:sec:contents}
\begin{enumerate}
    \item This appendix has a large number of figures, so we provide a table of contents below to help navigate the sections.
    \item Claude Code (sonnet/v4.5) was used to help generate the tables and figures in this paper (especially with \texttt{pgfplots}).
\end{enumerate}

\begin{table}[ht!]
\small
    \centering
\begin{tabular}{lp{4cm}}
\toprule
\Cref{app:sec:paper:inconsistencies} & Examples of Undiscovered Inconsistencies From A Late Draft of This Paper \\
\Cref{app:sec:undiscovered} & Examples of Undiscovered Inconsistencies in \ARXIV{} and \PG{}  \\
\Cref{app:sec:open} & Open-Source and Links  \\
\Cref{app:sec:reproducing} & Reproducing this Work and Implementation Details \\
\Cref{app:sec:evidence_metric} & Evidence Metric \\
\Cref{app:sec:description_metric} & Description Metric \\
\Cref{app:sec:dataset} & Dataset \\
\Cref{app:sec:results} & More Results \\
\Cref{app:sec:prompts} & Prompts \\
\bottomrule
\end{tabular}
\end{table}

\section{Examples of Undiscovered Inconsistencies From A Late Draft of This Paper}\label{app:sec:paper:inconsistencies}
We ran \model{gpt-5} and \model{gemini-2.5-pro} on a late draft of this work. We put one intentional inconsistency within the work, which \model{gemini-2.5-pro} found. Together, the two models found five meaningful and previously undiscovered issues and recovered the intentional error. Overall, this anecdotal case study suggests that models are helpful for real-world use.

\begin{figure*}[ht!]
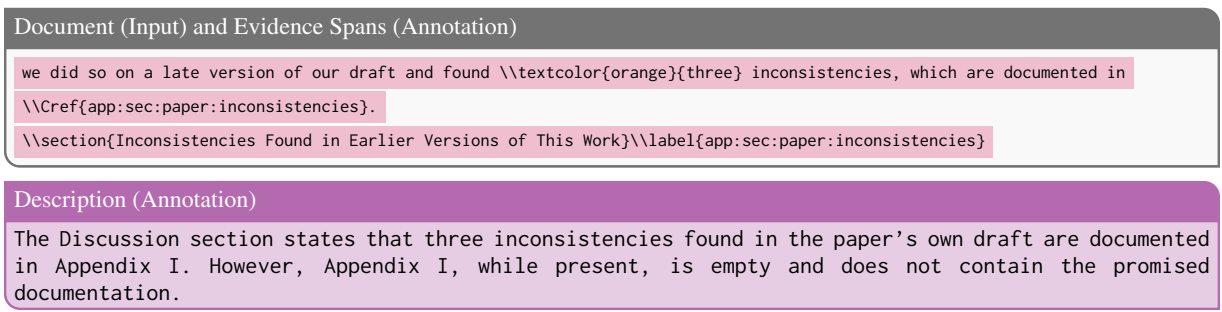

\small
\begin{tcolorbox}[
        enhanced jigsaw, 
        colback=gray!5!white, 
        colframe=gray!90!black, 
        boxrule=1pt, 
        arc=5pt, 
        sharp corners=downhill, 
        left=0pt, right=0pt, top=0pt, bottom=0pt,
        fontupper=\scriptsize\ttfamily,
        title=Document (Input) and Evidence Spans (Annotation)
    ]
\SPANEX{we did so on a late version of our draft and found \\textcolor{orange}{three} inconsistencies, which are documented in}\\
\SPANEX{\\Cref{app:sec:paper:inconsistencies}.}\\
\SPANEX{\\section{Inconsistencies Found in Earlier Versions of This Work}\\label{app:sec:paper:inconsistencies}}
\end{tcolorbox}
 \begin{tcolorbox}[
        enhanced jigsaw, 
        colback=PURPLE!25, 
        colframe=PURPLE!75, 
        boxrule=1pt, 
        arc=5pt, 
        sharp corners=downhill, 
                fontupper=\ttfamily,
        left=0pt, right=0pt, top=0pt, bottom=0pt,
        title=Description  (Annotation)
    ]
    The text claims three inconsistencies are documented in the cited appendix section, but the referenced section appears without the promised documentation, indicating missing expected content.
\end{tcolorbox}

\caption{\textcolor{OliveGreen}{\textbf{Real}} inconsistency found by \model{gpt-5} in an earlier draft. \model{gemini-2.5-pro} found this same issue. We note that this would have been fixed anyway, as it was a placeholder value for this analysis. \textit{We do not count this toward the number of novel inconsistencies found.}}
\label{app:fig:dataset:example:meta:unanticipated:GPT5_2_01}
\end{figure*}

\begin{figure*}[ht!]
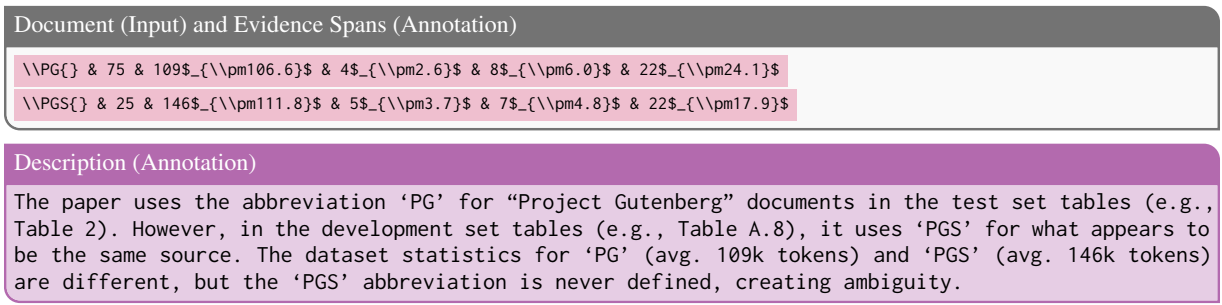

\small
\begin{tcolorbox}[
        enhanced jigsaw, 
        colback=gray!5!white, 
        colframe=gray!90!black, 
        boxrule=1pt, 
        arc=5pt, 
        sharp corners=downhill, 
        left=0pt, right=0pt, top=0pt, bottom=0pt,
        fontupper=\scriptsize\ttfamily,
        title=Document (Input) and Evidence Spans (Annotation)
    ]
\SPANEX{\\PG{} & 75 & 109$_{\\pm106.6}$ & 4$_{\\pm2.6}$ & 8$_{\\pm6.0}$ & 22$_{\\pm24.1}$}\\
\SPANEX{\\PGS{} & 25 & 146$_{\\pm111.8}$ & 5$_{\\pm3.7}$ & 7$_{\\pm4.8}$ & 22$_{\\pm17.9}$}
\end{tcolorbox}
 \begin{tcolorbox}[
        enhanced jigsaw, 
        colback=PURPLE!25, 
        colframe=PURPLE!75, 
        boxrule=1pt, 
        arc=5pt, 
        sharp corners=downhill, 
                fontupper=\ttfamily,
        left=0pt, right=0pt, top=0pt, bottom=0pt,
        title=Description  (Annotation)
    ]
The paper uses the abbreviation `PG' for ``Project Gutenberg'' documents in the test set tables (e.g., Table 2). However, in the development set tables (e.g., Table A.8), it uses `PGS' for what appears to be the same source. The dataset statistics for `PG' (avg. 109k tokens) and `PGS' (avg. 146k tokens) are different, but the `PGS' abbreviation is never defined, creating ambiguity.
\end{tcolorbox}

\caption{\textcolor{orange}{\textbf{Helpful}} comment made by \model{gemini-2.5-pro} on an earlier draft. We did not define PGS in the development section (an oversight). (PGS stands for ``Project Gutenberg Seen'', as the \PG{} development data may be leaked to models.)}
\label{app:fig:dataset:example:meta:unanticipated:GEMINI_2_05}
\end{figure*}

\begin{figure*}[ht!]
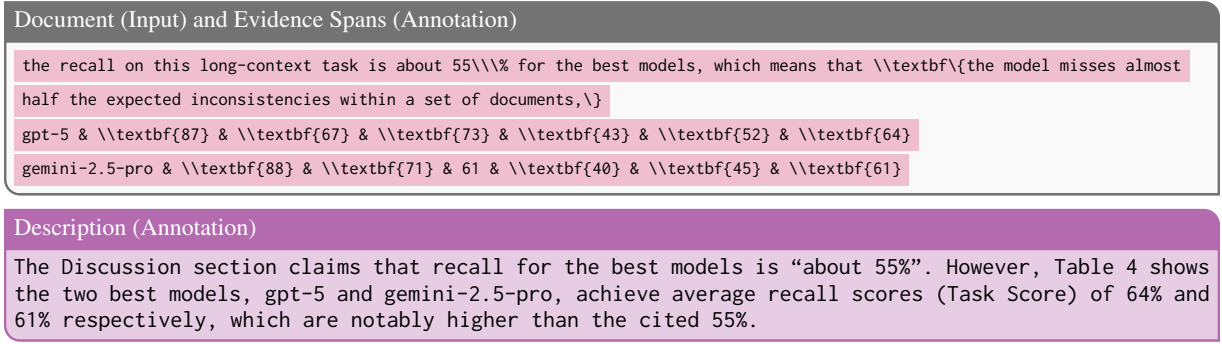

\small
\begin{tcolorbox}[
        enhanced jigsaw, 
        colback=gray!5!white, 
        colframe=gray!90!black, 
        boxrule=1pt, 
        arc=5pt, 
        sharp corners=downhill, 
        left=0pt, right=0pt, top=0pt, bottom=0pt,
        fontupper=\scriptsize\ttfamily,
        title=Document (Input) and Evidence Spans (Annotation)
    ]
        \SPANEX{the recall on this long-context task is about 55\\\% for the best models, which means that \\textbf\{the model misses almost}\\
        \SPANEX{half the expected inconsistencies within a set of documents,\}}\\
        \SPANEX{gpt-5 & \\textbf{87} & \\textbf{67} & \\textbf{73} & \\textbf{43} & \\textbf{52} & \\textbf{64}}\\
        \SPANEX{gemini-2.5-pro & \\textbf{88} & \\textbf{71} & 61 & \\textbf{40} & \\textbf{45} & \\textbf{61}}
\end{tcolorbox}
 \begin{tcolorbox}[
        enhanced jigsaw, 
        colback=PURPLE!25, 
        colframe=PURPLE!75, 
        boxrule=1pt, 
        arc=5pt, 
        sharp corners=downhill, 
                fontupper=\ttfamily,
        left=0pt, right=0pt, top=0pt, bottom=0pt,
        title=Description  (Annotation)
    ]
The Discussion section claims that recall for the best models is ``about 55\%''. However, Table 4 shows the two best models, gpt-5 and gemini-2.5-pro, achieve average recall scores (Task Score) of 64\% and 61\% respectively, which are notably higher than the cited 55\%.
\end{tcolorbox}
\caption{\textcolor{OliveGreen}{\textbf{Real}} inconsistency found by \model{gemini-2.5-pro}. We intentionally placed this inconsistency into a late draft of the paper. \textit{We do not count this toward the number of novel inconsistencies found.}}
\label{app:fig:dataset:example:meta:unanticipated:GEMINI_2_02}
\end{figure*}

\begin{figure*}[ht!]
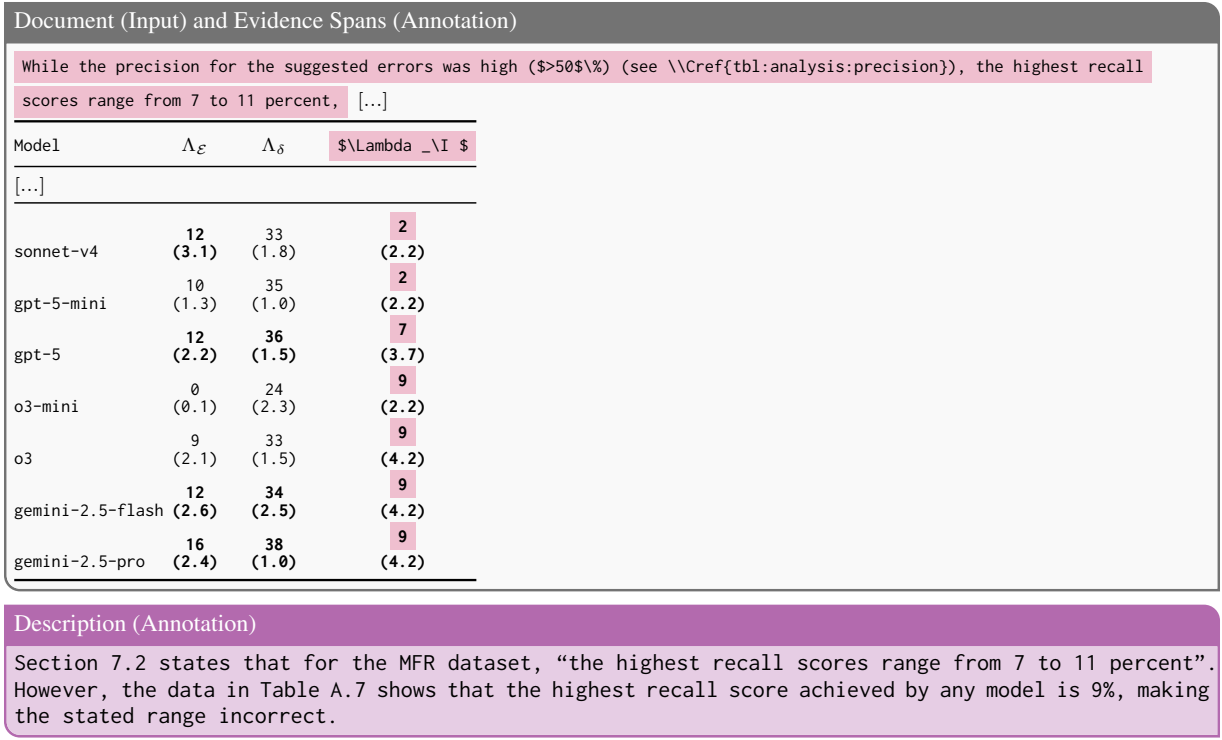

\small
\begin{tcolorbox}[
        enhanced jigsaw, 
        colback=gray!5!white, 
        colframe=gray!90!black, 
        boxrule=1pt, 
        arc=5pt, 
        sharp corners=downhill, 
        left=0pt, right=0pt, top=0pt, bottom=0pt,
        fontupper=\scriptsize\ttfamily,
        title=Document (Input) and Evidence Spans (Annotation)
    ]
\SPANEX{While the precision for the suggested errors was high ($>50$\%) (see \\Cref{tbl:analysis:precision}), the highest recall}\\
\SPANEX{scores range from 7 to 11 percent,}
    $[...]$\\
\begin{tabular}{@{}l@{\hspace{3pt}}ccc@{}}
\toprule
Model & $\Lambda_\E$ & $\Lambda_\T$ & \SPANEX{$\Lambda_\I$} \\
\midrule
$[...]$\\
\midrule
sonnet-v4 & \textbf{\shortstack{12 \\[-0.35ex] {\scriptsize (3.1)}}} & \shortstack{33 \\[-0.35ex] {\scriptsize (1.8)}} & \textbf{\shortstack{\SPANEX{2} \\[-0.35ex] {\scriptsize (2.2)}}} \\
gpt-5-mini & \shortstack{10 \\[-0.35ex] {\scriptsize (1.3)}} & \shortstack{35 \\[-0.35ex] {\scriptsize (1.0)}} & \textbf{\shortstack{\SPANEX{2} \\[-0.35ex] {\scriptsize (2.2)}}} \\
gpt-5 & \textbf{\shortstack{12 \\[-0.35ex] {\scriptsize (2.2)}}} & \textbf{\shortstack{36 \\[-0.35ex] {\scriptsize (1.5)}}} & \textbf{\shortstack{\SPANEX{7} \\[-0.35ex] {\scriptsize (3.7)}}} \\
o3-mini & \shortstack{0 \\[-0.35ex] {\scriptsize (0.1)}} & \shortstack{24 \\[-0.35ex] {\scriptsize (2.3)}} & \textbf{\shortstack{\SPANEX{9} \\[-0.35ex] {\scriptsize (2.2)}}} \\
o3 & \shortstack{9 \\[-0.35ex] {\scriptsize (2.1)}} & \shortstack{33 \\[-0.35ex] {\scriptsize (1.5)}} & \textbf{\shortstack{\SPANEX{9} \\[-0.35ex] {\scriptsize (4.2)}}} \\
gemini-2.5-flash & \textbf{\shortstack{12 \\[-0.35ex] {\scriptsize (2.6)}}} & \textbf{\shortstack{34 \\[-0.35ex] {\scriptsize (2.5)}}} & \textbf{\shortstack{\SPANEX{9} \\[-0.35ex] {\scriptsize (4.2)}}} \\
gemini-2.5-pro & \textbf{\shortstack{16 \\[-0.35ex] {\scriptsize (2.4)}}} & \textbf{\shortstack{38 \\[-0.35ex] {\scriptsize (1.0)}}} & \textbf{\shortstack{\SPANEX{9} \\[-0.35ex] {\scriptsize (4.2)}}} \\
\bottomrule
\end{tabular}
\end{tcolorbox}
 \begin{tcolorbox}[
        enhanced jigsaw, 
        colback=PURPLE!25, 
        colframe=PURPLE!75, 
        boxrule=1pt, 
        arc=5pt, 
        sharp corners=downhill, 
                fontupper=\ttfamily,
        left=0pt, right=0pt, top=0pt, bottom=0pt,
        title=Description  (Annotation)
    ]
Section 7.2 states that for the MFR dataset, ``the highest recall scores range from 7 to 11 percent''. However, the data in Table A.7 shows that the highest recall score achieved by any model is 9\%, making the stated range incorrect.
\end{tcolorbox}

\caption{\textcolor{OliveGreen}{\textbf{Real}} inconsistency found by \model{gemini-2.5-pro} in an earlier draft. The issue arose due to an out-of-date table in the appendix.}
\label{app:fig:dataset:example:meta:unanticipated:GEMINI_2_03}
\end{figure*}

\begin{figure*}[ht!]
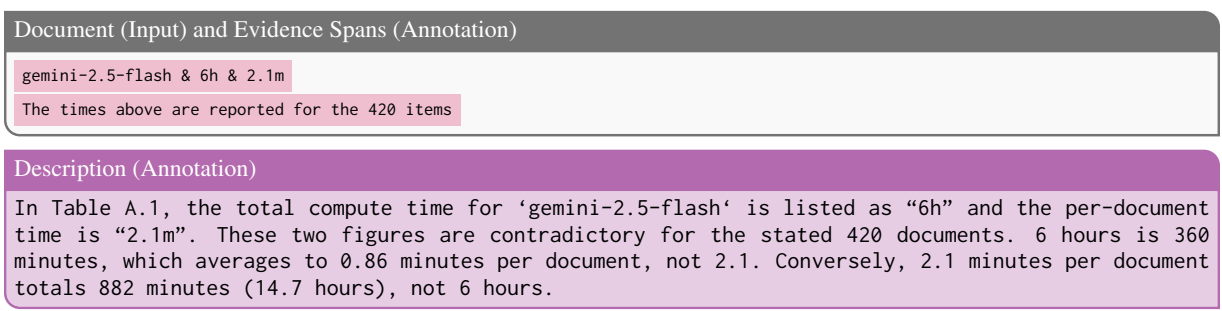

\small
\begin{tcolorbox}[
        enhanced jigsaw, 
        colback=gray!5!white, 
        colframe=gray!90!black, 
        boxrule=1pt, 
        arc=5pt, 
        sharp corners=downhill, 
        left=0pt, right=0pt, top=0pt, bottom=0pt,
        fontupper=\scriptsize\ttfamily,
        title=Document (Input) and Evidence Spans (Annotation)
    ]
\SPANEX{gemini-2.5-flash & 6h & 2.1m}\\
\SPANEX{The times above are reported for the 420 items}
\end{tcolorbox}
 \begin{tcolorbox}[
        enhanced jigsaw, 
        colback=PURPLE!25, 
        colframe=PURPLE!75, 
        boxrule=1pt, 
        arc=5pt, 
        sharp corners=downhill, 
                fontupper=\ttfamily,
        left=0pt, right=0pt, top=0pt, bottom=0pt,
        title=Description  (Annotation)
    ]
In Table A.1, the total compute time for `gemini-2.5-flash` is listed as ``6h'' and the per-document time is ``2.1m''. These two figures are contradictory for the stated 420 documents. 6 hours is 360 minutes, which averages to 0.86 minutes per document, not 2.1. Conversely, 2.1 minutes per document totals 882 minutes (14.7 hours), not 6 hours.
\end{tcolorbox}

\caption{\textcolor{OliveGreen}{\textbf{Real}} inconsistency found by \model{gemini-2.5-pro} in an earlier draft.}
\label{app:fig:dataset:example:meta:unanticipated:GEMINI_2_06}
\end{figure*}

\begin{figure*}[ht!]
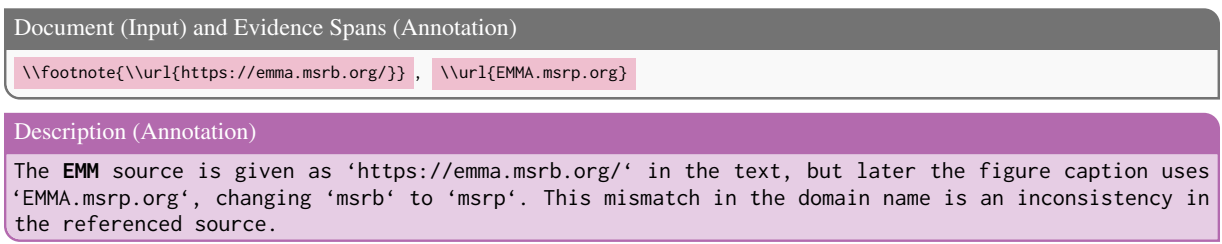

\small
\begin{tcolorbox}[
        enhanced jigsaw, 
        colback=gray!5!white, 
        colframe=gray!90!black, 
        boxrule=1pt, 
        arc=5pt, 
        sharp corners=downhill, 
        left=0pt, right=0pt, top=0pt, bottom=0pt,
        fontupper=\scriptsize\ttfamily,
        title=Document (Input) and Evidence Spans (Annotation)
    ]
\SPANEX{\\footnote{\\url{https://emma.msrb.org/}}}, \SPANEX{\\url{EMMA.msrp.org}}
\end{tcolorbox}
 \begin{tcolorbox}[
        enhanced jigsaw, 
        colback=PURPLE!25, 
        colframe=PURPLE!75, 
        boxrule=1pt, 
        arc=5pt, 
        sharp corners=downhill, 
                fontupper=\ttfamily,
        left=0pt, right=0pt, top=0pt, bottom=0pt,
        title=Description  (Annotation)
    ]
        The \USPF{} source is given as `https://emma.msrb.org/` in the text, but later the figure caption uses `EMMA.msrp.org`, changing `msrb` to `msrp`. This mismatch in the domain name is an inconsistency in the referenced source.
\end{tcolorbox}
\caption{\textcolor{OliveGreen}{\textbf{Real}} inconsistency found by \model{gpt-5} after the first round of fixes to model-identified inconsistencies.}
\label{app:fig:dataset:example:meta:unanticipated:GPT5_3_00}
\end{figure*}

\begin{figure*}[ht!]
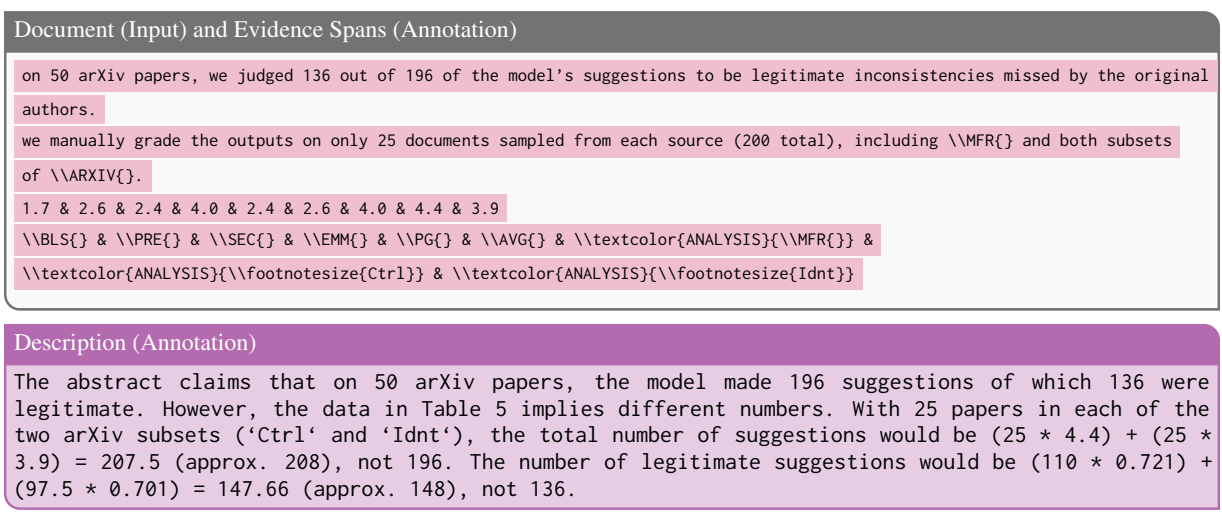

\small
\begin{tcolorbox}[
        enhanced jigsaw, 
        colback=gray!5!white, 
        colframe=gray!90!black, 
        boxrule=1pt, 
        arc=5pt, 
        sharp corners=downhill, 
        left=0pt, right=0pt, top=0pt, bottom=0pt,
        fontupper=\scriptsize\ttfamily,
        title=Document (Input) and Evidence Spans (Annotation)
    ]
\SPANEX{on 50 arXiv papers, we judged 136 out of 196 of the model's suggestions to be legitimate inconsistencies missed by the original}\\
\SPANEX{authors.}\\
\SPANEX{we manually grade the outputs on only 25 documents sampled from each source (200 total), including \\MFR{} and both subsets}\\
\SPANEX{of \\ARXIV{}.}\\
\SPANEX{1.7 & 2.6 & 2.4 & 4.0 & 2.4 & 2.6 & 4.0 & 4.4 & 3.9}\\
\SPANEX{\\BLS{} & \\PRE{} & \\SEC{} & \\EMM{} & \\PG{} & \\AVG{} & \\textcolor{ANALYSIS}{\\MFR{}} &}\\
\SPANEX{\\textcolor{ANALYSIS}{\\footnotesize{Ctrl}} & \\textcolor{ANALYSIS}{\\footnotesize{Idnt}}}\\
\end{tcolorbox}
 \begin{tcolorbox}[
        enhanced jigsaw, 
        colback=PURPLE!25, 
        colframe=PURPLE!75, 
        boxrule=1pt, 
        arc=5pt, 
        sharp corners=downhill, 
                fontupper=\ttfamily,
        left=0pt, right=0pt, top=0pt, bottom=0pt,
        title=Description  (Annotation)
    ]
The abstract claims that on 50 arXiv papers, the model made 196 suggestions of which 136 were legitimate. However, the data in Table 5 implies different numbers. With 25 papers in each of the two arXiv subsets (`Ctrl` and `Idnt`), the total number of suggestions would be (25 * 4.4) + (25 * 3.9) = 207.5 (approx. 208), not 196. The number of legitimate suggestions would be (110 * 0.721) + (97.5 * 0.701) = 147.66 (approx. 148), not 136.
\end{tcolorbox}
\caption{\textcolor{orange}{\textbf{Noteworthy}} comment made by \model{gemini-2.5-pro}. This apparent mismatch is intentional, as we exclude the 12 known inconsistencies that were found (\Cref{sec:analysis:hard}) from the 208 total predictions when describing the analysis in the abstract.}
\label{app:fig:dataset:example:meta:unanticipated:GEMINI_3_00}
\end{figure*}

\model{gpt-5} found three real issues (\Cref{app:fig:dataset:example:meta:unanticipated:GPT5_2_00,app:fig:dataset:example:meta:unanticipated:GPT5_2_01,app:fig:dataset:example:meta:unanticipated:GPT5_2_02}). One of those issues would have been found anyway because it was a placeholder for this analysis. The model did not find the inconsistency we intentionally inserted. \model{gemini-2.5-pro} made one incorrect suggestion, two helpful suggestions (\Cref{app:fig:dataset:example:meta:unanticipated:GEMINI_2_01,app:fig:dataset:example:meta:unanticipated:GEMINI_2_05}), and found four real issues (\Cref{app:fig:dataset:example:meta:unanticipated:GEMINI_2_02,app:fig:dataset:example:meta:unanticipated:GEMINI_2_03,app:fig:dataset:example:meta:unanticipated:GEMINI_2_04,app:fig:dataset:example:meta:unanticipated:GEMINI_2_06}). One of the real issues was the inconsistency we intentionally inserted. Another of the raised issues overlapped with one found by \model{gpt-5}. After fixing these inconsistencies in our draft and running the models again, \model{gpt-5} found one more issue (\Cref{app:fig:dataset:example:meta:unanticipated:GPT5_3_00}), and \model{gemini-2.5-pro} found an interesting apparent mismatch between the abstract and the analysis section (\Cref{app:fig:dataset:example:meta:unanticipated:GEMINI_3_00}), which arises because we do not fully describe the experiment in the abstract.

\begin{figure*}[ht!]
\small
\begin{tcolorbox}[
        enhanced jigsaw, 
        colback=gray!5!white, 
        colframe=gray!90!black, 
        boxrule=1pt, 
        arc=5pt, 
        sharp corners=downhill, 
        left=0pt, right=0pt, top=0pt, bottom=0pt,
        fontupper=\scriptsize\ttfamily,
        title=Document (Input) and Evidence Spans (Annotation)
    ]
\SPANEX{containing 375 test and 125 development problems.}\\
\SPANEX{In the test set 83 out of 420 documents (at most, depending on the model and tokenizer) had their text truncated.}
\end{tcolorbox}
 \begin{tcolorbox}[
        enhanced jigsaw, 
        colback=PURPLE!25, 
        colframe=PURPLE!75, 
        boxrule=1pt, 
        arc=5pt, 
        sharp corners=downhill, 
                fontupper=\ttfamily,
        left=0pt, right=0pt, top=0pt, bottom=0pt,
        title=Description  (Annotation)
    ]
      The document defines the test set as 375 items, but later refers to the ``test set'' as having 420 documents when discussing truncation, creating a contradiction about test set size.
\end{tcolorbox}

\caption{\textcolor{OliveGreen}{\textbf{Real}} inconsistency found by \model{gpt-5} in an earlier draft. The 420 value includes the \WLD{} data. }
\label{app:fig:dataset:example:meta:unanticipated:GPT5_2_00}
\end{figure*}

\begin{figure*}[ht!]
\small
\begin{tcolorbox}[
        enhanced jigsaw, 
        colback=gray!5!white, 
        colframe=gray!90!black, 
        boxrule=1pt, 
        arc=5pt, 
        sharp corners=downhill, 
        left=0pt, right=0pt, top=0pt, bottom=0pt,
        fontupper=\scriptsize\ttfamily,
        title=Document (Input) and Evidence Spans (Annotation)
    ]
\SPANEX{gpt-5-mini}\\
\SPANEX{\\model{gpt-5-nano}}
\end{tcolorbox}
 \begin{tcolorbox}[
        enhanced jigsaw, 
        colback=PURPLE!25, 
        colframe=PURPLE!75, 
        boxrule=1pt, 
        arc=5pt, 
        sharp corners=downhill, 
                fontupper=\ttfamily,
        left=0pt, right=0pt, top=0pt, bottom=0pt,
        title=Description  (Annotation)
    ]
The Methods list ``gpt-5-nano'' as a tested model, whereas results tables and later sections report ``gpt-5-mini'' creating an inconsistency in the model variant named as evaluated.
\end{tcolorbox}
\caption{\textcolor{OliveGreen}{\textbf{Real}} inconsistency found by \model{gpt-5} in an earlier draft.}
\label{app:fig:dataset:example:meta:unanticipated:GPT5_2_02}
\end{figure*}

\begin{figure*}[ht!]
\small
\begin{tcolorbox}[
        enhanced jigsaw, 
        colback=gray!5!white, 
        colframe=gray!90!black, 
        boxrule=1pt, 
        arc=5pt, 
        sharp corners=downhill, 
        left=0pt, right=0pt, top=0pt, bottom=0pt,
        fontupper=\scriptsize\ttfamily,
        title=Document (Input) and Evidence Spans (Annotation)
    ]
\SPANEX{the best-performing model (\\model{gpt-5}) recovered 64\\\% of the inserted inconsistencies.}\\
\SPANEX{even the best models miss almost half of the inconsistencies in \\datasetname{}, demonstrating that inconsistency detection} \SPANEX{is still a challenging task.}\\
\end{tcolorbox}
 \begin{tcolorbox}[
        enhanced jigsaw, 
        colback=PURPLE!25, 
        colframe=PURPLE!75, 
        boxrule=1pt, 
        arc=5pt, 
        sharp corners=downhill, 
                fontupper=\ttfamily,
        left=0pt, right=0pt, top=0pt, bottom=0pt,
        title=Description  (Annotation)
    ]
The abstract states that the best model, gpt-5, recovered 64\% of inconsistencies, but concludes by stating that even the best models miss ``almost half'' (50\%) of them. A 64\% recovery rate means a 36\% miss rate, which is substantially different from ``almost half''. This mischaracterization is repeated in the Discussion section.
\end{tcolorbox}

\caption{\textcolor{orange}{\textbf{Helpful}} comment made by \model{gemini-2.5-pro} on an earlier draft. We would argue that ``almost half'' is a fair characterization.}
\label{app:fig:dataset:example:meta:unanticipated:GEMINI_2_01}
\end{figure*}

\begin{figure*}[ht!]
\small
\begin{tcolorbox}[
        enhanced jigsaw, 
        colback=gray!5!white, 
        colframe=gray!90!black, 
        boxrule=1pt, 
        arc=5pt, 
        sharp corners=downhill, 
        left=0pt, right=0pt, top=0pt, bottom=0pt,
        fontupper=\scriptsize\ttfamily,
        title=Document (Input) and Evidence Spans (Annotation)
    ]
\SPANEX{we did so on a late version of our draft and found \\textcolor{orange}{three} inconsistencies, which are documented in}\\
\SPANEX{\\Cref{app:sec:paper:inconsistencies}.}\\
\SPANEX{\\section{Inconsistencies Found in Earlier Versions of This Work}\\label{app:sec:paper:inconsistencies}}
\end{tcolorbox}
 \begin{tcolorbox}[
        enhanced jigsaw, 
        colback=PURPLE!25, 
        colframe=PURPLE!75, 
        boxrule=1pt, 
        arc=5pt, 
        sharp corners=downhill, 
                fontupper=\ttfamily,
        left=0pt, right=0pt, top=0pt, bottom=0pt,
        title=Description  (Annotation)
    ]
The Discussion section states that three inconsistencies found in the paper's own draft are documented in Appendix I. However, Appendix I, while present, is empty and does not contain the promised documentation.
\end{tcolorbox}

\caption{\textcolor{OliveGreen}{\textbf{Real}} inconsistency found by \model{gemini-2.5-pro} in an earlier draft. \model{gpt-5} found this same issue. We note that this would have been fixed anyway, as it was a placeholder value for this analysis. \textit{We do not count this toward the number of novel inconsistencies found.}}
\label{app:fig:dataset:example:meta:unanticipated:GEMINI_2_04}
\end{figure*}

\clearpage
\section{Examples of Undiscovered Inconsistencies in \ARXIV{} and \PG{}}\label{app:sec:undiscovered}
See \Cref{app:fig:dataset:example:meta:unanticipated:ROW_73,app:fig:dataset:example:meta:unanticipated:ROW_86,app:fig:dataset:example:meta:unanticipated:ROW_119,app:fig:dataset:example:meta:unanticipated:ROW_228-2}. To improve readability, we made minor formatting changes to the text and removed long sections of text, replaced by ``[...]''.

\clearpage
\begin{figure*}[ht!]
\small
\begin{tcolorbox}[
        enhanced jigsaw, 
        colback=gray!5!white, 
        colframe=gray!90!black, 
        boxrule=1pt, 
        arc=5pt, 
        sharp corners=downhill, 
        left=0pt, right=0pt, top=0pt, bottom=0pt,
        fontupper=\scriptsize\ttfamily,
        title=Document (Input) and Evidence Spans (Annotation)
    ]
\begin{tabular}{|l|ccc|ccc|ccc|}
			\hline
                \multicolumn{1}{|c|}{Model}
								& \multicolumn{3}{c|}{14LAP}                       & \multicolumn{3}{c|}{\SPANEX{14RES}}               & \multicolumn{3}{c|}{15RES}\\
								& P              	& R              	& F1             	& P      	& R              	& F1             & P              & R              & F1           \\ \hline

			\SPANPLAIN{CMLA+~\cite{wang2017multi}}          	    & 30.10          	& 36.90          & 33.20          	& 39.20  & 39.80          & 37.00          & 41.30          & 42.10          & 41.70 \\
			\SPANPLAIN{RINANTE+ ~\cite{peng2020knowing}}       		& 21.70          	& 18.70          & 20.10          	& 29.90  & 30.10          & 30.00          & 25.70          & 22.30          & 23.90\\
                \SPANEX{WhatHowWhy~\cite{peng2020knowing}}             & 37.38             & 50.38          & 42.87            & \SPANEX{43.24}  & \SPANEX{48.07}          & \SPANEX{63.66}          & 51.46          & 57.51          & 52.32\\
 \end{tabular}
 
\end{tcolorbox}
 \begin{tcolorbox}[
        enhanced jigsaw, 
        colback=PURPLE!25, 
        colframe=PURPLE!75, 
        boxrule=1pt, 
        arc=5pt, 
        sharp corners=downhill, 
                fontupper=\ttfamily,
        left=0pt, right=0pt, top=0pt, bottom=0pt,
        title=Description  (Annotation)
    ]
In the 14RES column, the F1 score (63.66) exceeds both precision (43.24) and recall (48.07), which is impossible since F1 must lie between P and R.
\end{tcolorbox}

\caption{\textcolor{OliveGreen}{\textbf{Real}} inconsistency found by \model{gpt-5} from \ARXIV{} with type \textit{Numeric}.}
\label{app:fig:dataset:example:meta:unanticipated:ROW_119}
\end{figure*}

\begin{figure*}[ht!]
\small
\begin{tcolorbox}[
        enhanced jigsaw, 
        colback=gray!5!white, 
        colframe=gray!90!black, 
        boxrule=1pt, 
        arc=5pt, 
        sharp corners=downhill, 
        left=0pt, right=0pt, top=0pt, bottom=0pt,
        fontupper=\scriptsize\ttfamily,
        title=Document (Input) and Evidence Spans (Annotation)
    ]
\SPANPLAIN{[...]}\\
\SPANPLAIN{\begin{figure}[t]}\\
\SPANPLAIN{\centering}
\SPANPLAIN{\includegraphics[width=\linewidth]{figs/fig4.pdf}}\\
\SPANEX{\caption{Speedup comparison.}}\\
\SPANPLAIN{[...]}\\
\SPANPLAIN{Our hypothesis is that larger training datasets will lead to improved model performance, as indicated by reductions in cross-entropy loss.} \SPANEX{Figure~\ref{fig:fig4} depicts these scaling curves for the different model sizes.}\\

\end{tcolorbox}
 \begin{tcolorbox}[
        enhanced jigsaw, 
        colback=PURPLE!25, 
        colframe=PURPLE!75, 
        boxrule=1pt, 
        arc=5pt, 
        sharp corners=downhill, 
                fontupper=\ttfamily,
        left=0pt, right=0pt, top=0pt, bottom=0pt,
        title=Description  (Annotation)
    ]
Figure 4 is introduced earlier as a speedup comparison, but later it is referenced as showing scaling curves. This misreference creates an inconsistency between the figures stated content (speedup) and the later claim (scaling), making the pointer to Figure 4 for scaling incorrect.
\end{tcolorbox}

\caption{\textcolor{OliveGreen}{\textbf{Real}} inconsistency found by \model{gpt-5} from \ARXIV{} with type \textit{Structural}.}
\label{app:fig:dataset:example:meta:unanticipated:ROW_86}
\end{figure*}

\begin{figure*}[ht!]
\small
\begin{tcolorbox}[
        enhanced jigsaw, 
        colback=gray!5!white, 
        colframe=gray!90!black, 
        boxrule=1pt, 
        arc=5pt, 
        sharp corners=downhill, 
        left=0pt, right=0pt, top=0pt, bottom=0pt,
        fontupper=\scriptsize\ttfamily,
        title=Document (Input) and Evidence Spans (Annotation)
    ]
\SPANEX{These include [...]
HellaSwag\~\cite{zellers2019hellaswag}, [...] and WinoGrande~\cite{sakaguchi2019winogrande}.}

\SPANEX{We demonstrate the results of this evaluation suite in~\Cref{tab:downstream}.} 

$[\dots]$\\


\SPANPLAIN{\caption{Zero-shot benchmark accuracy on downstream datasets.} \label{tab:downstream}}\\
\begin{tabular}{llcccccccc}
\toprule
Size & Model & Train. toks. & ARC-e & ARC-c & BoolQ & PIQA & OBQA & WinoG. & Avg. \\
\end{tabular}

\end{tcolorbox}
 \begin{tcolorbox}[
        enhanced jigsaw, 
        colback=PURPLE!25, 
        colframe=PURPLE!75, 
        boxrule=1pt, 
        arc=5pt, 
        sharp corners=downhill, 
                fontupper=\ttfamily,
        left=0pt, right=0pt, top=0pt, bottom=0pt,
        title=Description  (Annotation)
    ]
The text claims HellaSwag is part of the reported evaluation suite and that the results of this suite are shown in the referenced table. However, the table that follows does not report HellaSwag results, so an expected metric is missing from the presented results.
\end{tcolorbox}

\caption{\textcolor{OliveGreen}{\textbf{Real}} inconsistency found by \model{gpt-5} from \ARXIV{} with type \textit{Structural}.}
\label{app:fig:dataset:example:meta:unanticipated:ROW_73}
\end{figure*}

\begin{figure*}[ht!]
\small
\begin{tcolorbox}[
        enhanced jigsaw, 
        colback=gray!5!white, 
        colframe=gray!90!black, 
        boxrule=1pt, 
        arc=5pt, 
        sharp corners=downhill, 
        left=0pt, right=0pt, top=0pt, bottom=0pt,
        fontupper=\scriptsize\ttfamily,
        title=Document (Input) and Evidence Spans (Annotation)
    ]
\begin{tabular}{lrr}
     Flour, per bag of 24.5 pounds       &     1.721   & 1.790 \\
\SPANEX{Corn meal, per pound} &          \SPANEX{.055}&  \SPANEX{.067}\\
              Potatoes, per peck        &     .579 & .645 \\
            Sugar, per pound          &     .122 & .110 \\
\end{tabular}

\SPANPLAIN{The factors as worked out in Table 1 are ratios between yearly average retail prices and yearly average Dun's index numbers. Even retail
prices, however, have some seasonal swing.}
[...] \SPANEX{In the main, however, Table 2 is fairly accurate as it stands. It will be noted that with the exception of hog products, wheat}
\SPANEX{and potatoes, retail prices in September of 1919 tended to be lower than their normal ratio to Dun's index number.}
\end{tcolorbox}
 \begin{tcolorbox}[
        enhanced jigsaw, 
        colback=PURPLE!25, 
        colframe=PURPLE!75, 
        boxrule=1pt, 
        arc=5pt, 
        sharp corners=downhill, 
                fontupper=\ttfamily,
        left=0pt, right=0pt, top=0pt, bottom=0pt,
        title=Description  (Annotation)
    ]
The text claims only hog products, wheat, and potatoes were above their index prices, but corn meal’s actual price (0.067) exceeded its index price (0.055). This contradicts the stated set of exceptions.
\end{tcolorbox}

\caption{\textcolor{OliveGreen}{\textbf{Real}} inconsistency found by \model{gpt-5} from \PG{} with type \textit{Numeric}.}
\label{app:fig:dataset:example:meta:unanticipated:ROW_228-2}
\end{figure*}

\clearpage
\section{Open-Source and Links}\label{app:sec:open}

The dataset and terms of usage are available at: \url{https://drive.google.com/drive/folders/1x4IwHH2g_iwEtOrwYu_QLJy5tbjyRzD6?usp=drive_link}. Note that documents in the dataset have been modified to include inconsistencies. We will provide a HuggingFace version after the paper is reviewed. Code will also be made available after review. The guides used for annotation and dataset creation are available at: \url{https://drive.google.com/drive/folders/18qOjCICcMPL_NSzJjirpTxx6G4ecwlfB?usp=drive_link}.

\section{Reproducing this Work and Implementation Details}\label{app:sec:reproducing}
\subsection{Language Model Software and Hardware}
For the open-source models we use huggingface transformer\footnote{\url{https://huggingface.co/docs/transformers/index}} weights and VLLM \citep{kwon2023efficient} for inference. For hardware, we use a \texttt{P4de-24, 8 x nvidia A100} AWS instance. For the closed-source models, we use the \texttt{LiteLLM}\footnote{\url{https://github.com/BerriAI/litellm}} interface to interact with their endpoints. For the closed-source models, all items were run synchronously in an online manner. 

Runtimes are presented in \Cref{app:tbl:computetime}. The times are reported for the 420 items in the \datasetname{} test set and \WLD{}. Open-source models were run on a \texttt{P4de-24, 8 x nvidia A100} AWS instance using VLLM. The closed-source models were run by the respective model providers' hardware via API. Because these values are based on single runs, we expect there to be some variance over compute times, especially for the closed-source models.

For each model's outputs on \datasetname{}, the grader (\model{gpt-4.1}) completed in 5m to 30m. Larger models tended to output more inconsistencies and thus took longer to grade because each answer was graded independently.

\begin{table}[ht!]
\small
    \centering
\begin{tabular}{lrr}
\toprule
& \multicolumn{2}{c}{Time}\\
Model & Total & Per Doc \\
\midrule
gemma-3-12b-it & 20m & 3s  \\
gemma-3-27b-it & 30m & 5s \\
gpt-oss-20b & 1h 5m & 10s \\
gpt-oss-120b & 1h 5m & 10s \\
\midrule
sonnet-v4 & 10h & 1.5m \\
gpt-5-mini & 10h& 1.5m \\
gpt-5 & 10h & 1.5m \\
o3-mini & 3h & 0.4m \\
o3  & 9h & 1.5m \\
gemini-2.5-flash & 6h &  0.9m \\
gemini-2.5-pro & 15h &  2.1m \\
\bottomrule
\end{tabular}
\caption{Compute time per document across \datasetname{} (and \WLD{}). All times are rounded up for legibility.}
\label{app:tbl:computetime}
\end{table}

\subsection{Handling Span Generation}\label{app:sec:span}
\citet{kasner2025largelanguagemodelsspan} discuss options for span generation. Our prompts (and grading code) instruct the models to present the answer in a tagged format: 
\begin{lstlisting}[language=Python, escapechar=!, backgroundcolor=\color{white}]
<answer>
    <evidence>!$\e_1$!</evidence>
    !$\dots$!
    <evidence>!$\e_n$!</evidence>
    <description>!$\T$!</description>
</answer>
\end{lstlisting}
We follow the recommendation of \citet{kasner2025largelanguagemodelsspan} to ``[list] textual content of the spans'', and expect the generated evidence spans are to appear verbatim within the original document.\footnote{Note that the textual contents of many evidence spans are likely to occur multiple times in the documents in \datasetname{}, especially for evidence such as row/column labels or table cell values, so there is ambiguity in which particular mention of the evidence text is the intended span in the document. Additionally, because the text may appear multiple times, text matching heuristics to identify the particular mention will not be able to correctly disambiguate all mentions. However, the approaches of XML-tagging the entire document or having the model generate span offsets have more severe drawbacks \cite{kasner2025largelanguagemodelsspan}, and so we do not use them.} In pilot testing, we found that models were generally successful at producing answers in this format, especially compared to other options (such as JSON and Markdown). For responses indicating no inconsistency was present, we specify that the response should be:
\begin{lstlisting}[language=Python, escapechar=!, backgroundcolor=\color{white}]
<answer></answer>
\end{lstlisting}

\section{Evidence Metric}\label{app:sec:evidence_metric}
\begin{figure*}[]
\centering
\includegraphics[width=0.95 \linewidth]{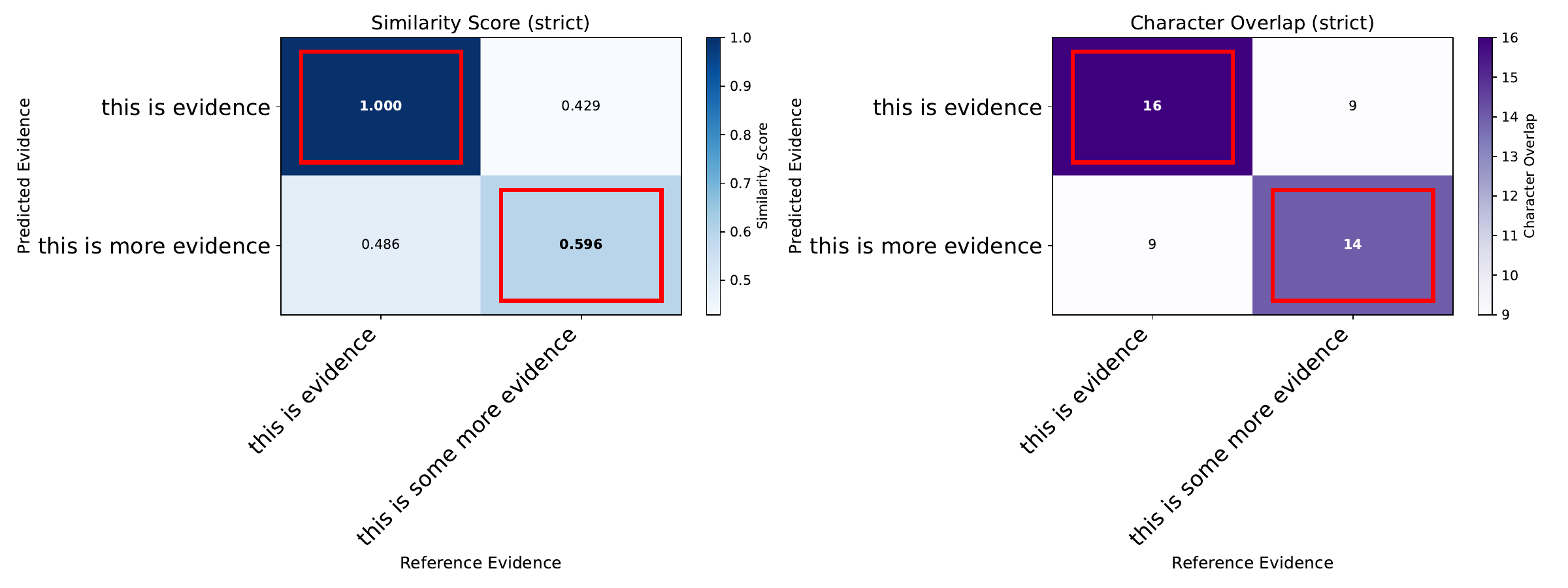}
\caption{Similarity scores and character overlaps computed for the strict evidence metric with predicted evidence $\hat{\E} = \{\textrm{``this is evidence'', ``this is more evidence''}\}$ and reference evidence $\E = \{\textrm{``this is evidence'', ``this is some more evidence''}\}$. Red outlines indicate the matched evidence determined via the similarity scores and superimposed on the character overlap matrix. The score for this example is then computed from the total amount of character overlap in the matched evidence, the total length of the predicted evidence, and the total length of the reference evidence.}
\label{fig:evidence_strict_illustration}
\end{figure*}

\begin{figure*}[]
\centering
\includegraphics[width=0.95 \linewidth]{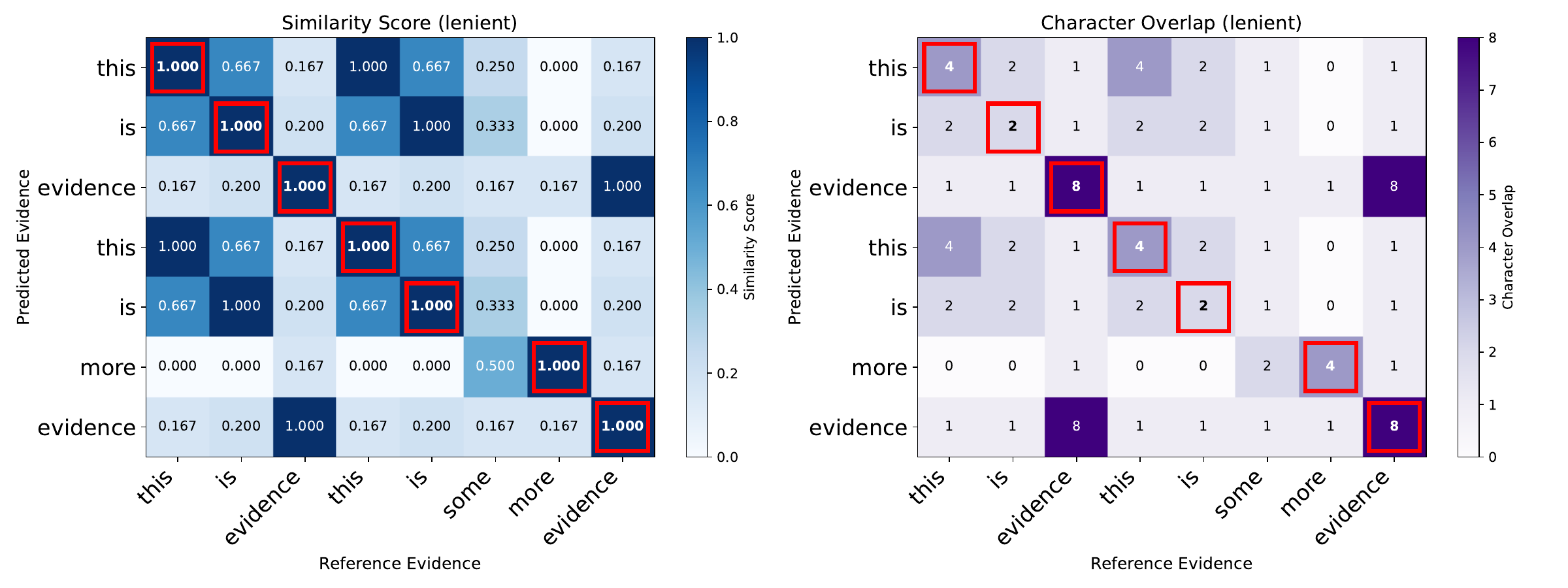}
\caption{Similarity scores and character overlaps computed for the lenient evidence metric with predicted evidence $\hat{\E} = \{\textrm{``this is evidence'', ``this is more evidence''}\}$ and reference evidence $\E = \{\textrm{``this is evidence'', ``this is some more evidence''}\}$. Red outlines indicate the matched evidence determined via the similarity scores and superimposed on the character overlap matrix. The score for this example is then computed from the total amount of character overlap in the matched evidence, the total length of the predicted evidence, and the total length of the reference evidence.}
\label{fig:evidence_lenient_illustration}
\end{figure*}

We support two versions of the lexical evidence metric: one strict and one lenient, each making different decisions about certain conditions on the predictions. Only the strict version is reported. For a concrete illustration of how the computation of the strict and lenient metrics differ see \Cref{fig:evidence_lenient_illustration,fig:evidence_strict_illustration}, and for pseudocode see \Cref{app:alg}.

The first decision point is whether the metric should be aware of span boundaries. The strict metric considers each piece of evidence \ee~as a separate span, while the lenient metric considers the totality of the evidence \ee~as a sequence of words.\footnote{For example, the strict metric views $\E = \{\textrm{``this is evidence'', ``this is more evidence''}\}$ as two spans, and the lenient metric views it as $\E = \{\textrm{``this'', ``is'', ``evidence'', ``this'', ``is'', ``more'', ``evidence''}\}$.} The lenient approach allows for disagreements on segmentation that do not affect the overall information conveyed, such as whether a piece of text is one long span or two shorter spans.

The second decision point is whether the amount of string overlap (and ultimately string similarity, see below) between predicted and reference evidence text should be based on their longest common substring (LCStr) or longest common subsequence (LCSeq).\footnote{The longest common substring of $x$ and $y$ is the longest \textit{contiguous} string that appears both in $x$ and in $y$. The longest common subsequence removes the contiguity constraint, allowing other text to be interspersed within the common subsequence.} LCStr (strict) more harshly penalizes predictions that differ from the reference even by one character, whereas LCSeq (lenient) allows for small differences (due to miscopying or using a slightly different piece of evidence than the reference does).

The third decision point is whether the predicted evidence should appear verbatim in the document. The strict approach requires that the predicted evidence occurs in the document and marks any predicted evidence that does not appear in the document as ineligible for matching to reference evidence. The lenient approach allows for mistakes in copying the evidence text.

We evaluate the predicted evidence against the reference evidence using a bipartite matching approach (discussed next), which aligns the predicted and reference evidence text. From the matching, we calculate the true positive count as the total amount of string overlap in the matching pairs of text, total predicted count as the amount of evidence text predicted by the model, and the total reference count as the amount of evidence text given in the reference answer.\footnote{All quantities are computed at the character level.}

To obtain the matching between predicted evidence and reference evidence, we first compute similarity scores between each predicted element and reference element.\footnote{For the strict metric, each element is a span. For the lenient metric, each element is a word, since we have removed the span boundaries from the computation.} The score is based on the string overlap between the two elements. Specifically, the similarity score is the F1 based on string overlap between the predicted element $x$ and reference element $y$, where $w$ is the sequence of overlapping characters (computed by LCStr or LCSeq), precision is $|w|/|x|$, recall is $|w|/|y|$, and F1 is the harmonic mean of precision and recall. The matching is then obtained from the similarity scores with a linear sum assignment algorithm to globally maximize the total score among matched predicted elements and reference elements. 

The total amount of character overlap among the matched elements is the true positive count achieved by the model on the example, which is then normalized by the character length of all the predicted evidence text to get precision or by the character length of all the reference evidence text to get recall. Intuitively, the precision is the proportion of predicted evidence text that appears in the reference evidence, and the recall is the proportion of reference evidence text that is covered by the predicted evidence (both conditioned on the alignment between predicted and reference evidence).\footnote{Certain cases can be scored automatically instead, without considering the content of the predicted evidence, such as when the model response is not well-formed, when the predicted and reference evidence are both empty (such as for consistent documents), and when either the reference or the predicted answer is empty and the other is not.}

To determine the ability of the evidence metrics to distinguish correct from incorrect responses, we compared their scores against manual judgments of evidence quality. We manually graded 202 responses with binary labels for whether the predicted evidence was similar enough to the reference evidence to correctly identify the inconsistency. The 202 responses were predictions on 101 internal development examples from two models, \texttt{Qwen3-30B-A3B-Instruct-2507} and \texttt{openai gpt-oss-120b}. Of the 202 responses, 61 were auto-gradable, and the remaining 141 were scored by the evidence metric.

\begin{figure}[]
\centering
\includegraphics[width=0.95 \linewidth]{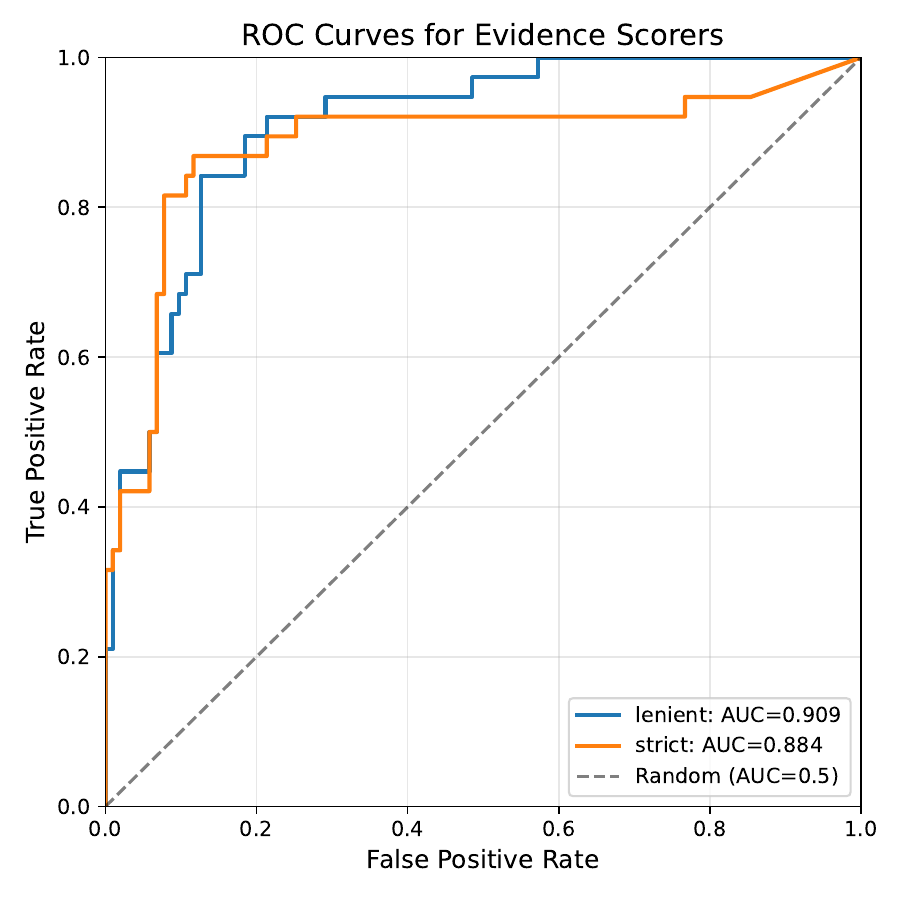}
\caption{ROC curves for evidence scorers based on 141 non-autograded examples from internal development data. Both the strict and lenient metrics can distinguish correct from incorrect evidence sets.}
\label{fig:evidence_roc_nonautograded}
\end{figure}

One could classify correct and incorrect evidence predictions by imposing a threshold on the scalar metric scores. \Cref{fig:evidence_roc_nonautograded} shows ROC curves that consider all such thresholds as well as the areas under the curve (AUC) for the strict and lenient metrics, based on the 141 non-autograded examples. Both metrics achieve high AUCs of 0.884 and 0.909, respectively, indicating that they are effective at distinguishing good predicted evidence from bad predicted evidence.

\algnewcommand{\LineComment}[1]{\State \(\triangleright\) #1}
\renewcommand{\algorithmicrequire}{\textbf{Input:}}
\renewcommand{\algorithmicensure}{\textbf{Output:}}
\begin{algorithm*}
\begin{algorithmic}[1]
\Function{OverlapScoreFn}{$x, y, \text{overlap\_fn}$}
  \If{$\text{overlap\_fn} = \text{LCStr}$}
      \State $\text{overlap\_block} \gets \text{longest contiguous matching block between $x$ and $y$}$
  \Else
      \State $\text{overlap\_blocks} \gets \text{all matching blocks (allowing gaps) between $x$ and $y$}$
  \EndIf
  \State $\text{overlap\_amount} \gets$ total characters in matching block(s)
  \State $\text{precision} \gets \text{overlap\_amount}~/~|x|$
  \State $\text{recall} \gets \text{overlap\_amount}~/~|y|$
  \State $\text{F1} \gets 2 \times \text{precision} \times \text{recall}~/~(\text{precision} + \text{recall})$
  \State \Return $\text{F1}, \text{overlap\_amount}$
\EndFunction
\item[]
\end{algorithmic}

\begin{algorithmic}[1]
\Require $\text{predicted\_spans}$, $\text{reference\_spans}$, $\text{mode} \in \{\text{strict}, \text{lenient}\}$, $\text{document\_text}$

\LineComment{Configure based on mode}
\If{$\text{mode} = \text{strict}$}
  \State $\text{predictions} \gets \text{predicted\_spans}$
  \State $\text{references} \gets \text{reference\_spans}$
  \State $\text{overlap\_fn} \gets \text{LCStr}$
  \State $\text{require\_verbatim} \gets \text{true}$
\Else
  \State $\text{predictions} \gets \text{flatten(predicted\_spans)}$
  \State $\text{references} \gets \text{flatten(reference\_spans)}$
  \State $\text{overlap\_fn} \gets \text{LCSeq}$
  \State $\text{require\_verbatim} \gets \text{false}$
\EndIf
\item[]
\LineComment{Build similarity matrix $S$ and character overlap matrix $C$}
\ForAll{$(\text{pred}_{i}, \text{ref}_{j}) \in$ predictions $\times$ references}
  \If{require\_verbatim and $\text{pred}_{i}$ $\notin \text{document\_text}$}
      \State $S[i,j] \gets -\infty$
      \State $C[i,j] \gets 0$
  \Else
      \State $\text{similarity\_score}, \text{overlap\_amount} \gets \Call{OverlapScoreFn}{\text{pred}_{i}, \text{ref}_{j}, \text{overlap\_fn}}$
      \State $S[i,j] \gets \text{similarity\_score}$
      \State $C[i,j] \gets \text{overlap\_amount}$
  \EndIf
\EndFor
\item[]
\LineComment{Get matched pairs of evidence}
\State $\text{matches} \gets \text{linear\_sum\_assignment}(S, \text{maximize}=\text{true})$
\State $\text{valid\_matches} \gets \text{filter(matches, lambda $i$, $j$: $S[i,j] \neq -\infty$)}$ \Comment{Remove spurious matches}
\item[]
\LineComment{Compute metrics}
\State $\text{total\_char\_overlap} \gets \sum_{(i,j) \in \text{valid\_matches}} C[i,j]$
\State $\text{total\_char\_predicted} \gets \sum_{p \in \text{predictions}} |p|$ \Comment{Includes non-verbatim and unmatched predictions}
\State $\text{total\_char\_reference} \gets \sum_{r \in \text{references}} |r|$ \Comment{Includes unmatched references}
\State $\text{precision} \gets \text{total\_char\_overlap}~/~\text{total\_char\_predicted}$
\State $\text{recall} \gets \text{total\_char\_overlap}~/~\text{total\_char\_reference}$
\State $\text{F1} \gets 2 \times \text{precision} \times \text{recall}~/~(\text{precision} + \text{recall})$

\Ensure precision, recall, F1
\end{algorithmic}
\caption{Evidence Metric Computation}
\label{app:alg}
\end{algorithm*}

\section{Description Metric}\label{app:sec:description_metric}

\begin{figure}[]
\centering
\includegraphics[width=0.95 \linewidth]{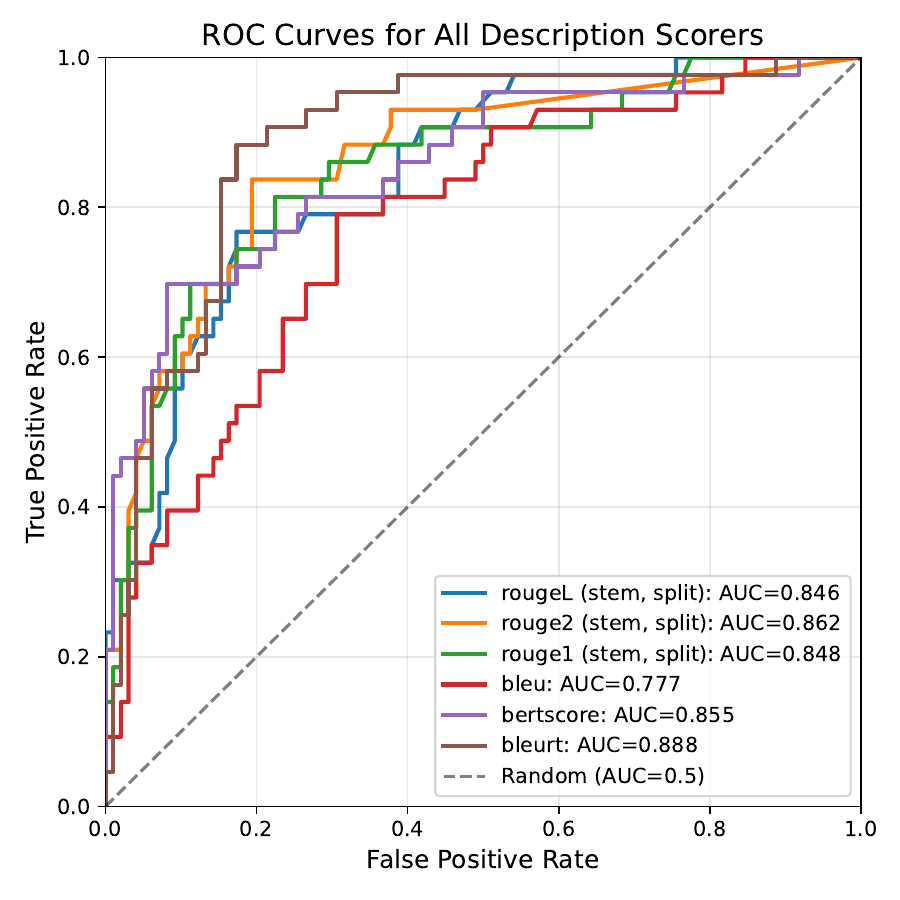}
\caption{ROC curves for all candidate description scorers based on 141 non-autograded examples from internal development data. All metrics can distinguish correct from incorrect descriptions, with BLEURT performing the best.}
\label{fig:description_roc_nonautograded}
\end{figure}

To determine the ability of the description metric to distinguish correct from incorrect responses, we use the same procedure as in \Cref{app:sec:evidence_metric}. We manually graded with binary labels the same 202 responses for whether the predicted description was similar enough to the reference description to correctly characterize the inconsistency. 61 of the responses were auto-gradable and 141 were scored by the description metric.

We considered a range of potential description metrics, some lexical and some neural, each of which provides a scalar score for the predicted description. \Cref{fig:description_roc_nonautograded} shows ROC curves for the description metrics we considered, based on the 141 non-autograded examples. Despite neural metrics like BERTScore \cite{Zhang2020BERTScore} exhibiting low sensitivity to numerical differences \cite{huang2025finnueexposingrisksusing}, the neural metric BLEURT obtains the highest AUC (0.888). ROUGE-2 obtains the second best AUC (0.862), and BLEU obtains the lowest (0.777). Based on these findings, we use BLEURT as the description metric.

\section{Dataset}\label{app:sec:dataset}
Documents from each data source in \datasetname{} were converted into Markdown documents using \url{https://extract.kensho.com/app} and \url{https://github.com/kensho-technologies/kenverters}, except for a subset of \BLS{} documents which were already in plain text. Most of the conversions were successful and high fidelity. Some \BLS{} documents required manual editing. For other document sources, when the conversion process failed we replaced the document with an alternative sample. Auxiliary files accompanying the \ARXIV{} documents were combined into a single plain LaTeX document with no additional markup. The auxillary \WLD{} documents  come from a diversely formatted set of PDF documents. Some of these documents were many hundreds of pages long. Because the formatting was non-uniform (and non-standard) we manually edited the the converted outputs, and because of the length, in some cases, we trimmed the documents to the relevant sections that contained the inconsistency identified by the expert. Note that these sections are still long, averaging 48k tokens.

\begin{figure*}[ht!]
\centering
\includegraphics[width=0.95 \linewidth]{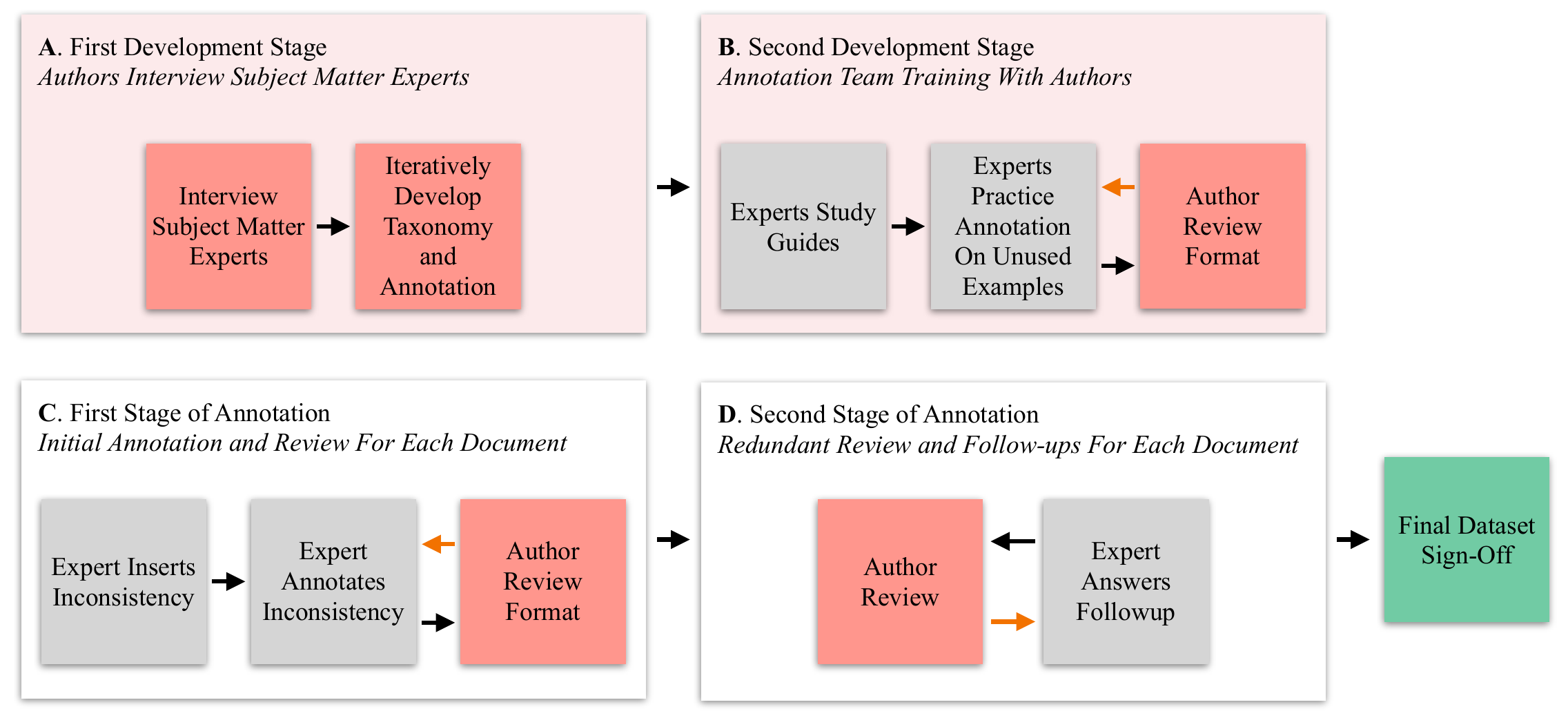}
\caption{Annotation and task quality workflow. The first two stages (A, B) are part of the dataset and task design. The next two stages (C, D) cover iterative annotation. First (\textbf{A}), we interviewed financial experts to help scope a useful taxonomy of inconsistency types. Next (\textbf{B}), our annotation team of financial experts reviewed the taxonomy of errors and annotation scheme. Then (\textbf{C}), the annotation team created and inserted inconsistencies into the documents. This was an iterative process of data review and validation which was followed by an additional review process (\textbf{D}). The complete annotation process required significant resources. On average, creating, annotating, and reviewing each inconsistency took two hours.}
\label{app:fig:annotation}
\end{figure*}

\Cref{app:fig:annotation} covers the flow of annotation and dataset creation.

\begin{figure*}[t]
    \centering
    \includegraphics[width=\linewidth]{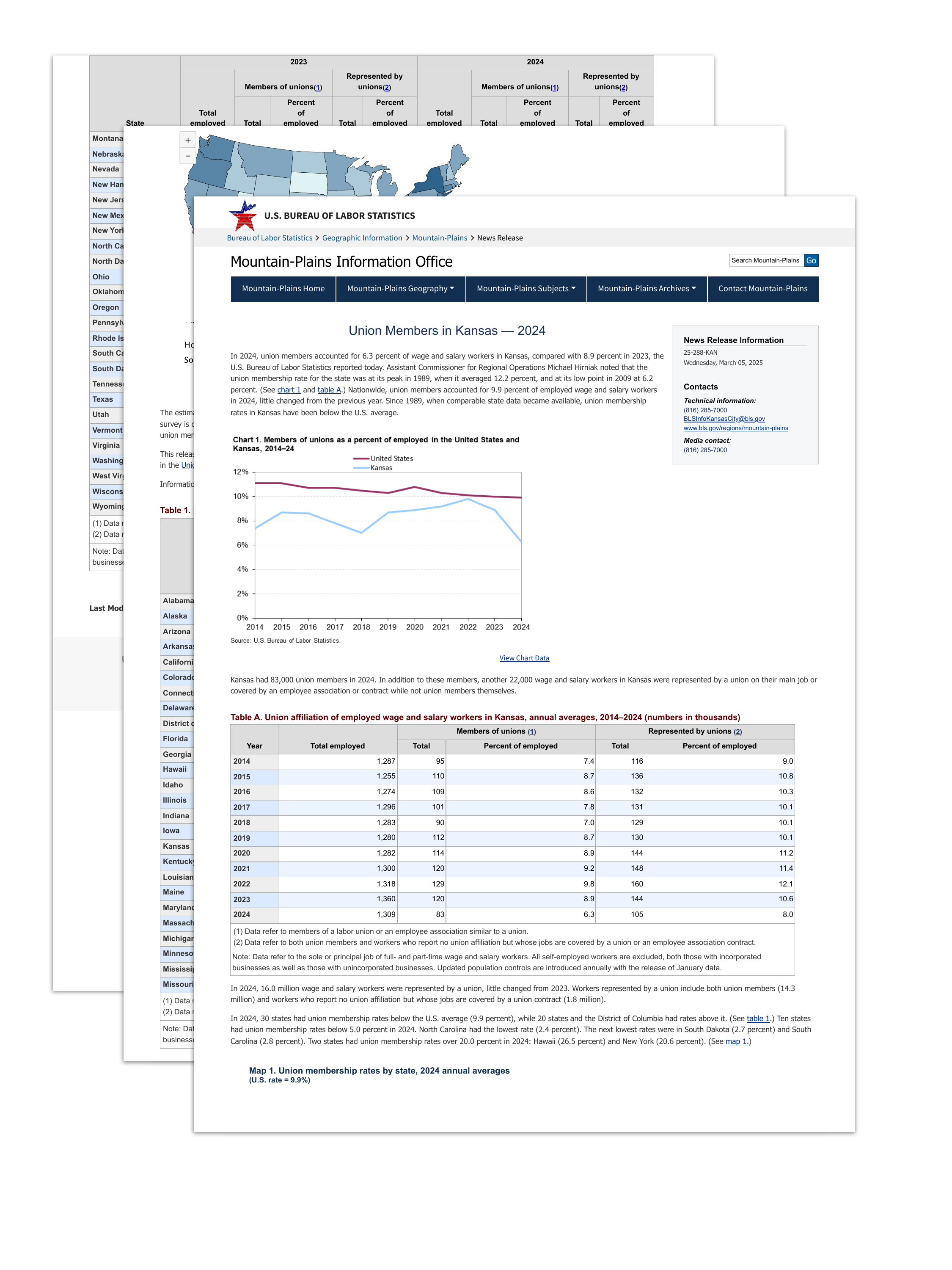}
    \caption{Pages from an example from \BLS{}. See original at \url{https://www.bls.gov/regions/mountain-plains/news-release/2025/unionmembership_kansas_20250305.htm}.} 
    \label{app:fig:dataset:example:BLS}
\end{figure*}

\begin{figure*}[t]
    \centering
    \includegraphics[width=\linewidth]{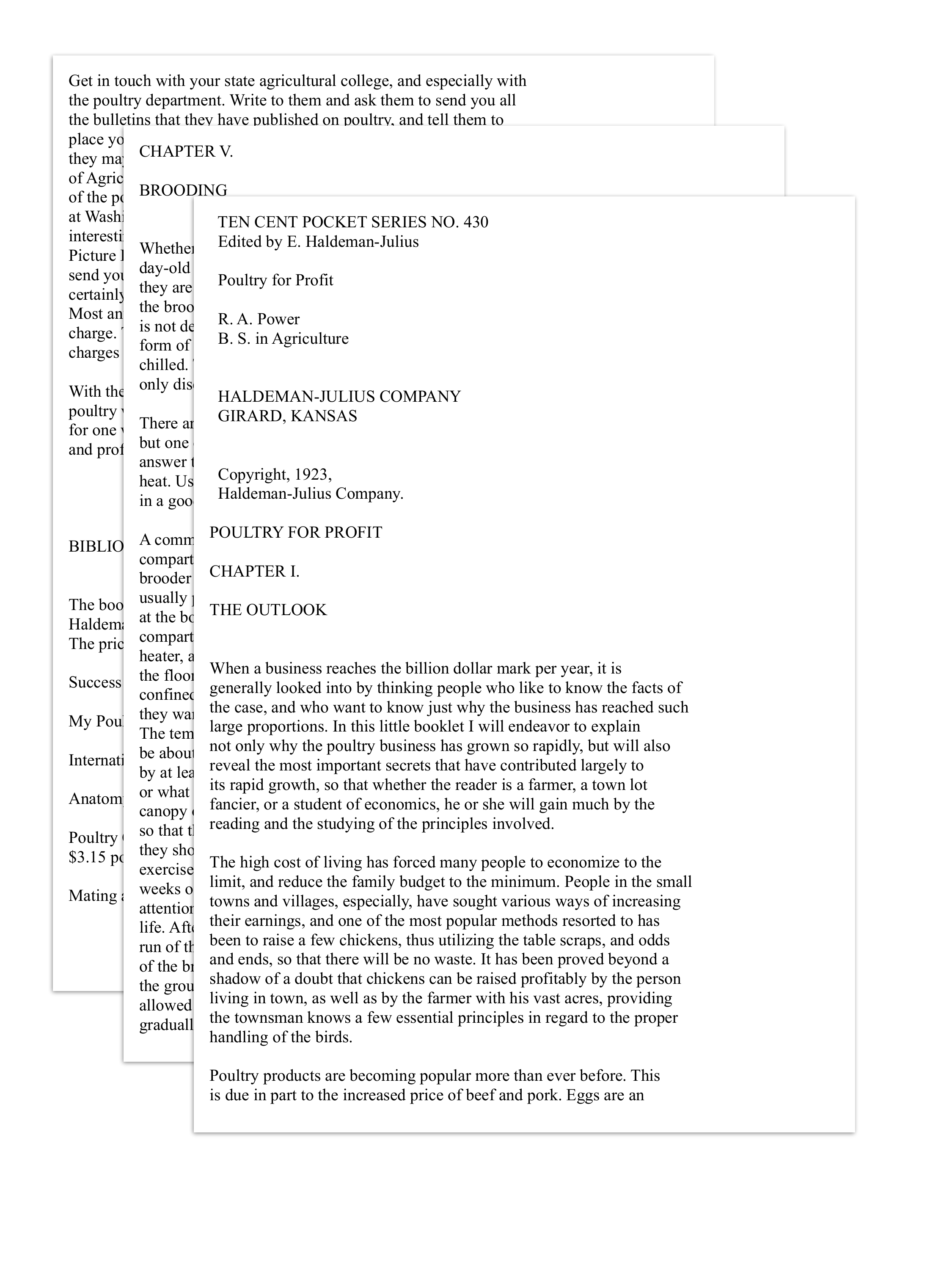}
    \caption{Pages from an example from \PG{}. See original at \url{https://www.gutenberg.org/ebooks/75419}.} 
    \label{app:fig:dataset:example:PG}
\end{figure*}

\begin{figure*}[t]
    \centering
    \includegraphics[width=\linewidth]{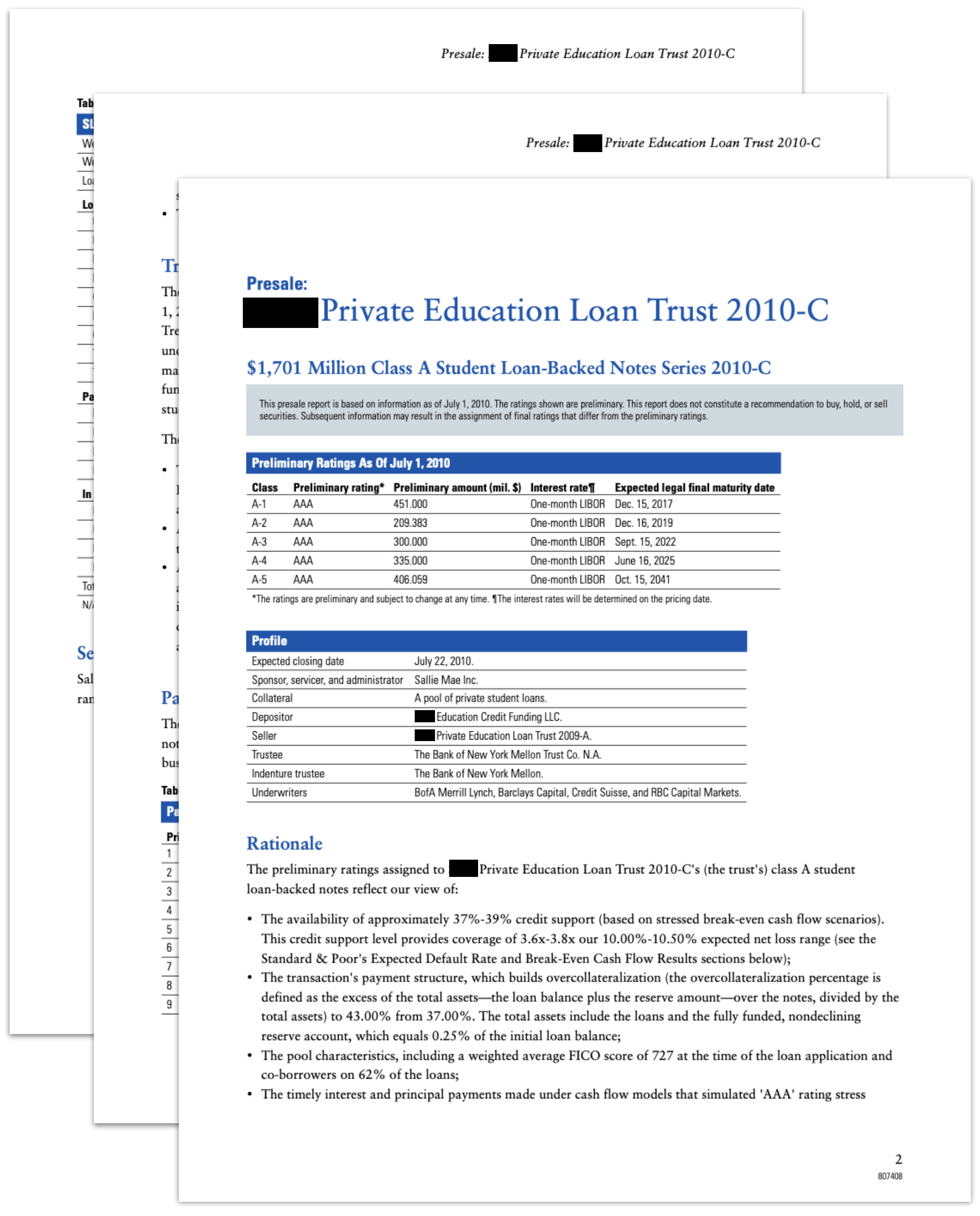}
    \caption{Pages from an example from \PRE{}. Documents were anonymized.} 
    \label{app:fig:dataset:example:PRE}
\end{figure*}

\begin{figure*}[t]
    \centering
    \includegraphics[width=\linewidth]{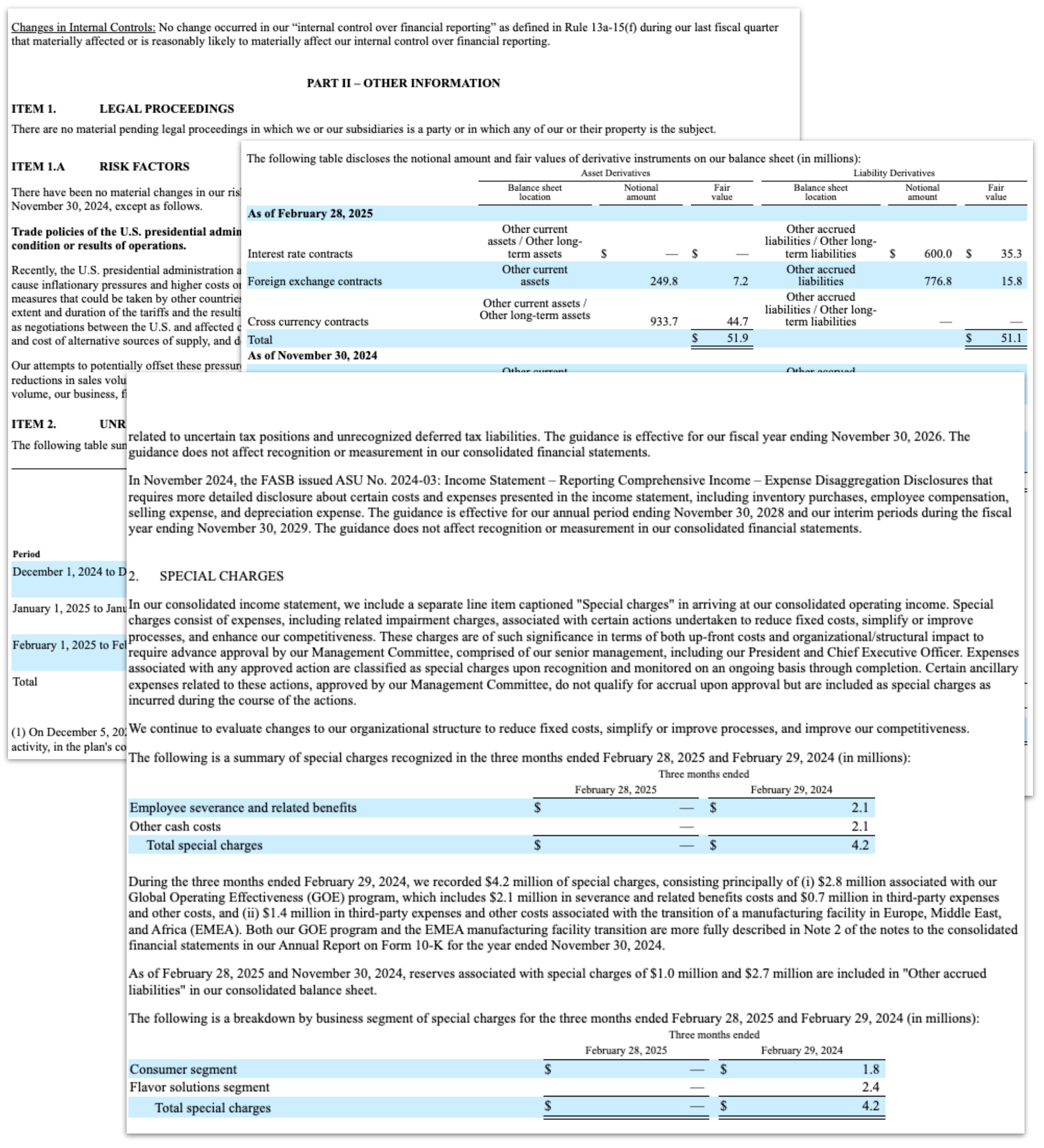}
    \caption{Pages from an example from \SEC{}. See original at \url{https://www.sec.gov/ix?doc=/Archives/edgar/data/63754/000006375425000024/mkc-20250228.htm}.}
    \label{app:fig:dataset:example:SEC}
\end{figure*}

\begin{figure*}[t]
    \centering
    \includegraphics[width=\linewidth]{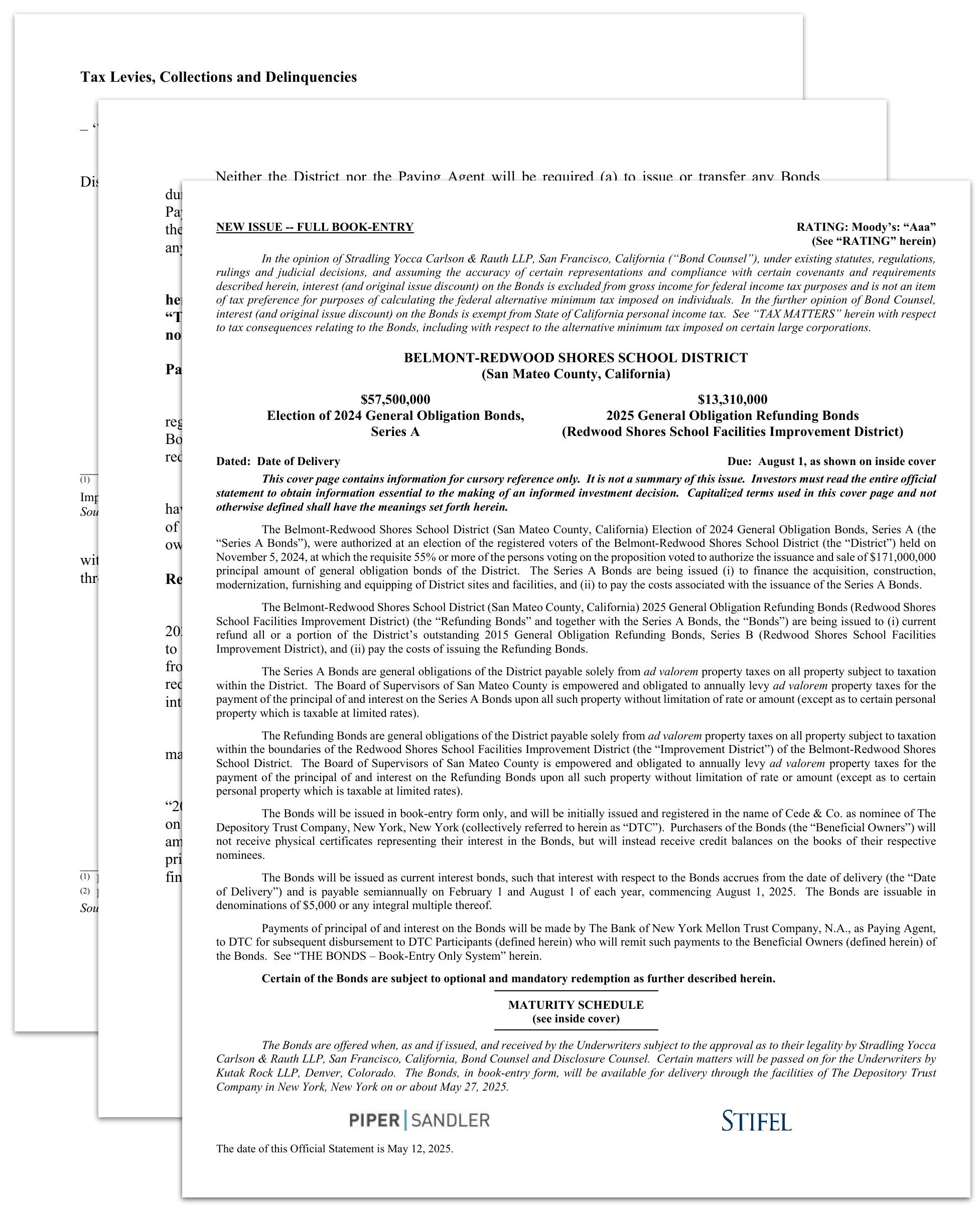}
    \caption{Pages from an example from \USPF{}. See original at \url{https://emma.msrb.org/P11854654-P11420157-P11863682.pdf}.}
    \label{app:fig:dataset:example:USPF}
\end{figure*}

\begin{figure*}[t]
    \centering
    \includegraphics[width=\linewidth]{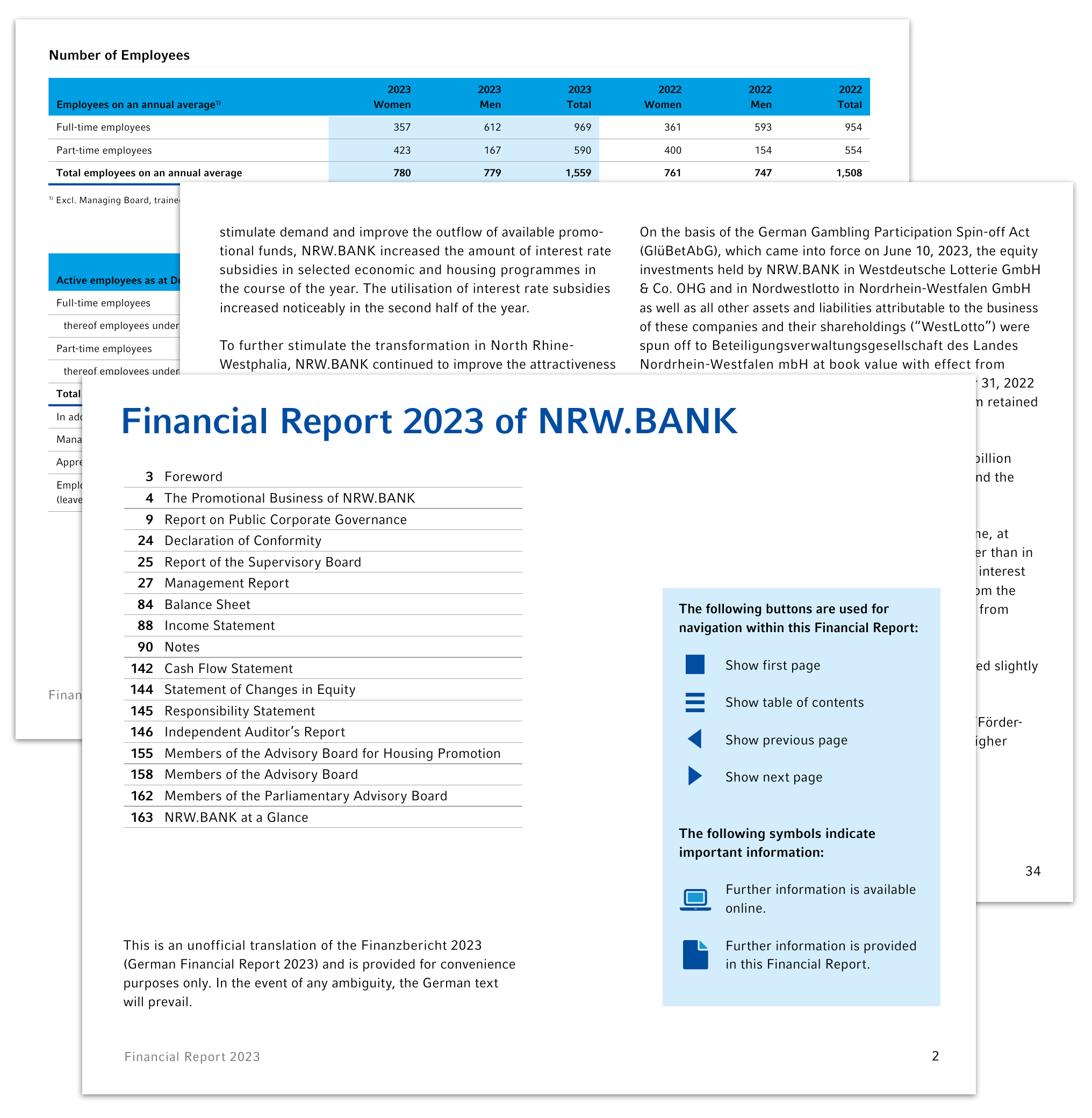}
    \caption{Pages from an example from \WLD{}. See original at \url{https://www.nrwbank.de/export/.galleries/downloads/Info-und-Service/Finanzberichte/nrwbank-financial-report-2023.pdf}.}
    \label{app:fig:dataset:example:WLD}
\end{figure*}

\Cref{app:fig:dataset:example:BLS,app:fig:dataset:example:PRE,app:fig:dataset:example:SEC,app:fig:dataset:example:USPF,app:fig:dataset:example:WLD,app:fig:dataset:example:PG} show examples of documents from the sources in \datasetname{} and \WLD{}.

Additional information about data from the sources is:

\paragraph{\BLS{}} 91 documents were released between January and April 2025. 9 documents were released between September and December 2025. Charts, which are out of scope for this work, were removed from the document. 

\paragraph{\PRE{}} These documents are not generally public and were published 2020 or before. Because we required explicit permission to access these documents, we expect most models are not trained on them. We removed the names of the primary stakeholders of these documents as a condition of use.

\paragraph{\SEC{}} We selected documents filed in January and February 2025.

\paragraph{\USPF{}} These reports show the financial health of local governments and provide ongoing updates to investors who bought their bonds. We selected documents released May 2025 and later.

\paragraph{\PG{}} We selected documents that were uploaded between September 2024 and June 2025 and which we predicted to not appear in pre-training corpora (see below). The items in the development set of this source in particular are not date-protected, with many uploaded in 2017 or before. We also removed the Project Gutenberg license and all references to Project Gutenberg from the documents per the requirement of their license.

To predict whether a candidate \PG{} document has been seen during model pre-training, we sample from the document 5 lines of more than 50 characters each and for each line obtain the number of occurrences it has within 3 corpora (OLMo-2-0325-32B-Instruct \cite{olmo20242olmo2furious}, Dolma-v1.7 \cite{soldaini-etal-2024-dolma}, and RedPajama \cite{weber2024redpajama}) using Infini-gram \cite{liu2024infinigram} and whether it appears in the Pile \cite{gao2020pile800gbdatasetdiverse} using Data Portraits \cite{marone2023dataportraits}.\footnote{Additional context before and after the sampled line was included in the input to Data Portraits to reduce false negatives.} Documents with a line that appears in the searched corpora are considered to be potentially seen by the models.
\\\\
\noindent\textcolor{ANALYSIS}{\textit{Analysis sources.}}
\paragraph{\textcolor{ANALYSIS}{\WLD{}}} These documents range in release date, and we expect it is possible that models were exposed to these documents.

\paragraph{\textcolor{ANALYSIS}{\ARXIV{}}} These documents range in release date, and it is likely that models were exposed to them. These documents often have multiple versions, particularly the subset with identified inconsistencies, for which versions with and without the inconsistency are available by construction. We do not know if models were trained on multiple versions of these papers.


\section{More Results}\label{app:sec:results}
\subsection{Development Data Results}
See \Cref{app:dev:tbl:dataset:dev}, \Cref{app:dev:tbl:results_list_basics}, and \Cref{app:dev:tbl:results:metrics} for results on the development set. There are not many notable differences between development and test performances. The development test size is relatively small per source (25 vs. 75 in the test set), making statistical tests less informative. 

That being said, the main area we were curious to observe potential differences in is performance on inconsistencies in \PG{} between development and test settings. \PG{} development instances were \textit{not} date-protected are were possibly in the pre-training corpora because they were present in the corpora we searched (\Cref{app:sec:dataset}). In the aforementioned tables, we denote \PG{} as \PGS{} (Project Gutenberg Seen) to make this distinction explicit. In general, however, the performance on average was lower for the development set versus the test set for \PG{} and not statistically different under the LMJ task metric, $\Lambda_\I$. Even though most models performed worse on \PGS{} compared to \PG{}, we still consider it good practice to select documents unlikely to be in general pre-training data; larger datasets for more statistical power are needed to draw firmer conclusions.

We did not work to overly prompt tune the models with the development set. Instead, it was used to validate our grading model, which has high agreement with manual judgments (Cohen's kappa of 0.96).

\subsection{Extended Results}
\Cref{app:tbl:results_list_hard_se} reports all three metrics for the \WLD{} source. Basic response statistics across sources and models are in \Cref{app:tbl:results_list_basics_appendix}. \Cref{app:tbl:results_list_lexical_se} reports the evidence, description, and task scores with standard errors. \Cref{app:tab:input_length_recall} reports the linear regression coefficients between input length, inconsistency type, and task score, with each dataset source as the controls.

\subsection{Duplicate Hits in Task Scores (Recall) Are Rare}
\label{app:sec:results:duplicates}
Models could return the same answer more than once within a list of answers. Additionally, the model grader could grade more than once answer as correct within a list of answers because each answer is graded independently. If these events occurred at high frequency, it would suggest the grader had a general high false positive rate.

In \Cref{app:tbl:duplicates}, we show the rate at which the LMJ grader returns True (or 1) for more than one answer in a single response. Across all the closed source models, this occurs in 13/2940 generations (less than 0.05\%). For both open source OpenAI models, it occurs in 3--6\% of generations, and it does not occur at all for both Gemma models. Upon manual inspection of the responses from the closed source models, we observed that 10/13 times were legitimate duplicates (the models truly did return the same answer more than once). For each of the other 3/13, one answer was correctly graded True (matching the expected inconsistency), and the other was a borderline case. The borderline prediction contained much of the content of the gold answer but missed some information or was the same issue present in different places of the document.

Overall, the occurrence of the grader model marking multiple answers in a single response as matching the expected inconsistency was rare enough that it does not seem to impact our results or indicate an unacceptable false positive rate.

\begin{table}[ht!]
    \centering
    \small
\begin{tabular}{lr}
\toprule
 & Duplicates \\
\midrule
gemma-3-12b-it & 0  / 420\\
gemma-3-27b-it & 0  / 420\\
gpt-oss-20b & 25  / 420\\
gpt-oss-120b & 13  / 420\\
\midrule
sonnet-v4 & 2 / 420 \\
gpt-5-mini & 2 / 420 \\
gpt-5 & 0  / 420\\
o3-mini & 0 / 420 \\
o3 & 2  / 420\\
gemini-2.5-flash & 2  / 420\\
gemini-2.5-pro & 5  / 420\\
\bottomrule
\end{tabular}
\caption{Number of times a model returned a response in which the grader model (\texttt{gpt-4.1}) found that more than one answer matched the expected inconsistency.}
\label{app:tbl:duplicates}
\end{table}

\begin{table*}[ht!]
\small
\centering
\begin{tabular}{@{}l@{\hspace{3pt}}ccc@{}}
\toprule
Model & $\Lambda_\E$ & $\Lambda_\T$ & $\Lambda_\I$ \\
\midrule
gemma-3-12b-it & \shortstack{4 \\[-0.35ex] {\scriptsize (0.8)}} & \shortstack{36 \\[-0.35ex] {\scriptsize (0.9)}} & \textbf{\shortstack{2 \\[-0.35ex] {\scriptsize (2.2)}}} \\
gemma-3-27b-it & \shortstack{6 \\[-0.35ex] {\scriptsize (1.3)}} & \textbf{\shortstack{38 \\[-0.35ex] {\scriptsize (0.8)}}} & \textbf{\shortstack{2 \\[-0.35ex] {\scriptsize (2.2)}}} \\
gpt-oss-20b & \shortstack{4 \\[-0.35ex] {\scriptsize (1.4)}} & \shortstack{29 \\[-0.35ex] {\scriptsize (1.9)}} & \shortstack{0 \\[-0.35ex] {\scriptsize (0.0)}} \\
gpt-oss-120b & \shortstack{5 \\[-0.35ex] {\scriptsize (1.1)}} & \shortstack{34 \\[-0.35ex] {\scriptsize (1.4)}} & \textbf{\shortstack{4 \\[-0.35ex] {\scriptsize (3.1)}}} \\
\midrule
sonnet-v4 & \textbf{\shortstack{15 \\[-0.35ex] {\scriptsize (3.4)}}} & \textbf{\shortstack{35 \\[-0.35ex] {\scriptsize (1.9)}}} & \textbf{\shortstack{11 \\[-0.35ex] {\scriptsize (4.7)}}} \\
gpt-5-mini & \shortstack{10 \\[-0.35ex] {\scriptsize (1.3)}} & \shortstack{35 \\[-0.35ex] {\scriptsize (1.0)}} & \shortstack{2 \\[-0.35ex] {\scriptsize (2.2)}} \\
gpt-5 & \textbf{\shortstack{12 \\[-0.35ex] {\scriptsize (2.2)}}} & \textbf{\shortstack{36 \\[-0.35ex] {\scriptsize (1.5)}}} & \textbf{\shortstack{7 \\[-0.35ex] {\scriptsize (3.7)}}} \\
o3-mini & \shortstack{0 \\[-0.35ex] {\scriptsize (0.1)}} & \shortstack{24 \\[-0.35ex] {\scriptsize (2.3)}} & \shortstack{2 \\[-0.35ex] {\scriptsize (2.2)}} \\
o3 & \shortstack{9 \\[-0.35ex] {\scriptsize (2.1)}} & \shortstack{33 \\[-0.35ex] {\scriptsize (1.5)}} & \textbf{\shortstack{9 \\[-0.35ex] {\scriptsize (4.2)}}} \\
gemini-2.5-flash & \textbf{\shortstack{12 \\[-0.35ex] {\scriptsize (2.6)}}} & \textbf{\shortstack{34 \\[-0.35ex] {\scriptsize (2.5)}}} & \textbf{\shortstack{9 \\[-0.35ex] {\scriptsize (4.2)}}} \\
gemini-2.5-pro & \textbf{\shortstack{16 \\[-0.35ex] {\scriptsize (2.4)}}} & \textbf{\shortstack{38 \\[-0.35ex] {\scriptsize (1.0)}}} & \textbf{\shortstack{9 \\[-0.35ex] {\scriptsize (4.2)}}} \\
\bottomrule
\end{tabular}
\caption{Evidence, description, and task scores (with standard errors) on \WLD{} (response-level metrics). For all three scores, higher is better. Scores are presented as percents (out of 100). 
For results with the same mean, the variance in the paired differences can still differ.
}
\label{app:tbl:results_list_hard_se}
\end{table*}

\begin{table*}[ht!]
\small
\centering
\begin{tabular}{@{}l|r@{\hspace{2pt}}r@{\hspace{2pt}}r@{\hspace{2pt}}r@{\hspace{2pt}}r|r@{\hspace{2pt}}r@{\hspace{2pt}}r@{\hspace{2pt}}r@{\hspace{2pt}}r|r@{\hspace{2pt}}r@{\hspace{2pt}}r@{\hspace{2pt}}r@{\hspace{2pt}}r@{}}
\toprule
 & \multicolumn{5}{c}{Valid} & \multicolumn{5}{c}{kTok/\DD} & \multicolumn{5}{c}{$\hat{\I}$/\DD} \\
Model & \BLS{} & \PG{} & \PRE{} & \SEC{} & \USPF{} & \BLS{} & \PG{} & \PRE{} & \SEC{} & \USPF{} & \BLS{} & \PG{} & \PRE{} & \SEC{} & \USPF{} \\
\midrule
gemma-3-12b-it & 100.0 & 100.0 & 100.0 & 100.0 & 98.7 & 0.6 & 0.6 & 0.7 & 0.6 & 0.6 & 2.5 & 4.4 & 3.5 & 3.1 & 3.5 \\
gemma-3-27b-it & 100.0 & 100.0 & 100.0 & 100.0 & 98.7 & 0.8 & 0.8 & 0.9 & 0.9 & 0.9 & 4.1 & 5.3 & 4.4 & 4.3 & 4.6 \\
gpt-oss-20b & 100.0 & 94.7 & 98.7 & 93.3 & 96.0 & 4.8 & 4.3 & 6.4 & 6.9 & 4.8 & 2.7 & 2.3 & 2.4 & 2.2 & 2.1 \\
gpt-oss-120b & 100.0 & 98.7 & 100.0 & 100.0 & 98.7 & 2.6 & 2.5 & 2.7 & 2.9 & 2.7 & 2.2 & 1.5 & 1.9 & 1.8 & 1.9 \\
\midrule
sonnet-v4 & 100.0 & 100.0 & 100.0 & 92.0 & 96.0 & 3.2 & 4.8 & 4.5 & 6.5 & 4.8 & 1.5 & 1.4 & 2.2 & 1.3 & 1.8 \\
gpt-5-mini & 100.0 & 100.0 & 100.0 & 100.0 & 100.0 & 4.7 & 5.2 & 6.0 & 7.3 & 6.0 & 1.8 & 2.3 & 2.8 & 2.3 & 3.4 \\
gpt-5 & 97.3 & 100.0 & 98.7 & 100.0 & 100.0 & 9.1 & 8.6 & 13.7 & 12.9 & 10.0 & 1.9 & 2.9 & 3.4 & 2.8 & 4.3 \\
o3-mini & 100.0 & 100.0 & 100.0 & 97.3 & 100.0 & 2.3 & 2.7 & 3.4 & 3.0 & 2.7 & 1.3 & 1.5 & 1.5 & 1.2 & 1.3 \\
o3 & 100.0 & 100.0 & 100.0 & 98.7 & 98.7 & 4.8 & 4.1 & 7.2 & 6.9 & 3.8 & 1.4 & 1.7 & 2.2 & 1.7 & 2.2 \\
gemini-2.5-flash & 100.0 & 90.7 & 100.0 & 89.3 & 97.3 & 6.4 & 11.8 & 8.3 & 15.9 & 13.2 & 2.7 & 5.2 & 3.8 & 3.4 & 5.3 \\
gemini-2.5-pro & 100.0 & 100.0 & 100.0 & 97.3 & 100.0 & 8.7 & 10.6 & 11.1 & 17.2 & 16.0 & 2.7 & 5.1 & 5.3 & 4.5 & 6.9 \\
\bottomrule
\end{tabular}

\vspace{1em}

\begin{tabular}{@{}l|r@{\hspace{2pt}}r@{\hspace{2pt}}r@{\hspace{2pt}}r@{\hspace{2pt}}r|r@{\hspace{2pt}}r@{\hspace{2pt}}r@{\hspace{2pt}}r@{\hspace{2pt}}r@{}}
\toprule
 & \multicolumn{5}{c}{\ee/$\hat{\I}$} & \multicolumn{5}{c}{Tok/\ee} \\
Model & \BLS{} & \PG{} & \PRE{} & \SEC{} & \USPF{} & \BLS{} & \PG{} & \PRE{} & \SEC{} & \USPF{} \\
\midrule
gemma-3-12b-it & 2.9 & 2.0 & 2.4 & 2.5 & 2.3 & 12.1 & 14.1 & 18.6 & 19.4 & 18.9 \\
gemma-3-27b-it & 2.9 & 2.2 & 2.6 & 2.6 & 2.4 & 17.3 & 21.3 & 24.9 & 22.3 & 25.7 \\
gpt-oss-20b & 6.0 & 4.9 & 6.6 & 6.7 & 5.1 & 7.6 & 8.8 & 6.6 & 6.9 & 12.1 \\
gpt-oss-120b & 4.1 & 3.1 & 3.5 & 2.7 & 2.7 & 13.6 & 16.0 & 16.1 & 21.2 & 20.7 \\
\midrule
sonnet-v4 & 3.7 & 4.4 & 4.4 & 6.1 & 4.1 & 8.8 & 11.3 & 9.6 & 6.7 & 11.2 \\
gpt-5-mini & 2.6 & 2.5 & 2.7 & 2.6 & 2.3 & 33.8 & 29.4 & 36.3 & 29.5 & 32.8 \\
gpt-5 & 2.6 & 2.3 & 2.5 & 2.6 & 2.2 & 22.7 & 24.1 & 28.4 & 23.3 & 30.8 \\
o3-mini & 4.4 & 5.4 & 3.1 & 7.7 & 5.1 & 8.5 & 10.3 & 17.0 & 6.2 & 12.2 \\
o3 & 2.1 & 2.6 & 2.1 & 2.5 & 2.3 & 25.4 & 19.3 & 28.7 & 21.7 & 23.0 \\
gemini-2.5-flash & 6.2 & 2.7 & 4.1 & 4.8 & 3.4 & 13.4 & 47.3 & 25.1 & 22.2 & 39.6 \\
gemini-2.5-pro & 3.1 & 2.2 & 3.6 & 4.5 & 3.5 & 18.8 & 28.1 & 19.2 & 13.1 & 22.7 \\
\bottomrule
\end{tabular}
\caption{\textbf{Model response statistics across data sources.} Format validity rate (\%), mean tokens per document (in thousands) (kTok$/\D$), mean answers per response ($\hat{\I}/\D$), mean evidence spans per answer ($\e/\hat{\I}$), and mean tokens per evidence span (Tok/\ee).}
\label{app:tbl:results_list_basics_appendix}
\end{table*}

\begin{table*}[ht!]
\small
\centering
\begin{tabular}{@{}l@{\hspace{3pt}}r@{\hspace{3pt}}r@{\hspace{3pt}}r@{\hspace{3pt}}r@{\hspace{3pt}}r@{\hspace{3pt}}|@{\hspace{3pt}}r@{}}
\toprule
& \multicolumn{6}{c}{\textbf{Evidence Score ($\Lambda_\E$)}} \\
\midrule
Model & \BLS{} & \PRE{} & \SEC{} & \USPF{} & \PG{} & \AVG{} \\
\midrule
gemma-3-12b-it & \shortstack{18 \\[-0.35ex] {\scriptsize (2.4)}} & \shortstack{5 \\[-0.35ex] {\scriptsize (0.6)}} & \shortstack{6 \\[-0.35ex] {\scriptsize (1.2)}} & \shortstack{3 \\[-0.35ex] {\scriptsize (0.5)}} & \shortstack{4 \\[-0.35ex] {\scriptsize (1.0)}} & \shortstack{7 \\[-0.35ex] {\scriptsize (2.5)}} \\[2pt]
gemma-3-27b-it & \shortstack{23 \\[-0.35ex] {\scriptsize (2.7)}} & \shortstack{8 \\[-0.35ex] {\scriptsize (0.9)}} & \shortstack{4 \\[-0.35ex] {\scriptsize (0.7)}} & \shortstack{1 \\[-0.35ex] {\scriptsize (0.4)}} & \shortstack{0 \\[-0.35ex] {\scriptsize (0.2)}} & \shortstack{7 \\[-0.35ex] {\scriptsize (3.7)}} \\[2pt]
gpt-oss-20b & \shortstack{30 \\[-0.35ex] {\scriptsize (3.4)}} & \shortstack{11 \\[-0.35ex] {\scriptsize (2.2)}} & \shortstack{3 \\[-0.35ex] {\scriptsize (0.6)}} & \shortstack{2 \\[-0.35ex] {\scriptsize (0.9)}} & \shortstack{2 \\[-0.35ex] {\scriptsize (1.3)}} & \shortstack{10 \\[-0.35ex] {\scriptsize (4.8)}} \\[2pt]
gpt-oss-120b & \shortstack{35 \\[-0.35ex] {\scriptsize (3.5)}} & \shortstack{14 \\[-0.35ex] {\scriptsize (2.2)}} & \shortstack{7 \\[-0.35ex] {\scriptsize (1.7)}} & \shortstack{5 \\[-0.35ex] {\scriptsize (1.3)}} & \shortstack{4 \\[-0.35ex] {\scriptsize (1.5)}} & \shortstack{13 \\[-0.35ex] {\scriptsize (5.1)}} \\
\midrule
sonnet-v4 & \textbf{\shortstack{44 \\[-0.35ex] {\scriptsize (3.3)}}} & \textbf{\shortstack{35 \\[-0.35ex] {\scriptsize (3.6)}}} & \shortstack{13 \\[-0.35ex] {\scriptsize (2.3)}} & \shortstack{14 \\[-0.35ex] {\scriptsize (2.9)}} & \shortstack{9 \\[-0.35ex] {\scriptsize (2.3)}} & \textbf{\shortstack{23 \\[-0.35ex] {\scriptsize (6.2)}}} \\[2pt]
gpt-5-mini & \shortstack{37 \\[-0.35ex] {\scriptsize (3.0)}} & \shortstack{27 \\[-0.35ex] {\scriptsize (2.9)}} & \shortstack{25 \\[-0.35ex] {\scriptsize (2.9)}} & \shortstack{17 \\[-0.35ex] {\scriptsize (2.3)}} & \shortstack{9 \\[-0.35ex] {\scriptsize (1.9)}} & \textbf{\shortstack{23 \\[-0.35ex] {\scriptsize (4.2)}}} \\[2pt]
gpt-5 & \textbf{\shortstack{43 \\[-0.35ex] {\scriptsize (3.2)}}} & \textbf{\shortstack{36 \\[-0.35ex] {\scriptsize (3.2)}}} & \textbf{\shortstack{42 \\[-0.35ex] {\scriptsize (3.2)}}} & \textbf{\shortstack{27 \\[-0.35ex] {\scriptsize (3.4)}}} & \textbf{\shortstack{18 \\[-0.35ex] {\scriptsize (2.9)}}} & \textbf{\shortstack{33 \\[-0.35ex] {\scriptsize (4.2)}}} \\[2pt]
o3-mini & \shortstack{23 \\[-0.35ex] {\scriptsize (3.6)}} & \shortstack{5 \\[-0.35ex] {\scriptsize (1.4)}} & \shortstack{1 \\[-0.35ex] {\scriptsize (0.3)}} & \shortstack{0 \\[-0.35ex] {\scriptsize (0.2)}} & \shortstack{1 \\[-0.35ex] {\scriptsize (1.3)}} & \shortstack{6 \\[-0.35ex] {\scriptsize (3.8)}} \\[2pt]
o3 & \textbf{\shortstack{40 \\[-0.35ex] {\scriptsize (3.0)}}} & \shortstack{29 \\[-0.35ex] {\scriptsize (3.2)}} & \shortstack{25 \\[-0.35ex] {\scriptsize (2.9)}} & \shortstack{17 \\[-0.35ex] {\scriptsize (2.9)}} & \shortstack{10 \\[-0.35ex] {\scriptsize (2.3)}} & \textbf{\shortstack{24 \\[-0.35ex] {\scriptsize (4.7)}}} \\[2pt]
gemini-2.5-flash & \textbf{\shortstack{42 \\[-0.35ex] {\scriptsize (3.5)}}} & \shortstack{25 \\[-0.35ex] {\scriptsize (2.9)}} & \shortstack{19 \\[-0.35ex] {\scriptsize (2.9)}} & \shortstack{13 \\[-0.35ex] {\scriptsize (2.6)}} & \shortstack{5 \\[-0.35ex] {\scriptsize (1.0)}} & \textbf{\shortstack{21 \\[-0.35ex] {\scriptsize (5.5)}}} \\[2pt]
gemini-2.5-pro & \textbf{\shortstack{45 \\[-0.35ex] {\scriptsize (3.6)}}} & \textbf{\shortstack{38 \\[-0.35ex] {\scriptsize (3.1)}}} & \textbf{\shortstack{36 \\[-0.35ex] {\scriptsize (3.3)}}} & \textbf{\shortstack{27 \\[-0.35ex] {\scriptsize (3.3)}}} & \shortstack{12 \\[-0.35ex] {\scriptsize (2.1)}} & \textbf{\shortstack{32 \\[-0.35ex] {\scriptsize (5.1)}}} \\
\bottomrule
\end{tabular}
\begin{tabular}{@{}@{\hspace{3pt}}r@{\hspace{3pt}}r@{\hspace{3pt}}r@{\hspace{3pt}}r@{\hspace{3pt}}r@{\hspace{3pt}}|@{\hspace{3pt}}r@{}}
\toprule
\multicolumn{6}{c}{\textbf{Description Score ($\Lambda_\T$)}} \\
\midrule
\BLS{} & \PRE{} & \SEC{} & \USPF{} & \PG{} & \AVG{} \\
\midrule
\shortstack{43 \\[-0.35ex] {\scriptsize (1.4)}} & \shortstack{37 \\[-0.35ex] {\scriptsize (0.8)}} & \shortstack{35 \\[-0.35ex] {\scriptsize (0.8)}} & \shortstack{34 \\[-0.35ex] {\scriptsize (0.8)}} & \shortstack{33 \\[-0.35ex] {\scriptsize (0.9)}} & \shortstack{36 \\[-0.35ex] {\scriptsize (1.6)}} \\[2pt]
\shortstack{45 \\[-0.35ex] {\scriptsize (1.3)}} & \shortstack{38 \\[-0.35ex] {\scriptsize (0.7)}} & \shortstack{36 \\[-0.35ex] {\scriptsize (0.7)}} & \shortstack{34 \\[-0.35ex] {\scriptsize (0.8)}} & \shortstack{34 \\[-0.35ex] {\scriptsize (0.8)}} & \shortstack{37 \\[-0.35ex] {\scriptsize (1.9)}} \\[2pt]
\shortstack{45 \\[-0.35ex] {\scriptsize (1.9)}} & \shortstack{32 \\[-0.35ex] {\scriptsize (1.9)}} & \shortstack{27 \\[-0.35ex] {\scriptsize (1.7)}} & \shortstack{28 \\[-0.35ex] {\scriptsize (1.4)}} & \shortstack{28 \\[-0.35ex] {\scriptsize (1.6)}} & \shortstack{32 \\[-0.35ex] {\scriptsize (3.0)}} \\[2pt]
\shortstack{47 \\[-0.35ex] {\scriptsize (2.0)}} & \shortstack{36 \\[-0.35ex] {\scriptsize (1.5)}} & \shortstack{34 \\[-0.35ex] {\scriptsize (0.9)}} & \shortstack{34 \\[-0.35ex] {\scriptsize (1.0)}} & \shortstack{32 \\[-0.35ex] {\scriptsize (1.5)}} & \shortstack{37 \\[-0.35ex] {\scriptsize (2.4)}} \\
\midrule
\textbf{\shortstack{53 \\[-0.35ex] {\scriptsize (1.5)}}} & \shortstack{43 \\[-0.35ex] {\scriptsize (1.4)}} & \shortstack{30 \\[-0.35ex] {\scriptsize (2.0)}} & \shortstack{33 \\[-0.35ex] {\scriptsize (1.7)}} & \shortstack{32 \\[-0.35ex] {\scriptsize (2.1)}} & \textbf{\shortstack{38 \\[-0.35ex] {\scriptsize (3.9)}}} \\[2pt]
\shortstack{46 \\[-0.35ex] {\scriptsize (1.2)}} & \shortstack{43 \\[-0.35ex] {\scriptsize (1.0)}} & \shortstack{39 \\[-0.35ex] {\scriptsize (0.9)}} & \shortstack{38 \\[-0.35ex] {\scriptsize (1.1)}} & \shortstack{38 \\[-0.35ex] {\scriptsize (1.5)}} & \textbf{\shortstack{41 \\[-0.35ex] {\scriptsize (1.3)}}} \\[2pt]
\shortstack{50 \\[-0.35ex] {\scriptsize (1.5)}} & \textbf{\shortstack{46 \\[-0.35ex] {\scriptsize (1.4)}}} & \textbf{\shortstack{44 \\[-0.35ex] {\scriptsize (1.1)}}} & \textbf{\shortstack{41 \\[-0.35ex] {\scriptsize (1.2)}}} & \textbf{\shortstack{43 \\[-0.35ex] {\scriptsize (1.5)}}} & \textbf{\shortstack{45 \\[-0.35ex] {\scriptsize (1.4)}}} \\[2pt]
\shortstack{39 \\[-0.35ex] {\scriptsize (2.7)}} & \shortstack{31 \\[-0.35ex] {\scriptsize (1.8)}} & \shortstack{19 \\[-0.35ex] {\scriptsize (2.1)}} & \shortstack{23 \\[-0.35ex] {\scriptsize (1.5)}} & \shortstack{22 \\[-0.35ex] {\scriptsize (2.0)}} & \shortstack{27 \\[-0.35ex] {\scriptsize (3.3)}} \\[2pt]
\shortstack{49 \\[-0.35ex] {\scriptsize (1.1)}} & \shortstack{42 \\[-0.35ex] {\scriptsize (1.4)}} & \shortstack{39 \\[-0.35ex] {\scriptsize (1.5)}} & \shortstack{35 \\[-0.35ex] {\scriptsize (1.4)}} & \shortstack{37 \\[-0.35ex] {\scriptsize (1.9)}} & \textbf{\shortstack{40 \\[-0.35ex] {\scriptsize (2.2)}}} \\[2pt]
\textbf{\shortstack{53 \\[-0.35ex] {\scriptsize (1.4)}}} & \shortstack{45 \\[-0.35ex] {\scriptsize (1.0)}} & \shortstack{38 \\[-0.35ex] {\scriptsize (1.7)}} & \shortstack{39 \\[-0.35ex] {\scriptsize (1.3)}} & \shortstack{36 \\[-0.35ex] {\scriptsize (1.7)}} & \textbf{\shortstack{42 \\[-0.35ex] {\scriptsize (2.7)}}} \\[2pt]
\textbf{\shortstack{54 \\[-0.35ex] {\scriptsize (1.2)}}} & \textbf{\shortstack{48 \\[-0.35ex] {\scriptsize (1.2)}}} & \textbf{\shortstack{42 \\[-0.35ex] {\scriptsize (1.3)}}} & \textbf{\shortstack{43 \\[-0.35ex] {\scriptsize (1.2)}}} & \textbf{\shortstack{44 \\[-0.35ex] {\scriptsize (1.5)}}} & \textbf{\shortstack{46 \\[-0.35ex] {\scriptsize (1.9)}}} \\
\bottomrule
\end{tabular}
\begin{tabular}{@{}r@{\hspace{3pt}}r@{\hspace{3pt}}r@{\hspace{3pt}}r@{\hspace{3pt}}r|@{\hspace{3pt}}r@{}}
\toprule
\multicolumn{6}{c}{\textbf{Task Score ($\Lambda_\I$)}} \\
\midrule
 \BLS{} & \PRE{} & \SEC{} & \USPF{} & \PG{} & \AVG{} \\
\midrule
 \shortstack{57 \\[-0.35ex] {\scriptsize (5.7)}} & \shortstack{25 \\[-0.35ex] {\scriptsize (5.0)}} & \shortstack{7 \\[-0.35ex] {\scriptsize (2.9)}} & \shortstack{3 \\[-0.35ex] {\scriptsize (1.9)}} & \shortstack{9 \\[-0.35ex] {\scriptsize (3.4)}} & \shortstack{20 \\[-0.35ex] {\scriptsize (9.0)}} \\[2pt]
 \shortstack{61 \\[-0.35ex] {\scriptsize (5.6)}} & \shortstack{27 \\[-0.35ex] {\scriptsize (5.1)}} & \shortstack{7 \\[-0.35ex] {\scriptsize (2.9)}} & \shortstack{8 \\[-0.35ex] {\scriptsize (3.1)}} & \shortstack{12 \\[-0.35ex] {\scriptsize (3.8)}} & \shortstack{23 \\[-0.35ex] {\scriptsize (9.2)}} \\[2pt]
 \shortstack{21 \\[-0.35ex] {\scriptsize (4.7)}} & \shortstack{3 \\[-0.35ex] {\scriptsize (1.9)}} & \shortstack{1 \\[-0.35ex] {\scriptsize (1.3)}} & \shortstack{0 \\[-0.35ex] {\scriptsize (0.0)}} & \shortstack{3 \\[-0.35ex] {\scriptsize (1.9)}} & \shortstack{6 \\[-0.35ex] {\scriptsize (3.5)}} \\[2pt]
 \shortstack{28 \\[-0.35ex] {\scriptsize (5.2)}} & \shortstack{3 \\[-0.35ex] {\scriptsize (1.9)}} & \shortstack{0 \\[-0.35ex] {\scriptsize (0.0)}} & \shortstack{0 \\[-0.35ex] {\scriptsize (0.0)}} & \shortstack{1 \\[-0.35ex] {\scriptsize (1.3)}} & \shortstack{6 \\[-0.35ex] {\scriptsize (4.9)}} \\
\midrule
 \textbf{\shortstack{83 \\[-0.35ex] {\scriptsize (4.4)}}} & \shortstack{49 \\[-0.35ex] {\scriptsize (5.8)}} & \shortstack{20 \\[-0.35ex] {\scriptsize (4.6)}} & \shortstack{21 \\[-0.35ex] {\scriptsize (4.7)}} & \shortstack{23 \\[-0.35ex] {\scriptsize (4.8)}} & \shortstack{39 \\[-0.35ex] {\scriptsize (10.9)}} \\[2pt]
 \shortstack{80 \\[-0.35ex] {\scriptsize (4.6)}} & \shortstack{53 \\[-0.35ex] {\scriptsize (5.8)}} & \shortstack{47 \\[-0.35ex] {\scriptsize (5.8)}} & \textbf{\shortstack{37 \\[-0.35ex] {\scriptsize (5.6)}}} & \shortstack{39 \\[-0.35ex] {\scriptsize (5.6)}} & \shortstack{51 \\[-0.35ex] {\scriptsize (6.9)}} \\[2pt]
\textbf{\shortstack{87 \\[-0.35ex] {\scriptsize (3.9)}}} & \textbf{\shortstack{67 \\[-0.35ex] {\scriptsize (5.4)}}} & \textbf{\shortstack{73 \\[-0.35ex] {\scriptsize (5.1)}}} & \textbf{\shortstack{43 \\[-0.35ex] {\scriptsize (5.7)}}} & \textbf{\shortstack{52 \\[-0.35ex] {\scriptsize (5.8)}}} & \textbf{\shortstack{64 \\[-0.35ex] {\scriptsize (6.9)}}} \\[2pt]
 \shortstack{56 \\[-0.35ex] {\scriptsize (5.7)}} & \shortstack{17 \\[-0.35ex] {\scriptsize (4.4)}} & \shortstack{8 \\[-0.35ex] {\scriptsize (3.1)}} & \shortstack{3 \\[-0.35ex] {\scriptsize (1.9)}} & \shortstack{12 \\[-0.35ex] {\scriptsize (3.8)}} & \shortstack{19 \\[-0.35ex] {\scriptsize (8.5)}} \\[2pt]
\textbf{\shortstack{85 \\[-0.35ex] {\scriptsize (4.1)}}} & \shortstack{49 \\[-0.35ex] {\scriptsize (5.8)}} & \shortstack{57 \\[-0.35ex] {\scriptsize (5.7)}} & \shortstack{31 \\[-0.35ex] {\scriptsize (5.3)}} & \shortstack{40 \\[-0.35ex] {\scriptsize (5.7)}} & \shortstack{53 \\[-0.35ex] {\scriptsize (8.4)}} \\[2pt]
 \shortstack{75 \\[-0.35ex] {\scriptsize (5.0)}} & \shortstack{48 \\[-0.35ex] {\scriptsize (5.8)}} & \shortstack{36 \\[-0.35ex] {\scriptsize (5.5)}} & \shortstack{25 \\[-0.35ex] {\scriptsize (5.0)}} & \shortstack{31 \\[-0.35ex] {\scriptsize (5.3)}} & \shortstack{43 \\[-0.35ex] {\scriptsize (7.9)}} \\[2pt]
 \textbf{\shortstack{88 \\[-0.35ex] {\scriptsize (3.8)}}} & \textbf{\shortstack{71 \\[-0.35ex] {\scriptsize (5.3)}}} & \shortstack{61 \\[-0.35ex] {\scriptsize (5.6)}} & \textbf{\shortstack{40 \\[-0.35ex] {\scriptsize (5.7)}}} & \textbf{\shortstack{45 \\[-0.35ex] {\scriptsize (5.7)}}} & \textbf{\shortstack{61 \\[-0.35ex] {\scriptsize (7.8)}}} \\
\bottomrule
\end{tabular}
\caption{Evidence, description, and task scores (with standard errors) across the \datasetname{} test set (response-level metrics). For all three scores, higher is better. Scores are presented as percents (out of 100). These results are an extension of those shown in the main body. For the average column, we use a clustered test to account for information correlated within a data source.}
\label{app:tbl:results_list_lexical_se}
\end{table*}


\begin{table*}[t]
\centering
\small
\begin{tabular}{lcccllll}
\toprule
 & \multicolumn{3}{c}{Task Score ($\Lambda_\I$)} & \multicolumn{4}{c}{Coefficients (pp)} \\
\cmidrule(lr){2-4} \cmidrule(lr){5-8}
Model & Num & Non & Struc & Len & Num & Non & $R^2$ \\
\midrule
gemma-3-12b-it & 0.06 & 0.10 & 0.02 & -1.0* & +2.5 & +5.7 & 0.14 \\
gemma-3-27b-it & 0.10 & 0.03 & 0.02 & -1.4** & +5.1 & -1.4 & 0.23 \\
gpt-oss-20b & 0.26 & 0.25 & 0.06 & -2.6*** & +12.8** & +12.4* & 0.29 \\
gpt-oss-120b & 0.27 & 0.38 & 0.06 & -2.7*** & +13.9** & +25.3*** & 0.30 \\
\midrule
sonnet-v4 & 0.45 & 0.49 & 0.23 & -4.0*** & +12.4* & +17.4** & 0.30 \\
gpt-5-mini & 0.60 & 0.56 & 0.31 & -2.6* & +23.9*** & +19.3* & 0.15 \\
gpt-5 & 0.68 & 0.70 & 0.54 & -2.6* & +9.5 & +11.1 & 0.13 \\
o3-mini & 0.24 & 0.27 & 0.06 & -3.5*** & +11.8** & +14.8** & 0.30 \\
o3 & 0.58 & 0.59 & 0.39 & -4.3*** & +13.5* & +13.8 & 0.19 \\
gemini-2.5-flash & 0.50 & 0.46 & 0.28 & -2.8** & +16.2** & +12.3 & 0.16 \\
gemini-2.5-pro & 0.67 & 0.65 & 0.48 & -2.6* & +11.7* & +10.4 & 0.15 \\
\bottomrule
\end{tabular}
\caption{Task score by inconsistency type with regression coefficients for the types and document length. The average task score is shown for each type. Coefficients (pp): Len = change in task score ($\Lambda_\I$) per 10k tokens; Num/Non (which refers to Numeric/Non-numeric inconsistency types) columns indicate the effect compared to the structural inconsistency type (as a reference). Significance: *** $p<0.001$, ** $p<0.01$, * $p<0.05$.}
\label{app:tab:input_length_recall}
\end{table*}

\begin{table*}[t]
\centering
\small
\begin{tabular}{lrrrrr|c|rrrrr}
\toprule
 & \multicolumn{5}{c|}{Count} & & \multicolumn{5}{c}{Task Score ($\Lambda_\I$) Across Models} \\
\cmidrule{2-6} \cmidrule{8-12}
Length & \BLS{} & \PRE{} & \SEC{} & \USPF{} & \PG{} & & \BLS{} & \PRE{} & \SEC{} & \USPF{} & \PG{} \\
\midrule
0--10k & 52 & 32 & 0 & 0 & 1 & & 0.72 & 0.43 & -- & -- & 1.00 \\
10--20k & 9 & 38 & 1 & 1 & 7 & & 0.63 & 0.33 & 0.00 & 0.67 & 0.41 \\
20--30k & 6 & 5 & 12 & 3 & 7 & & 0.45 & 0.34 & 0.44 & 0.39 & 0.32 \\
30--40k & 3 & 0 & 26 & 4 & 6 & & 0.54 & -- & 0.26 & 0.31 & 0.30 \\
40--50k & 2 & 0 & 15 & 3 & 2 & & 0.43 & -- & 0.25 & 0.03 & 0.09 \\
50--60k & 2 & 0 & 9 & 4 & 7 & & 0.44 & -- & 0.34 & 0.30 & 0.36 \\
60--70k & 1 & 0 & 3 & 3 & 2 & & 0.00 & -- & 0.22 & 0.00 & 0.64 \\
70--80k & 0 & 0 & 1 & 10 & 4 & & 0.00 & -- & 0.23 & 0.22 & 0.18 \\
80--90k & 0 & 0 & 1 & 3 & 8 & & 0.00 & -- & 0.44 & 0.08 & 0.19 \\
90--100k & 0 & 0 & 1 & 5 & 1 & & -- & -- & 0.18 & 0.21 & 0.06 \\
100k & 0 & 0 & 6 & 39 & 30 & & -- & -- & 0.22 & 0.17 & 0.13 \\
\bottomrule
\end{tabular}
\caption{Data distribution and task score ($\Lambda_\I$) by document length and source. The document lengths are based on the \model{gpt-5} tokenizer. The task score is averaged across all models.}
\label{tab:dataset_bin_distribution}
\end{table*}

\begin{table*}
\small
\centering
\begin{tabular}{@{}l@{\hspace{4pt}}r@{\hspace{4pt}}r@{\hspace{4pt}}r@{\hspace{4pt}}r@{\hspace{4pt}}r@{}}
\toprule
 & Docs (\DD) & kTok/\DD & \ee/\DD & Tok/\ee & \ee~Pos\% \\
\midrule
\BLS{} & 25 & 17$_{\pm21.1}$ & 5$_{\pm3.1}$ & 19$_{\pm44.1}$ & 23$_{\pm18.8}$ \\
\PRE{} & 25 & 13$_{\pm6.9}$ & 5$_{\pm3.8}$ & 9$_{\pm9.5}$ & 41$_{\pm20.7}$ \\
\SEC{} & 26 & 43$_{\pm22.4}$ & 4$_{\pm2.5}$ & 14$_{\pm11.5}$ & 36$_{\pm26.5}$ \\
\USPF{} & 24 & 95$_{\pm57.2}$ & 4$_{\pm3.0}$ & 12$_{\pm11.1}$ & 26$_{\pm19.4}$ \\
\PGS{} & 25 & 146$_{\pm111.8}$ & 5$_{\pm3.7}$ & 7$_{\pm4.8}$ & 22$_{\pm17.9}$ \\
\midrule
\datasetname{} & 125 & 62$_{\pm76.8}$ & 5$_{\pm3.3}$ & 12$_{\pm21.9}$ & 30$_{\pm22.2}$ \\
\bottomrule
\end{tabular}
\caption{\textbf{Statistics by document source} \textbf{(development set)}: number of documents (\DD), mean tokens per document (in thousands) (kTok/\DD), mean evidence spans per document (\ee/\DD), mean tokens per evidence span (Tok/\ee), and mean relative position of evidence within documents (\ee~Pos\%). Subscripts denote standard deviations. \datasetname{} row reports statistics pooled over the sources above.}
\label{app:dev:tbl:dataset:dev}
\end{table*}
\begin{table*}
\small
\centering
\begin{tabular}{@{}l@{\hspace{4pt}}r@{\hspace{4pt}}r@{\hspace{4pt}}r@{\hspace{4pt}}r@{\hspace{4pt}}r@{}}
\toprule
Model & Valid & kTok/\DD & $\hat{\I}$/\DD & \ee/$\hat{\I}$ & Tok/\ee \\
\midrule
sonnet-v4 & 96.0 & 5.0$_{\pm0.2}$ & 1.50$_{\pm0.1}$ & 4.33$_{\pm0.4}$ & 10$_{\pm1.0}$ \\
gpt-5-mini & 100.0 & 6.1$_{\pm0.2}$ & 2.63$_{\pm0.1}$ & 2.55$_{\pm0.1}$ & 32$_{\pm2.0}$ \\
gpt-5 & 99.2 & 11$_{\pm0.3}$ & 3.00$_{\pm0.1}$ & 2.36$_{\pm0.1}$ & 29$_{\pm1.3}$ \\
o3-mini & 100.0 & 2.8$_{\pm0.1}$ & 1.38$_{\pm0.0}$ & 4.65$_{\pm0.5}$ & 12$_{\pm1.7}$ \\
o3 & 100.0 & 5.4$_{\pm0.3}$ & 1.77$_{\pm0.1}$ & 2.57$_{\pm0.2}$ & 22$_{\pm1.9}$ \\
gemini-2.5-flash & 95.2 & 11$_{\pm0.4}$ & 4.05$_{\pm0.3}$ & 3.68$_{\pm0.3}$ & 28$_{\pm2.7}$ \\
gemini-2.5-pro & 99.2 & 13$_{\pm0.4}$ & 5.18$_{\pm0.3}$ & 3.06$_{\pm0.1}$ & 22$_{\pm1.7}$ \\
\bottomrule
\end{tabular}
\caption{\textbf{Model response statistics over} \datasetname{} \textbf{development set.} Format validity rate (\%), mean tokens per document (in thousands) (kTok/$\D$), mean answers per response ($\hat{\I}$/$\D$), mean evidence spans per answer ($\e$/$\hat{\I}$), and mean tokens per evidence span (Tok/\ee). Subscripts denote standard deviations.}
\label{app:dev:tbl:results_list_basics}
\end{table*}

\begin{table*}[ht!]
\small
\centering
\begin{tabular}{@{}l@{\hspace{3pt}}r@{\hspace{3pt}}r@{\hspace{3pt}}r@{\hspace{3pt}}r@{\hspace{3pt}}r@{\hspace{3pt}}|@{\hspace{3pt}}r@{}}
\toprule
& \multicolumn{6}{c}{\textbf{Evidence Score ($\Lambda_\E$)}} \\
\midrule
Model & \BLS{} & \PRE{} & \SEC{} & \USPF{} & \PGS{} & \AVG{} \\
\midrule
sonnet-v4 & \textbf{\shortstack{35 \\[-0.35ex] {\scriptsize (6.9)}}} & \shortstack{25 \\[-0.35ex] {\scriptsize (5.4)}} & \shortstack{20 \\[-0.35ex] {\scriptsize (4.5)}} & \textbf{\shortstack{19 \\[-0.35ex] {\scriptsize (6.0)}}} & \shortstack{3 \\[-0.35ex] {\scriptsize (0.7)}} & \textbf{\shortstack{20 \\[-0.35ex] {\scriptsize (4.7)}}} \\[2pt]
gpt-5-mini & \shortstack{28 \\[-0.35ex] {\scriptsize (5.0)}} & \shortstack{19 \\[-0.35ex] {\scriptsize (4.3)}} & \shortstack{22 \\[-0.35ex] {\scriptsize (4.2)}} & \shortstack{17 \\[-0.35ex] {\scriptsize (5.0)}} & \shortstack{5 \\[-0.35ex] {\scriptsize (1.8)}} & \textbf{\shortstack{18 \\[-0.35ex] {\scriptsize (3.4)}}} \\[2pt]
gpt-5 & \textbf{\shortstack{37 \\[-0.35ex] {\scriptsize (5.1)}}} & \textbf{\shortstack{32 \\[-0.35ex] {\scriptsize (6.2)}}} & \textbf{\shortstack{35 \\[-0.35ex] {\scriptsize (5.3)}}} & \textbf{\shortstack{24 \\[-0.35ex] {\scriptsize (5.5)}}} & \shortstack{8 \\[-0.35ex] {\scriptsize (3.1)}} & \textbf{\shortstack{27 \\[-0.35ex] {\scriptsize (4.7)}}} \\[2pt]
o3-mini & \shortstack{19 \\[-0.35ex] {\scriptsize (6.6)}} & \shortstack{4 \\[-0.35ex] {\scriptsize (1.0)}} & \shortstack{0 \\[-0.35ex] {\scriptsize (0.2)}} & \shortstack{0 \\[-0.35ex] {\scriptsize (0.1)}} & \shortstack{0 \\[-0.35ex] {\scriptsize (0.1)}} & \shortstack{5 \\[-0.35ex] {\scriptsize (3.2)}} \\[2pt]
o3 & \textbf{\shortstack{35 \\[-0.35ex] {\scriptsize (6.0)}}} & \textbf{\shortstack{25 \\[-0.35ex] {\scriptsize (5.3)}}} & \shortstack{17 \\[-0.35ex] {\scriptsize (3.7)}} & \shortstack{15 \\[-0.35ex] {\scriptsize (4.6)}} & \shortstack{3 \\[-0.35ex] {\scriptsize (1.2)}} & \textbf{\shortstack{19 \\[-0.35ex] {\scriptsize (4.7)}}} \\[2pt]
gemini-2.5-flash & \textbf{\shortstack{34 \\[-0.35ex] {\scriptsize (5.8)}}} & \shortstack{20 \\[-0.35ex] {\scriptsize (5.2)}} & \shortstack{19 \\[-0.35ex] {\scriptsize (4.4)}} & \shortstack{17 \\[-0.35ex] {\scriptsize (4.8)}} & \shortstack{2 \\[-0.35ex] {\scriptsize (0.8)}} & \textbf{\shortstack{18 \\[-0.35ex] {\scriptsize (4.5)}}} \\[2pt]
gemini-2.5-pro & \textbf{\shortstack{40 \\[-0.35ex] {\scriptsize (6.2)}}} & \textbf{\shortstack{35 \\[-0.35ex] {\scriptsize (5.1)}}} & \textbf{\shortstack{31 \\[-0.35ex] {\scriptsize (4.9)}}} & \textbf{\shortstack{25 \\[-0.35ex] {\scriptsize (5.9)}}} & \textbf{\shortstack{18 \\[-0.35ex] {\scriptsize (4.3)}}} & \textbf{\shortstack{30 \\[-0.35ex] {\scriptsize (3.4)}}} \\
\bottomrule
\end{tabular}
\begin{tabular}{@{}@{\hspace{3pt}}r@{\hspace{3pt}}r@{\hspace{3pt}}r@{\hspace{3pt}}r@{\hspace{3pt}}r@{\hspace{3pt}}|@{\hspace{3pt}}r@{}}
\toprule
\multicolumn{6}{c}{\textbf{Description Score ($\Lambda_\T$)}} \\
\midrule
\BLS{} & \PRE{} & \SEC{} & \USPF{} & \PGS{} & \AVG{} \\
\midrule
\textbf{\shortstack{44 \\[-0.35ex] {\scriptsize (4.1)}}} & \shortstack{40 \\[-0.35ex] {\scriptsize (2.4)}} & \shortstack{34 \\[-0.35ex] {\scriptsize (3.0)}} & \shortstack{34 \\[-0.35ex] {\scriptsize (3.1)}} & \shortstack{31 \\[-0.35ex] {\scriptsize (3.2)}} & \shortstack{36 \\[-0.35ex] {\scriptsize (2.1)}} \\[2pt]
\shortstack{41 \\[-0.35ex] {\scriptsize (2.1)}} & \shortstack{37 \\[-0.35ex] {\scriptsize (1.8)}} & \shortstack{38 \\[-0.35ex] {\scriptsize (2.0)}} & \shortstack{37 \\[-0.35ex] {\scriptsize (1.5)}} & \shortstack{34 \\[-0.35ex] {\scriptsize (2.0)}} & \shortstack{38 \\[-0.35ex] {\scriptsize (1.1)}} \\[2pt]
\textbf{\shortstack{46 \\[-0.35ex] {\scriptsize (2.3)}}} & \shortstack{44 \\[-0.35ex] {\scriptsize (2.8)}} & \textbf{\shortstack{44 \\[-0.35ex] {\scriptsize (1.6)}}} & \shortstack{39 \\[-0.35ex] {\scriptsize (1.7)}} & \shortstack{36 \\[-0.35ex] {\scriptsize (1.8)}} & \textbf{\shortstack{42 \\[-0.35ex] {\scriptsize (1.7)}}} \\[2pt]
\shortstack{32 \\[-0.35ex] {\scriptsize (5.0)}} & \shortstack{34 \\[-0.35ex] {\scriptsize (3.1)}} & \shortstack{24 \\[-0.35ex] {\scriptsize (3.1)}} & \shortstack{24 \\[-0.35ex] {\scriptsize (2.8)}} & \shortstack{17 \\[-0.35ex] {\scriptsize (2.7)}} & \shortstack{26 \\[-0.35ex] {\scriptsize (2.6)}} \\[2pt]
\shortstack{40 \\[-0.35ex] {\scriptsize (3.6)}} & \shortstack{41 \\[-0.35ex] {\scriptsize (3.2)}} & \shortstack{39 \\[-0.35ex] {\scriptsize (1.5)}} & \shortstack{33 \\[-0.35ex] {\scriptsize (2.5)}} & \shortstack{28 \\[-0.35ex] {\scriptsize (2.1)}} & \shortstack{36 \\[-0.35ex] {\scriptsize (2.2)}} \\[2pt]
\textbf{\shortstack{47 \\[-0.35ex] {\scriptsize (3.2)}}} & \shortstack{42 \\[-0.35ex] {\scriptsize (2.1)}} & \textbf{\shortstack{38 \\[-0.35ex] {\scriptsize (3.0)}}} & \textbf{\shortstack{39 \\[-0.35ex] {\scriptsize (2.3)}}} & \shortstack{34 \\[-0.35ex] {\scriptsize (2.5)}} & \textbf{\shortstack{40 \\[-0.35ex] {\scriptsize (1.9)}}} \\[2pt]
\textbf{\shortstack{49 \\[-0.35ex] {\scriptsize (2.7)}}} & \textbf{\shortstack{50 \\[-0.35ex] {\scriptsize (2.1)}}} & \textbf{\shortstack{41 \\[-0.35ex] {\scriptsize (2.2)}}} & \textbf{\shortstack{43 \\[-0.35ex] {\scriptsize (1.8)}}} & \textbf{\shortstack{44 \\[-0.35ex] {\scriptsize (1.8)}}} & \textbf{\shortstack{45 \\[-0.35ex] {\scriptsize (1.6)}}} \\
\bottomrule
\end{tabular}
\begin{tabular}{@{}r@{\hspace{3pt}}r@{\hspace{3pt}}r@{\hspace{3pt}}r@{\hspace{3pt}}r|@{\hspace{3pt}}r@{}}
\toprule
 \multicolumn{6}{c}{\textbf{Task Score ($\Lambda_\I$)}} \\
\midrule
 \BLS{} & \PRE{} & \SEC{} & \USPF{} & \PGS{} & \AVG{} \\
\midrule
\shortstack{52 \\[-0.35ex] {\scriptsize (10.0)}} & \shortstack{36 \\[-0.35ex] {\scriptsize (9.6)}} & \shortstack{38 \\[-0.35ex] {\scriptsize (9.5)}} & \shortstack{29 \\[-0.35ex] {\scriptsize (9.3)}} & \shortstack{24 \\[-0.35ex] {\scriptsize (8.5)}} & \shortstack{36 \\[-0.35ex] {\scriptsize (4.2)}} \\[2pt]
 \shortstack{56 \\[-0.35ex] {\scriptsize (9.9)}} & \shortstack{32 \\[-0.35ex] {\scriptsize (9.3)}} & \shortstack{58 \\[-0.35ex] {\scriptsize (9.7)}} & \shortstack{25 \\[-0.35ex] {\scriptsize (8.8)}} & \shortstack{28 \\[-0.35ex] {\scriptsize (9.0)}} & \shortstack{40 \\[-0.35ex] {\scriptsize (6.4)}} \\[2pt]
\textbf{\shortstack{76 \\[-0.35ex] {\scriptsize (8.5)}}} & \textbf{\shortstack{68 \\[-0.35ex] {\scriptsize (9.3)}}} & \textbf{\shortstack{77 \\[-0.35ex] {\scriptsize (8.3)}}} & \textbf{\shortstack{38 \\[-0.35ex] {\scriptsize (9.9)}}} & \shortstack{32 \\[-0.35ex] {\scriptsize (9.3)}} & \textbf{\shortstack{58 \\[-0.35ex] {\scriptsize (8.7)}}} \\[2pt]
 \shortstack{44 \\[-0.35ex] {\scriptsize (9.9)}} & \shortstack{16 \\[-0.35ex] {\scriptsize (7.3)}} & \shortstack{8 \\[-0.35ex] {\scriptsize (5.2)}} & \shortstack{8 \\[-0.35ex] {\scriptsize (5.6)}} & \shortstack{4 \\[-0.35ex] {\scriptsize (3.9)}} & \shortstack{16 \\[-0.35ex] {\scriptsize (6.5)}} \\[2pt]
\textbf{\shortstack{68 \\[-0.35ex] {\scriptsize (9.3)}}} & \textbf{\shortstack{64 \\[-0.35ex] {\scriptsize (9.6)}}} & \shortstack{50 \\[-0.35ex] {\scriptsize (9.8)}} & \shortstack{25 \\[-0.35ex] {\scriptsize (8.8)}} & \shortstack{8 \\[-0.35ex] {\scriptsize (5.4)}} & \shortstack{43 \\[-0.35ex] {\scriptsize (10.3)}} \\[2pt]
\shortstack{56 \\[-0.35ex] {\scriptsize (9.9)}} & \shortstack{40 \\[-0.35ex] {\scriptsize (9.8)}} & \shortstack{50 \\[-0.35ex] {\scriptsize (9.8)}} & \shortstack{21 \\[-0.35ex] {\scriptsize (8.3)}} & \shortstack{16 \\[-0.35ex] {\scriptsize (7.3)}} & \shortstack{37 \\[-0.35ex] {\scriptsize (7.0)}} \\[2pt]
\textbf{\shortstack{68 \\[-0.35ex] {\scriptsize (9.3)}}} & \textbf{\shortstack{76 \\[-0.35ex] {\scriptsize (8.5)}}} & \textbf{\shortstack{65 \\[-0.35ex] {\scriptsize (9.3)}}} & \textbf{\shortstack{50 \\[-0.35ex] {\scriptsize (10.2)}}} & \textbf{\shortstack{56 \\[-0.35ex] {\scriptsize (9.9)}}} & \textbf{\shortstack{63 \\[-0.35ex] {\scriptsize (4.0)}}} \\
\bottomrule
\end{tabular}
\caption{Evidence, description, and task scores (with standard errors) across the \datasetname{} development set. The task score functions like recall for the inserted inconsistency across the models' predicted inconsistencies. Scores are presented as percents (out of 100).}
\label{app:dev:tbl:results:metrics}
\end{table*}

\clearpage
\section{Prompts}\label{app:sec:prompts}
We list the user, system, and grading prompt below.

\onecolumn
\begin{tcblisting}{
  breakable,
  title={User Prompt},
  enhanced jigsaw,
  colback=gray!5,
  colframe=gray!60,
  arc=2mm,
  boxrule=0.4pt,
  listing only,
  listing options={basicstyle=\ttfamily\small, breaklines=true}
}
Here is a new example for you to help me with.
```
<document>
{{problem}}
</document>
```
Find any contradictions or inconsistencies. Start by reasoning within <think> and </think> tags, then return any answers within <answer> and </answer> tags.
\end{tcblisting}
\begin{tcblisting}{
  breakable,
  title={System Prompt},
  enhanced jigsaw,
  colback=gray!5,
  colframe=gray!60,
  arc=2mm,
  boxrule=0.4pt,
  listing only,
  listing options={basicstyle=\ttfamily\small, breaklines=true}
}
I need to check this one document over and make sure there are no semantic errors like inconsistent information or contradicting facts. This really could be anything but not grammar. I can use a spellchecker for that later. For example, I want to make sure that the analysis is correct. The logic is correct. That type of thing.

Rules:
* Format your final answer within <answer> and </answer> tags
* Do NOT include comments within or outside of the tags
* Return the meaningful inconsistencies that matter and in the other that you find within the document. Track only the minimal information needed.
* Ignore spelling and grammar errors.
* Do not repeat the same basic inconsistency in your list of answers.
* Every answer within <answer> will be graded so only put your absolutely final and best results within those tags.

```
<think>...</think>
# first inconsistency
<answer>
<evidence>...</evidence>
<evidence>...</evidence>
<description>...</description>
</answer>
<think>...</think>
# second inconsistency
<answer>
<evidence>...</evidence>
<evidence>...</evidence>
<description>...</description>
</answer>
# etc.
```

First think through any potential error or things you need to check within <think> and </think> tags before giving your answer. Try to limit your thinking just to what you actually need. Format your outputs in the answer structure as shown in the examples below.

# Annotation Definitions

For each contradiction/inconsistency we want to collect the contradictory information present in the document, a description of the inconsistency, and the answers to two flags.

> trigger
Find the trigger first. The trigger is the first place top-to-bottom where the error becomes apparent. The rest of the needed evidence should appear in the document before the trigger. Note that the trigger is not itself always wrong, just the place where it is clear that there is now an error. (The trigger is part of the evidence, which we define next. You don't need to label it separately, it'll just always be the last piece of evidence.)

> evidence
Collect the minimal set of spans including the trigger that identify the error. This will generally be two spans, but could be many more, especially when we find errors in tables. Evidence spans must appear verbatim or else they will be wrong. Present the evidence in order of appearance within the document.

> description
Lastly, include a brief and complete description of why and how the spans form an inconsistency. The description should not include how to fix the contradiction. There will always be more than one way to fix the contradiction. However, showing exactly how and where the error arises would be helpful.

# Annotation Process

1. Find the trigger span. Call the position in the document of the trigger span *j*.
2. Now find the spans that the trigger contradicts, which we call the evidence. Include only evidence that is higher up in the document, with positions i < j.
3. Provide the description of why the spans are inconsistent.

# Some additional notes
- Within the description describe any key structural or background information needed to identify the error. Structural information could include patterns of bolding within a table (always max, or always best.) Background information could include some finance knowledge.

# What if there are two (or more) possible sets of evidence that end with the same trigger?
Pick the set of evidence closest to the trigger. Start by comparing spans of evidence from each candidate set furthest away (nearest the top of the document) and pick the set with the first span closer to the trigger.

- If candidates share the first k spans then compare the first differing span between sets and pick the set with the one closer to the trigger.
- If one set of evidence is a strict subset of another, choose the subset.

# What about meta errors?
- If information is expected to appear (for example in a footnote or appendix) but is missing, describe the problem. Do not attempt to show the absence of missing information. Only highlight the claim as evidence.
- If information is unexpected, then highlight why there is that expectation and the triggering span that violates the expectation.

# Example 1
<document>
Table 1
|  | $ |
|---|---|
| Val Ya | -1 |
| Val Yb | 0 |
| Val Yc | 3 |
| Val Y | 20 |

Val Y is the total of Ya Yb and Yc

{10 pages of text}

Table 2
|  | $ |
|---|---|
| Val X | 1 |
| Val Y | 2 |
| Val Z | 3 |
| Total | 6 |
</document>
<think>Note that the error is only within the first table. The trigger is the caption statement; from that point there is clearly a contradiction in the document. The values within the rows do not sum up to the total.</think>
<answer>
<evidence>Val Ya</evidence><evidence>-1</evidence>
<evidence>Val Yb</evidence><evidence>0</evidence>
<evidence>Val Yc</evidence><evidence>3</evidence>
<evidence>Val Y</evidence><evidence>20</evidence>
<evidence>Val Y is the total of Ya Yb and Yc</evidence>
<description>Val Y is reported to sum up to the value (-1) + 0 + 3 = 2 but is reported as having the value 20.</description>
</answer>


# Example 2
```
<document>
\newcommand{ \yes}[0]{$ \color{violet} \checkmark$}
\newcommand{ \nah}[0]{$ \color{red}$}


\begin{table*}[t]
\small \centering
\begin{tabular}{l|rrr|ccc}
\toprule
{ \bf Dataset} & { \bf \#Docs} & { \bf \#QAs} & { \bf \#Words} & { \bf Multi-page} & { \bf Numeric} & { \bf Tabular} \ \
\midrule
NarrativeQA & 1,572 & 46,765 & 63,000 & \yes & - & - \\
QuALITY & 381 & 6,737 & 5,159 & \yes & - & - \\
PDFTriage & 82 & 908 & 12,000 & \yes & \yes & \yes \\
\midrule
TAT-QA & 2,757 & 16,552 & 260 & - & \yes & \yes \\
FinQA & 2,789 & 8281 & 687 & - & \yes & \yes \\
\midrule
DocFinQA & 801 & 7,437 & 123453 & \yes & \yes & \yes \\
\bottomrule
\end{tabular}
\caption{Comprison of DocFinQA and existing Finance QA and Long Document QA dataset. DocFinQA includes { \bf multi-page} documents with both { \bf numeric} and { \bf tabular} data.}
\label{tab:dataset}
\end{table*}
</document>
<think>There do not appear to be any errors. </think>
<answer></answer>
```

# Example 3
```
<document>
\begin{center}
\begin{tabular}{||c | c c c||}
 \hline
 Col1 & Col2 & Col2 & Col3 \\ [0.5ex]
 \hline\hline
 1 & 6 & 7837 & 787 \\
 \hline
 2 & 7 & 78 & 5415 \\
 \hline
 3 & \textbf{545} & 778 & \textbf{7507} \\
 \hline
 4 & \textbf{545} & \textbf{18744} & 7560 \\
 \hline
 5 & 88 & 788 & 9344 \\ [1ex]
 \hline
\end{tabular}
\end{center}
\label{table:example_9}
</document>
<think>
Each column appears to have the maximum value bolded but there is a mistake: the maximums respectively are 545, 18744, 9344. However, in Col3, 7507 is bolded but 7507 should not be since it is not the maximum in its column. Instead, 9344 should be bolded in Col3.
</think>
<answer>
<evidence>\textbf{7507}</evidence>
<evidence>9344</evidence>
<description>The wrong cell in Column 4 ('Col3') is bolded because it is not the largest value in that column: 9344 is larger than 7507.</description>
</answer>
```

# Example 4
```
<document>

The profit for selling D is low because the materials costs are so high. (See the table.)

| Item | In Stock | Price |
| :-- | :------: | ----: |
| A |   True   | 23.99 | 
| B |   True   | 23.99 |
| C |  False   | 19.99 |
| D |  False   | 42.99 |

For the first time this year, Item C has the highest Price.

</document>
<answer>
<evidence>profit for selling D is low because the materials costs are so high</evidence>
<evidence>See the table.</evidence>
<description>The table is expected to report on profit or materials costs but this information is missing from the table.</description>
</answer>
<answer>
<evidence>C<evidence>
<evidence>19.99<evidence>
<evidence>D<evidence>
<evidence>42.99<evidence>
<evidence>For the first time this year, Item C has the highest Price.</evidence>
<description>Item C does not have the highest price. For example, Item D has a higher price.</description>
</answer>
```

# End of Examples.

Think hard and carefully and give the grade as directed in a compact manner. Feel free to plan and scope out key parts of the document to check and then provide each found inconsistency within answer tags.
Reasoning: high
\end{tcblisting}

\begin{tcblisting}{
  breakable,
  title={Grading Prompt},
  enhanced jigsaw,
  colback=gray!5,
  colframe=gray!60,
  arc=2mm,
  boxrule=0.4pt,
  listing only,
  listing options={basicstyle=\ttfamily\small, breaklines=true}
}
Your job is to grade a set of results. The task was finding the error in a technical document. The student is meant to provide a clear description of the error and evidence.


Compare the ground truth Gold description and evidence compared to the student's output. Then assign a grade of either CORRECT or INCORRECT for whether the student found the error. If the suggested evidence is wrong or incomplete but still makes clear the student found the error then consider the output CORRECT. If the evidence of description pertains to something else in the document please mark the answer INCORRECT.


First, I will give examples of each grade, and then you will grade a new example.


The following are examples of CORRECT predicted answers.
```
Gold answer:
<evidence>\cellcolor[HTML]{FF6D01}\textbf{71.5}</evidence>
<description>The color of 71.5 does not match highlighting pattern of other cells; it should be cyan, not orange.</description>

Predicted answer 1:
<evidence>\cellcolor[HTML]{FF6D01}\textbf{71.5}</evidence>
<description>Highlighting 71.5 orange is wrong. It should be cyan.</description>

Predicted answer 2:
<evidence>"\cellcolor[HTML]{FF6D01}\textbf{71.5}</evidence>
<description>71.5 does not match the pattern of highlighting.</description>

Gold answer:
<evidence>\textbf{62.8}</evidence>
<evidence>61.6</evidence>
<description>In Col 6, 61.6 should be bolded, not 62.8.</description>

Predicted Answer 1:
<evidence>61.6</evidence>
<evidence>62.8</evidence>
<description>61.6 bolded, not 62.8.</description>

Predicted Answer 2:
<evidence>61.6</evidence>
<evidence>\textbf{62.8}</evidence>
<description>In Col 6  in Table {tab:rm_results}, 61.6 appears to be the minimum value and because the table is bolding minimums this value should be bolded for Col 6 and not the currently bolded number. Please change the bolding to fix this error.</description>

```

These predicted answers are all CORRECT because:
    - They contain the important information in the gold target.
    - They do not contain any information that contradicts the gold target.
    - Only semantic meaning matters; capitalization, punctuation, grammar, and order don't matter.
    - Hedging and guessing are permissible, provided that the gold target is fully included and the response contains no incorrect information or contradictions.
    - The predicted answers are all clearly talking about the correct target
    - The json formatting does not need to be perfect or compile for the answer to be correct


The following are examples of INCORRECT predicted answers.
```
Gold answer:
<evidence>\cellcolor[HTML]{FF6D01}\textbf{71.5}</evidence>
<description>The color of 71.5 does not match highlighting pattern of other cells; it should be cyan, not orange.</description>

Predicted answer 1:
<description>There appears to be no semantic errors.</description>

Predicted answer 2:
<evidence>textbf</evidence>
<description>Color is wrong</description>

Predicted answer 3:
<evidence>\cellcolor[HTML]{FF6D01}\textbf{71.5}</evidence>
<description>Bolding seems wrong in Table {tab:dpo_vs_ppo}</description>

Predicted answer 4:
{}
```
These predicted answers are all INCORRECT because:
    - The predicted answer and gold answer must refer to the same error; in the examples above they do not clearly match.
    - The important information from the gold description is not included in the answer



Here is a new example. Simply assign a grade of either CORRECT or INCORRECT.
```
Original Document That Contains The Error from the Gold Answer:
<document>
{{problem_text}}
</document>

Gold answer:
{{gold_answer}}
Predicted answer:
{{model_answer}}
```


Grade the predicted answer of this new question as one of:
A: CORRECT
B: INCORRECT


Just return the letter "[[A]]" or "[[B]]" (corresponding to CORRECT or INCORRECT, respectively) within answer brackets.
\end{tcblisting}
\end{document}